\documentclass{article}

\PassOptionsToPackage{numbers}{natbib}
\usepackage[final]{neurips_2024}

\usepackage[utf8]{inputenc} % allow utf-8 input
\usepackage[T1]{fontenc}    % use 8-bit T1 fonts
\usepackage{hyperref}       % hyperlinks
\usepackage{url}            % simple URL typesetting
\usepackage{booktabs}       % professional-quality tables
\usepackage{amsfonts}       % blackboard math symbols
\usepackage{nicefrac}       % compact symbols for 1/2, etc.
\usepackage{microtype}      % microtypography
\usepackage{xcolor}         % colors
\usepackage{graphicx}
\usepackage{amsmath}
\usepackage{amssymb}
\usepackage{mathtools}
\usepackage{amsthm}

\usepackage{bm}
\usepackage{algorithm}
\usepackage{algpseudocode}
\usepackage{subfigure}
\usepackage{wrapfig}
\usepackage[capitalize,noabbrev]{cleveref}

\usepackage{multirow}
\usepackage{xspace}
\usepackage{caption}
\usepackage{thm-restate}
\usepackage{footmisc}

\newcommand{\z}{\mathbf{z}}
\newcommand{\gz}{\bm{z}}

\newcommand{\Vv}{\bm{v}}

\DeclareRobustCommand{\Q}{\text{\usefont{T1}{ptm}{m}{n} \textit{Q}}}
\newcommand\myldots{\ifmmode\ldots\else\makebox[0.5em][c]{.\hfil.\hfil.}\thinspace\fi}
\newcommand\mycdots{\ifmmode\cdots\else\makebox[0.5em][c]{.\hfil.\hfil.}\thinspace\fi}
\theoremstyle{plain}
\newtheorem{theorem}{Theorem}[section]

\newtheorem{lemma}[theorem]{Lemma}

\theoremstyle{definition}
\newtheorem{definition}[theorem]{Definition}

\theoremstyle{remark}

\title{FIFO-Diffusion: Generating Infinite Videos from Text without Training}

\author{
Jihwan Kim\footnotemark[1]~~$^1$ \quad Junoh Kang\footnotemark[1]~~$^1$ \quad Jinyoung Choi$^1$ \quad Bohyung Han$^{1,2}$ \\
Computer Vision Laboratory, $^1$ECE \& $^2$IPAI, Seoul National University
\\
{\tt\small \{kjh26720,junoh.kang, jin0.choi, bhhan\}@snu.ac.kr}
}

\begin{document}

\maketitle
\def\thefootnote{*}\footnotetext{indicates equal contribution.}\def\thefootnote{\arabic{footnote}}

\renewcommand{\arraystretch}{0.7}
\vspace{-5mm}
\scalebox{0.98}{
    \setlength{\tabcolsep}{1pt} \hspace{-3mm}
    \begin{tabular}{ccccc}
        \includegraphics[width=0.2\linewidth]{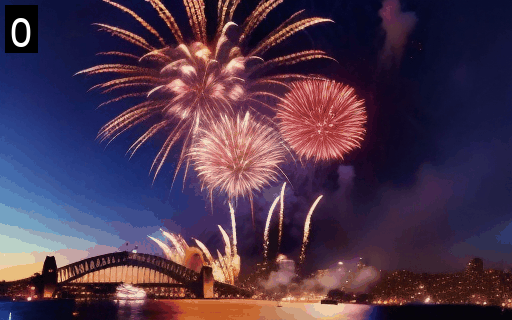} &
        \includegraphics[width=0.2\linewidth]{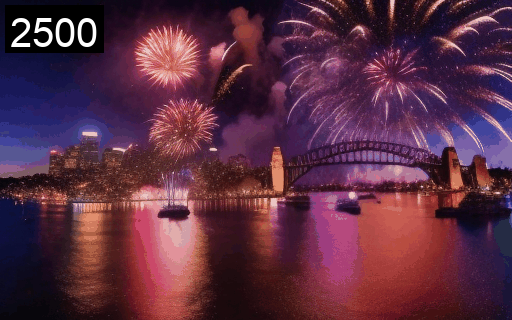} &
        \includegraphics[width=0.2\linewidth]{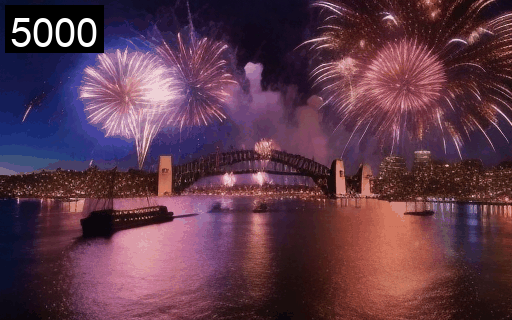} &
        \includegraphics[width=0.2\linewidth]{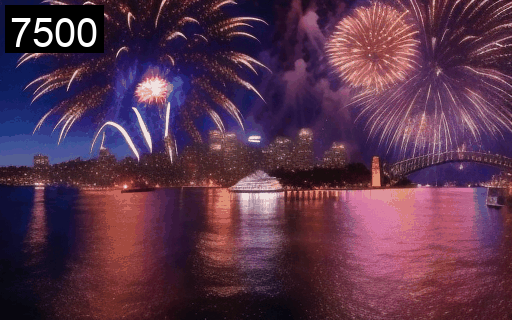} &
        \includegraphics[width=0.2\linewidth]{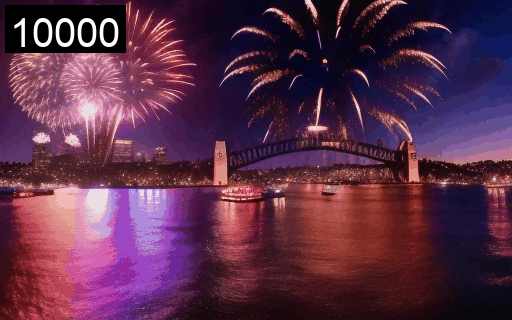} \vspace{0.5mm} \\
        \multicolumn{5}{c}{\small (a) \textsf{"A spectacular ﬁreworks display over Sydney Harbour, 4K, high resolution."}} \vspace{1.5mm}\\
        \includegraphics[width=0.2\linewidth]{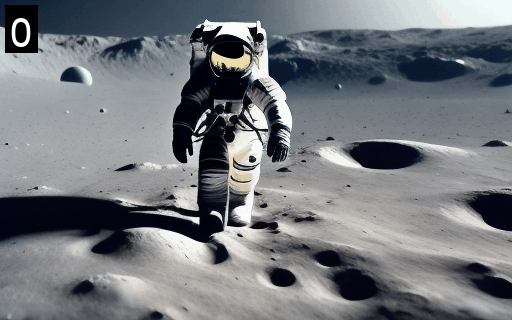} &
        \includegraphics[width=0.2\linewidth]{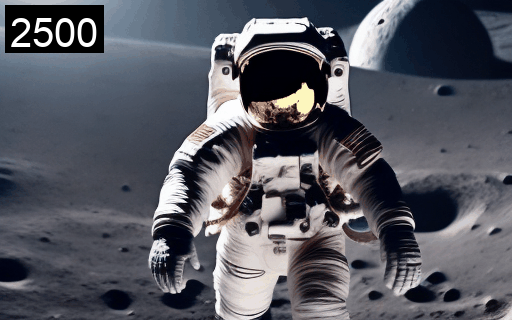} &
        \includegraphics[width=0.2\linewidth]{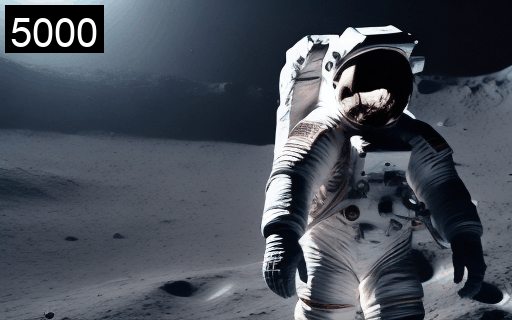} &
        \includegraphics[width=0.2\linewidth]{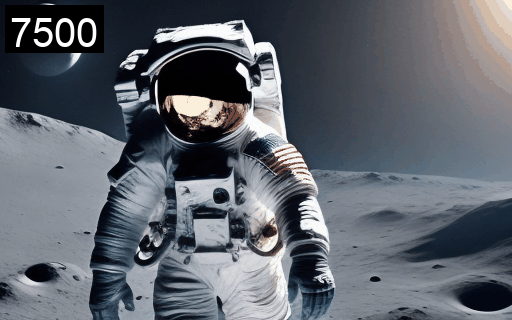} &
        \includegraphics[width=0.2\linewidth]{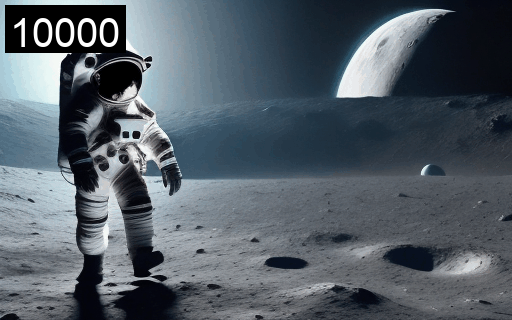} \vspace{0.5mm} \\
        \multicolumn{5}{c}{\small (b) \textsf{"An astronaut walking on the moon's surface, high-quality, 4K resolution."}} \vspace{1.5mm}\\
        \includegraphics[width=0.2\linewidth]{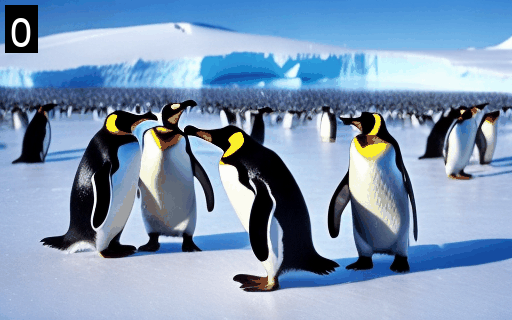} &
        \includegraphics[width=0.2\linewidth]{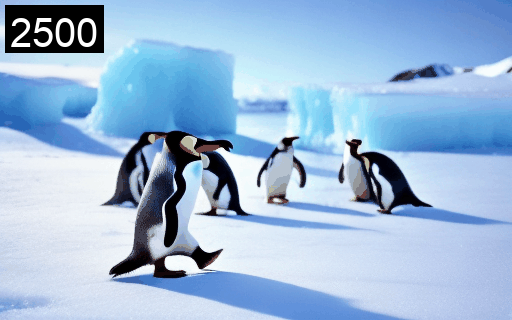} &
        \includegraphics[width=0.2\linewidth]{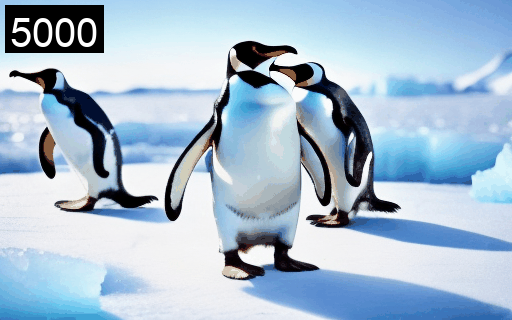} &
        \includegraphics[width=0.2\linewidth]{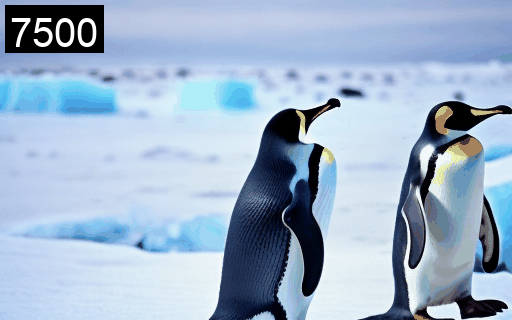} &
        \includegraphics[width=0.2\linewidth]{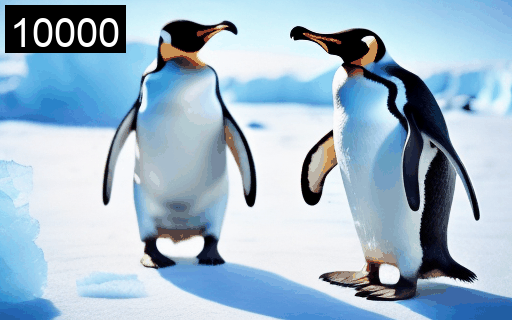} \vspace{0.5mm} \\
        \multicolumn{5}{c}{\small (c) \textsf{"A colony of penguins waddling on an Antarctic ice sheet, 4K, ultra HD."}} \\
    \end{tabular}
}
\captionof{figure}{
    Illustration of 10K-frame long videos generated by FIFO-Diffusion based on a pretrained text-conditional video generation model, VideoCrafter2~\citep{chen2024videocrafter2}.
    The number at the top-left corner of each image indicates the frame index.
    The results clearly show that FIFO-Diffusion can generate extremely long videos effectively based on the model trained on short clips~(16 frames) without quality degradation while preserving the dynamics and semantics of scenes. 
    }
    \vspace{3mm}
\label{fig:qual_long}

% !TEX root = ./../main.tex

\begin{abstract}
We propose a novel inference technique based on a pretrained diffusion model for text-conditional video generation.
Our approach, called FIFO-Diffusion, is conceptually capable of generating infinitely long videos without additional training.
This is achieved by iteratively performing diagonal denoising, which simultaneously processes a series of consecutive frames with increasing noise levels in a queue; our method dequeues a fully denoised frame at the head while enqueuing a new random noise frame at the tail.
However, diagonal denoising is a double-edged sword as the frames near the tail can take advantage of cleaner frames by forward reference but such a strategy induces the discrepancy between training and inference.
Hence, we introduce latent partitioning to reduce the training-inference gap and lookahead denoising to leverage the benefit of forward referencing.
Practically, FIFO-Diffusion consumes a constant amount of memory regardless of the target video length given a baseline model, while well-suited for parallel inference on multiple GPUs.
We have demonstrated the promising results and effectiveness of the proposed methods on existing text-to-video generation baselines.
Generated video examples and source codes are available at our project page\footnote{\url{https://jjihwan.github.io/projects/FIFO-Diffusion}.\label{fn:project}}.
\end{abstract}
% !TEX root = ./../main.tex
\section{Introduction}
\label{sec:introduction}
Diffusion probabilistic models have achieved remarkable success in generating high-quality images~\citep{ho2020denoising,song2021scorebased,dhariwal2021diffusion,rombach2022high}.
On top of the success in the image domain, there has been rapid progress in the generation of videos~\citep{ho2022video,singer2022make,zhou2022magic,wang2023modelscope}.
Despite the progress, long video generation still lags behind compared to image generation.
One reason is that video diffusion models~(VDMs) often consider a video as a single 4D tensor with an additional axis corresponding to time, which prevents the models from generating videos at scale.
An intuitive approach to generating a long video is autoregressive generation, which iteratively predicts a future frame given the previous ones.
However, in contrast to the transformer-based models~\citep{hong2023cogvideo,villegas2023phenaki}, diffusion-based models cannot directly adopt the autoregressive generation strategy due to the heavy computational costs incurred by iterative denoising steps for a single frame generation.
Instead, several recent works~\citep{ho2022video,he2022latent,voleti2022mcvd,luo2023videofusion,chen2023seine,blattmann2023align} adopt a chunked autoregressive generation strategy, which predicts several frames in parallel conditioned on few preceding ones, consequently reducing computational burden.
While these approaches are computationally tractable, they often leads to temporal inconsistency and discontinuous motion, especially between the chunks predicted separately, because the model captures a limited temporal context available in the last few---only one or two in practice---frames.

To address the limitation, we propose a novel inference technique, FIFO-Diffusion, which realizes arbitrarily long video generation without training based on a pretrained video generation model for short clips.
Our approach effectively alleviates the limitations of the chunked autoregressive method by enabling every frame to refer to a sufficient number of preceding frames. 
Our approach generates frames through diagonal denoising~(\cref{subsec:diagonal}) in a first-in-first-out manner using a queue, which contains a sequence of frames with different---monotonically increasing---noise levels over time.
At each step, a completely denoised frame at the head is popped out from the queue while a new random noise image is pushed back at the tail.
Diagonal denoising offers both advantage and disadvantage; noisier frames benefit from referring to cleaner ones while the model may suffer from training-inference gap because video models are generally trained to denoise frames with the same noise level.
To overcome this trade-off and embrace the advantage of diagonal denoising, we propose latent partitioning~(\cref{subsec:latent_partitioning}) and lookahead denoising~(\cref{subsec:look_ahead_denoising}). 
Latent partitioning reduces training-inference gap by narrowing the range of noise levels in to-be-denoised frames and enables inference with finer steps.
Lookahead denoising allows to-be-denoised frames to reference cleaner frames, thereby performing more accurate noise prediction.
Furthermore, both latent partitioning and lookahead denoising offer parallelizability on multiple GPUs.

\vspace{2mm}
Our main contributions are summarized below.
\begin{itemize}
    \item We propose FIFO-Diffusion through diagonal denoising, which is a training-free video generation technique for VDMs pretrained on short clips.
    Our approach denoises images with different noise levels for seamless video generation, enabling us to generate arbitrarily long videos.
    \item We introduce latent partitioning and lookahead denoising, which respectively reduce the training-inference gap incurred by diagonal denoising and allow the reference to less noisy frames for denoising, improving generation quality.
    \item FIFO-Diffusion requires a constant amount of memory regardless of the length of the generated videos given a baseline model. It is straightforward to run FIFO-Diffusion in parallel on multiple GPUs.
    \item Our experiments on four strong baselines, based on the U-Net~\citep{ronneberger2015unet} or DiT~\citep{peebles2023dit} architectures, show that FIFO-Diffusion generates extremely long videos including natural motion without degradation on quality over time.
\end{itemize}
% !TEX root = ./../main.tex

\section{Text-to-Video Diffusion Models}
\label{sec:pretrained_models}

We summarize the basic idea of text-conditional video generation techniques based on diffusion models.
They consist of a few key components: an encoder $\text{Enc}(\cdot)$, a decoder $\text{Dec}(\cdot)$, and a noise prediction network $\bm{\epsilon}_\theta(\cdot)$.
They learn the distribution of videos corresponding to text conditions, and the video is denoted by $\Vv \in \mathbb{R}^{f \times H \times W \times 3}$, where $f$ is the number of frames and $H\times W$ indicates the image resolution.
The encoder projects each frame onto the latent image space and the decoder reconstructs the frame from the latent.
A video latent $\z_0 =\text{Enc}(\Vv) = [\gz^1_0;\text{\myldots};\gz^f_0]\in \mathbb{R}^{f \times h \times w \times c}$ is obtained by concatenating projected frames and the latent diffusion model is trained to denoise its perturbed version, $\z_t$.
For noise $\bm{\epsilon} \sim \mathcal{N}(\mathbf{0}, \mathbf{I})$, a diffusion time step $t\sim \mathcal{U}([1,\text{\myldots},T])$, and a text condition $\bm{c}$, the model is trained to minimize the following loss:
\begin{align}\label{eq:diffusion_loss}
    \mathbb{E}_{\Vv,\bm{\epsilon}, t}
    \left[
        ||\bm{\epsilon}_\theta(\z_t;\bm{c},t) - \bm{\epsilon}||
    \right],
\end{align}
where the perturbed latent, $\z_t=s_t\z_0 + \sigma_t\bm{\epsilon}$, is obtained using predefined constants $\{s_t\}_{t=0}^T$ and $\{\sigma_t\}_{t=0}^T$, with the constraints $s_0=1$, $\sigma_0=0$ and $\sigma_T/s_T \gg 1$.

Following a time step schedule, $0=\tau_0 < \tau_1< \text{\mycdots} <\tau_{S}=T$, initialized by a diffusion scheduler, the model generates a video by iteratively denoising $[\gz_{\tau_S}^1;\text{\myldots};\gz_{\tau_S}^f] \sim \mathcal{N}(\mathbf{0}, \mathbf{I})$ over $S$ steps using a sampler $\Phi(\cdot)$ such as the DDIM sampler.
Each denoising step is expressed as
\begin{align}\label{eq:diffusion_step}
    [\gz_{\tau_{t-1}}^1;\text{\myldots};\gz_{\tau_{t-1}}^f] = \Phi([\gz_{\tau_{t}}^1;\text{\myldots};\gz_{\tau_{t}}^f], [\tau_t;\text{\myldots};\tau_t], \bm{c}; \bm{\epsilon}_\theta),
\end{align}
where $\gz_{\tau_t}^i$ denotes the latent of the $i^\text{th}$ frame at time step $\tau_t$.

\section{FIFO-Diffusion}
\label{sec:fifo-diffusion}

This section discusses how FIFO-Diffusion generates long videos consisting of $N$ frames using a pretrained model only for $f$ frames ($f \ll N$).
The proposed approach iteratively employs diagonal denoising~(\cref{subsec:diagonal}) over a predefined number of frames with different levels of noise.
Our method also incorporates latent partitioning~(\cref{subsec:latent_partitioning}) and lookahead denoising~(\cref{subsec:look_ahead_denoising}) to improve the output quality of FIFO-Diffusion based on diagonal denoising.

\begin{figure}[t]
    \centering
    \includegraphics[width=0.7\linewidth]{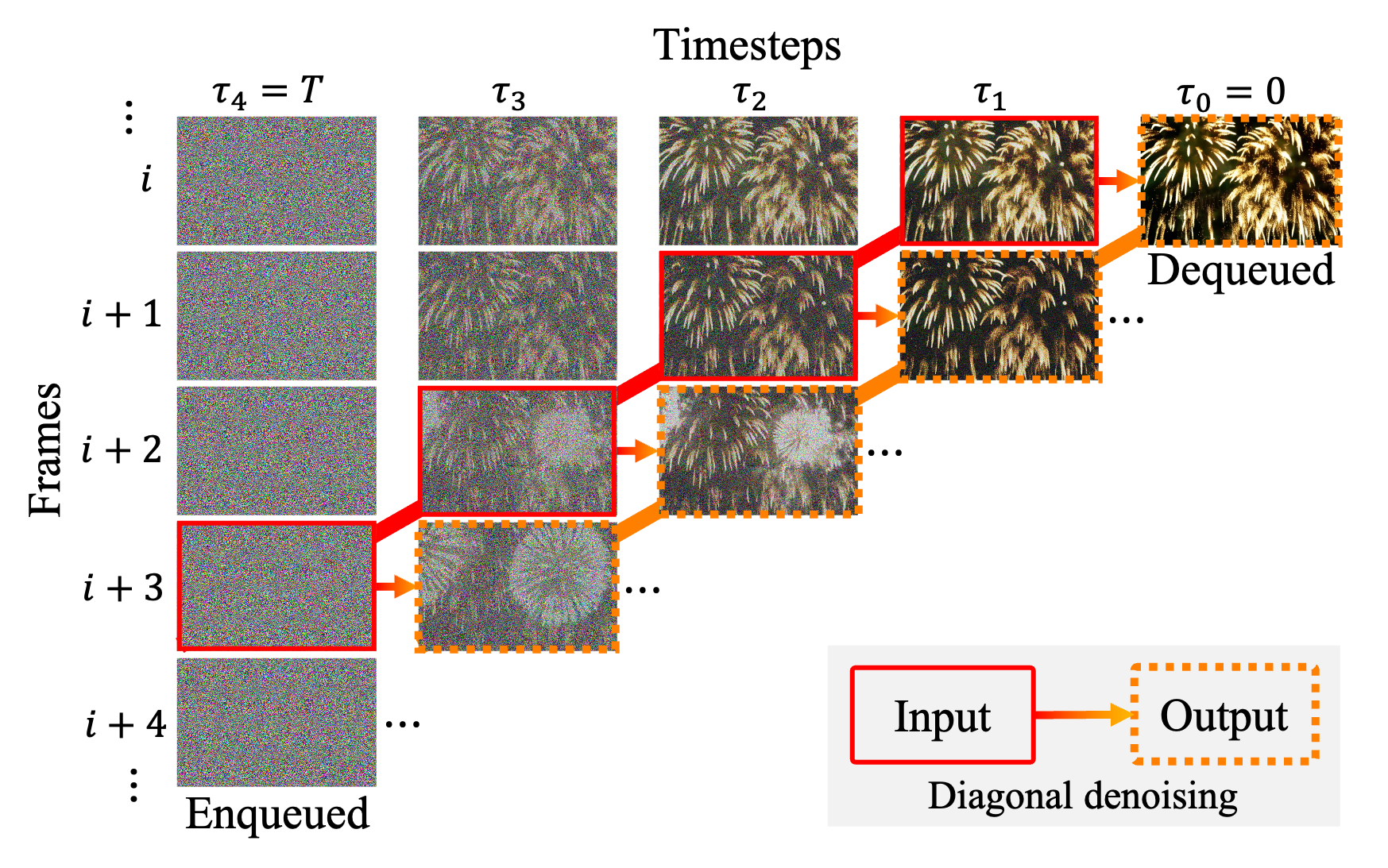} 
    \caption{
    Illustration of diagonal denoising with $f=4$.
    The frames surrounded by solid lines are model inputs while frames surrounded by dotted line are their denoised version.
    After denoising, the fully denoised instance at the top-right corner is dequeued while random noise is enqueued.
    }  \label{fig:diagonal_denoising}
%    \vspace{-3mm}
\end{figure}

\subsection{Diagonal denoising}
\label{subsec:diagonal}

Diagonal denoising processes a series of consecutive frames with increasing noise levels as depicted in \cref{fig:diagonal_denoising}.
To be specific, for a time step schedule $0=\tau_0 < \tau_1 < \text{\mycdots} < \tau_f=T$, each denoising step is defined as
\begin{align} \label{eq:diagonal_denoising_step}
    [\gz_{\tau_0}^1;\text{\myldots};\gz_{\tau_{f-1}}^f] = \Phi([\gz_{\tau_1}^1;\text{\myldots};\gz_{\tau_f}^f], [\tau_1;\text{\myldots};\tau_f], \bm{c}; \bm{\epsilon}_\theta).
\end{align}
Note that the latents along the diagonal, $[\gz_{\tau_1}^1;\text{\myldots};\gz_{\tau_{f}}^f]$, are stored in a queue, $\Q$, and diagonal denoising jointly considers the latents with different noise levels of $[\tau_1;\text{\myldots};\tau_f]$, in contrast to the standard method specified in \cref{eq:diffusion_step}.
\cref{alg:fifo_algorithm} in \cref{app:algorithm} illustrates how diagonal denoising in FIFO-Diffusion works.
After each denoising step with $[\gz_{\tau_1}^1;\text{\myldots};\gz_{\tau_{f}}^f]$, the foremost frame is dequeued as it arrives at the noise level $\tau_0=0$, and the new latent at noise level $\tau_f$ is enqueued.
As a result, the model generates frames in a first-in-first-out manner.

Additionally, the initial diagonal latents $[\gz_{\tau_1}^1;\text{\myldots};\gz_{\tau_{f}}^{f}]$ to initiate the diagonal denoising can be generated from $f$ random noises at time step $\tau_f$, similar to the the process described above.
Notably, our approach does not require pregenerated videos or additional training for the initial latent construction.
The detailed algorithm is presented in \cref{alg:fifo_init} in \cref{app:algorithm}.

\begin{figure}[t]
    \centering
    \setlength{\tabcolsep}{1mm}
    \scalebox{1}{
    \begin{tabular}[t]{cccc}
        \hspace{-4mm}\includegraphics[width=0.35\linewidth]{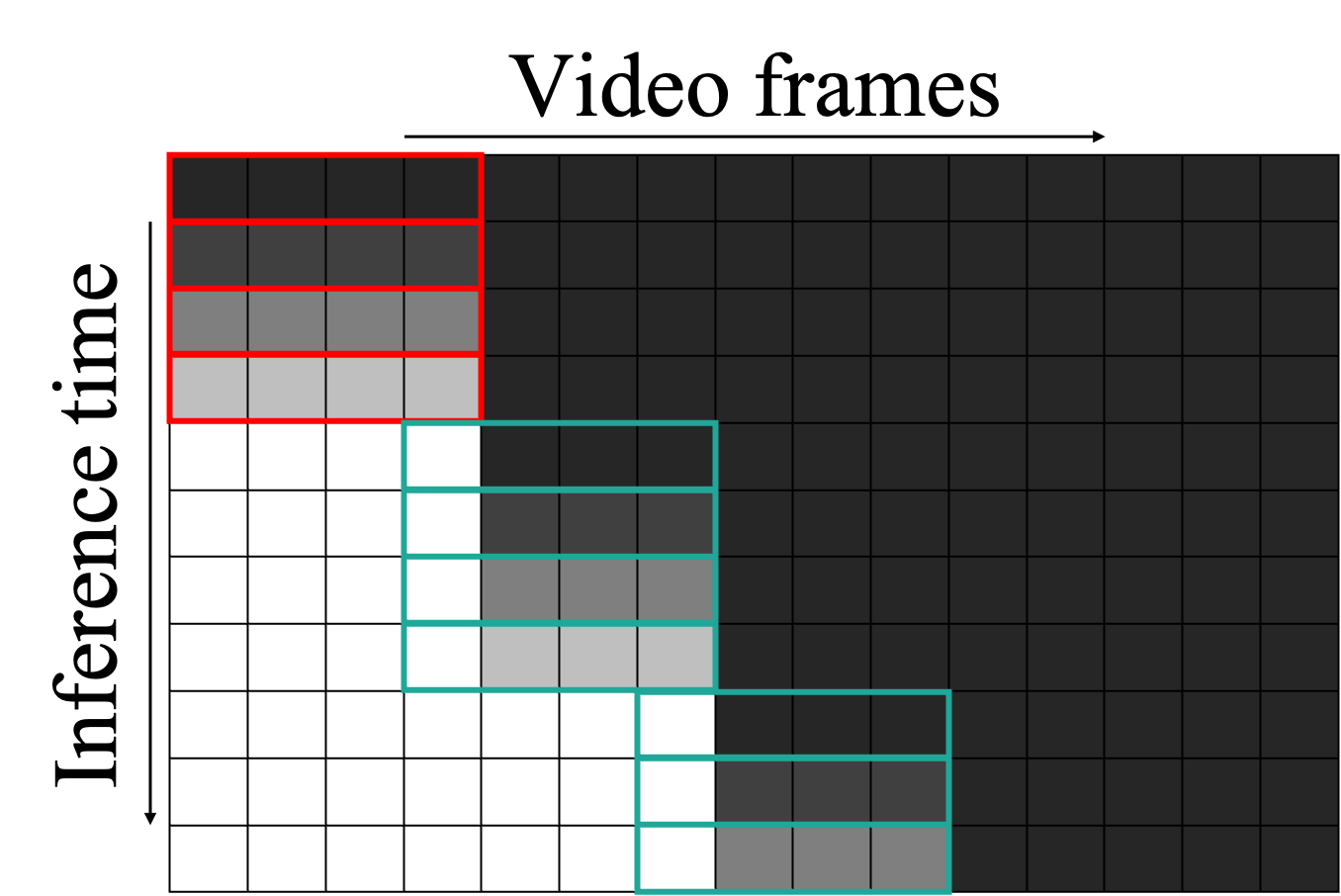} & & & \quad \quad
        \includegraphics[width=0.35\linewidth]{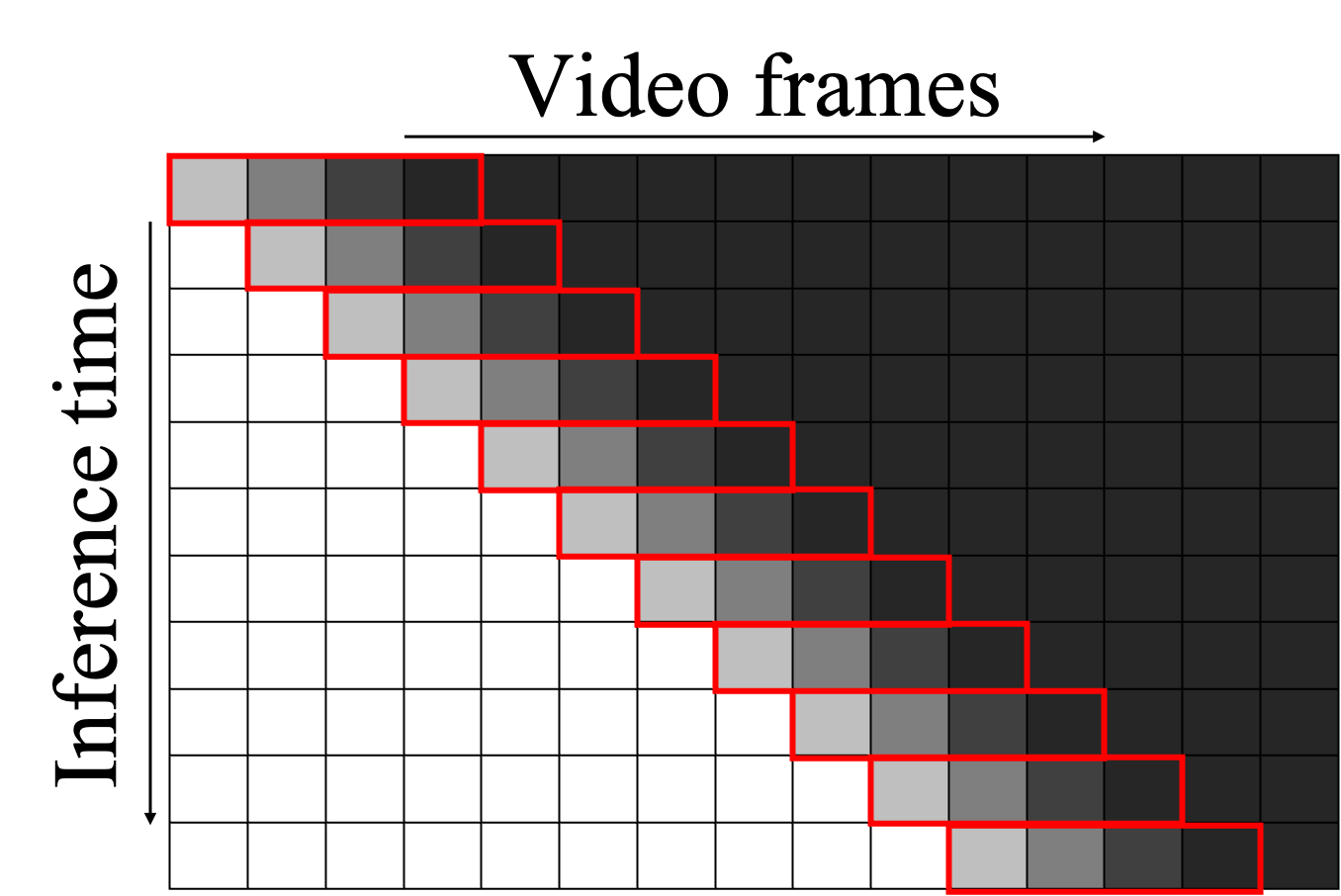} \\
        {(a) Chunked autoregressive} & & &{(b) FIFO-Diffusion}
    \end{tabular}
    }
    \caption{
        Comparison between the chunked autoregressive methods and FIFO-Diffusion proposed for long video generation.
        The random noises~(black) are iteratively denoised to image latents~(white) by the models.
        The red boxes indicate the denoising network in the pretrained base model while the green boxes denote the prediction network obtained by additional training.
    }
    \label{fig:compare}
    \vspace{-3mm}
\end{figure}

FIFO-Diffusion takes $f$ frames as input, regardless of the target video length, and generates an arbitrary number of frames by producing one frame per iteration using a sliding window approach.
Note that generating $N~(\gg f)$ frames for a video requires $\mathcal{O}(f)$ memory in each step~(see \cref{tab:memory_time}), which is independent of $N$.

Diagonal denoising allows us to generate consistent videos by sequentially propagating context to later frames.
\cref{fig:compare} illustrates the conceptual difference between chunked autoregressive methods~\citep{ho2022video,he2022latent,voleti2022mcvd,luo2023videofusion,chen2023seine,blattmann2023align} and FIFO-Diffusion.
The former often struggles to maintain long-term context across chunks since their conditioning---only the last generated frame---lacks contextual information propagated from previous frames.
In contrast, diagonal denoising progresses through the frame sequence with a stride of 1, allowing each frame to reference a sufficient number of preceding frames during generation. 
This approach enables the model to naturally extend the local consistency of a few frames to longer sequences. 
Additionally, FIFO-Diffusion requires no subnetworks or extra training, depending solely on a base model. 
This distinguishes it from existing autoregressive methods, which often require an additional prediction model or fine-tuning for masked frame outpainting.

\subsection{Latent partitioning}
\label{subsec:latent_partitioning}

Although diagonal denoising enables infinitely long video generation, it introduces a training-inference gap, as the model is trained to denoise all frames at uniform noise levels. 
To address this, we aim to reduce noise level differences in the input latents by extending the queue length $n$ times (from $f$ to $nf$ with $n > 1$), partitioning it into $n$ blocks, and processing each block independently. 
Note that the extended queue length increases the number of inference steps.

\begin{figure}[t]
    \centering
    \scalebox{1}{
        \setlength{\tabcolsep}{1pt}
        \begin{tabular}{ccc}
            \includegraphics[width=1\linewidth]{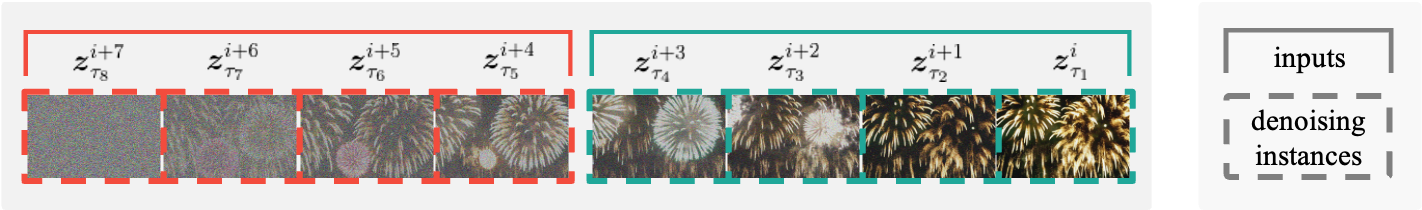} \\
            (a) Latent partitioning \vspace{2mm}\\
            \includegraphics[width=1\linewidth]{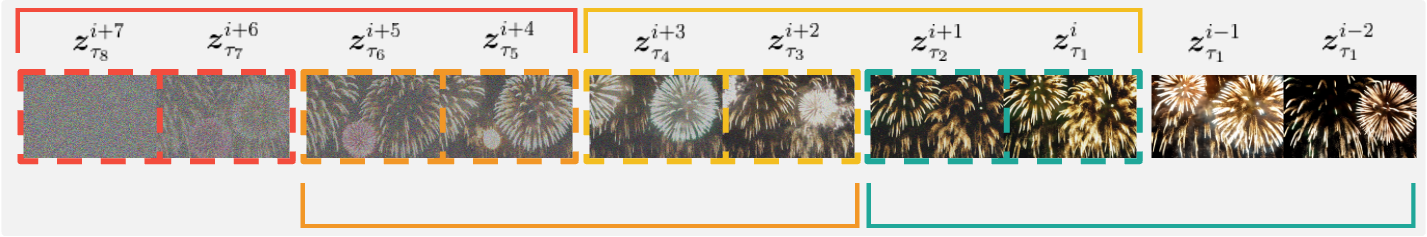} \\
            (b) Lookahead denoising \\
        \end{tabular}
    }
    \caption{
        Illustration of latent partitioning and lookahead denoising where $f=4$ and $n=2$.
        (a) Latent partitioning divides the diffusion process into $n$ parts to reduce the maximum noise level difference.
        (b) Lookahead denoising on (a) enables all frames to be denoised with an adequate number of former frames at the expense of two times more computation than (a).
    }
    \label{fig:lp_ld}
   \vspace{-3mm}
\end{figure}

\cref{alg:fifo_algorithm_lp} in \cref{app:algorithm} provides the procedure of FIFO-Diffusion with latent partitioning.
Let a queue $Q$ has diagonal latents $[\gz_{\tau_1}^1;\text{\myldots};\gz_{\tau_{nf}}^{nf}]$.
We partition $Q$ into $n$ blocks, $[\Q_0;\text{\myldots};\Q_{n-1}]$, of equal size $f$, then each block $\Q_k$ contains the latents at time steps $\bm{\tau}_k = [ \tau_{kf+1}; \text{\myldots}; \tau_{(k+1)f}]$.
Next, we apply diagonal denoising to each block in a divide-and-conquer manner~(See \cref{fig:lp_ld}~(a)).
At $k=0,\text{\mycdots},n-1$, each denoising step updates the queue as follows:
\begin{align} \label{eq:lp}
    \Q_k \leftarrow \Phi(\Q_k, \bm{\tau}_k, \bm{c}; \bm{\epsilon}_\theta).
\end{align}

Latent partitioning offers three key advantages for diagonal denoising.
First, it significantly reduces the maximum noise level difference between the latents from $|\sigma_{\tau_{nf}}-\sigma_{\tau_{1}}|$ to $\max_{k}{|\sigma_{\tau_{(k+1)f}}-\sigma_{\tau_{kf+1}}}|$.
The effectiveness of latent partitioning is supported theoretically and empirically by \cref{thm1} and \cref{tab:ablation}, respectively.
Second, latent partitioning improves throughput of inference by processing partitioned blocks in parallel on multiple GPUs~(see \cref{tab:memory_time}).
Last, it allows the diffusion process to leverage a large number of inference steps, $nf$ ($n \ge 2$), reducing discretization error during inference.

We now show in~\cref{thm1} that the gap incurred by diagonal denoising is bounded by the maximum noise level difference, which implies that the error can be reduced by narrowing the noise level differences of model inputs.

\begin{definition} \label{def0}
    We define $\z^{\text{vdm}}_{t}\coloneqq[\gz_{t}^1;\text{\myldots};\gz_{t}^f]$,
    where $\gz_{t}^i$ is the latent of the $i^\text{th}$ frame at time step $t$~(noise level of $\sigma_t = ct$ for a constant $c$).
    $\z^{\text{vdm}}_{t}$ satisfies the following ODE from \citep{karras2022elucidating}:
    \begin{align}
    	d\z_{t}^\text{vdm} = c\cdot\bm{\epsilon}(\z^{\text{vdm}}_{t}, t\cdot\bm{1})dt \label{thm1:ode},
    \end{align}
    for $\bm{1}=[1;\text{\myldots};1]$ and $\bm{\epsilon}(\cdot)$ is the scaled score function $-\sigma\nabla_{\z} \log p(\cdot)$.
\end{definition}

\begin{lemma} \label{lemma1}
    If $\bm{\epsilon}(\cdot)$ is bounded, then
    \begin{align*}
        ||\gz_{t}^i - \gz_{s}^i|| = O(|t - s|) ~~\text{for} ~~\forall i.
    \end{align*}
\end{lemma}

\begin{proof}
    Refer to \cref{app:proof}.
\end{proof}

\begin{theorem} \label{thm1}
    Assume the system satisfies the following two hypotheses:
    \begin{align*}
        &\text{(Hypothesis 1) $\bm{\epsilon}(\cdot)$ is bounded.} \\
        &\text{(Hypothesis 2) The diffusion model $\bm{\epsilon}_\theta(\cdot)$ is $K$-Lipschitz continuous.}
    \end{align*}
    Then, for diagonal latents $\z^{\text{diag}} = [\gz_{\tau_1}^1;\text{\myldots};\gz_{\tau_f}^f]$ and corresponding time steps $\bm{\tau}^{\text{diag}}=[\tau_1;\text{\myldots};\tau_f]$,
    \begin{align} 
        ||\bm\epsilon_\theta(\z^{\text{diag}}, \bm{\tau}^\text{diag})^i - \bm\epsilon(\z^{\text{vdm}}_{\tau_i}, \tau_i\cdot\bm{1})^i||   
        = ||\bm\epsilon_\theta(\z^{\text{vdm}}_{\tau_i}, \tau_i\cdot\bm{1})^i - \bm\epsilon(\z^{\text{vdm}}_{\tau_i}, \tau_i\cdot\bm{1})^i|| + O(|\sigma_{\tau_f} - \sigma_{\tau_1}|),  \label{thm1:main} 
    \end{align}
    where the $\bm\epsilon_\theta(\cdot)^i$ and $\bm\epsilon(\cdot)^i$ are $i^\text{th}$ element of $\bm\epsilon_\theta(\cdot)$ and $\bm\epsilon(\cdot)$, and $\tau_1<\text{\myldots}<\tau_f$.
    In other words, the error introduced by diagonal denoising is bounded by the noise level difference.
\end{theorem}
\begin{proof}
    The left-hand side of \cref{thm1:main} is bounded as:
    \begin{align*}
        ||\bm\epsilon_\theta(\z^{\text{diag}}, \bm{\tau}^\text{diag})^i &- \bm\epsilon(\z^{\text{vdm}}_{\tau_i}, \tau_i\cdot\bm{1})^i|| \\
        &\leq  
        ||\bm\epsilon_\theta(\z^{\text{diag}}, \bm{\tau}^\text{diag})^i - \bm\epsilon_\theta(\z_{\tau_i}^{\text{vdm}}, \tau_i \cdot \bm{1})^i||
        + ||\bm\epsilon_\theta(\z_{\tau_i}^{\text{vdm}}, \tau_i \cdot \bm{1})^i - \bm\epsilon(\z^{\text{vdm}}_{\tau_i}, \tau_i\cdot\bm{1})^i||,
    \end{align*}
    by triangle inequality. 
    Then, the first term of the right-hand side satisfies the following inequality:
    \begin{align*}
        ||\bm\epsilon_\theta (\z^{\text{diag}}, \bm{\tau}^\text{diag})^i - \bm\epsilon_\theta(\z_{\tau_i}^{\text{vdm}}, \tau_i \cdot \bm{1})^i||
        &\leq K||(\z^{\text{diag}}, \bm{\tau}^\text{diag}) - (\z^\text{vdm}_{\tau_i}, \tau_i \cdot \bm{1})|| \\
        &\leq K \sum_{j=1}^{f} (|| \gz_{\tau_j}^j - \gz_{\tau_i}^j || + |\tau_j - \tau_i|) 
        = O(|\sigma_{\tau_f} - \sigma_{\tau_1}|),
    \end{align*}
    which is from the Lipshitz continuity and \cref{lemma1}. 
    Furthermore, we provide justification for \textit{(Hypothesis 2)} in \cref{app:justification}.
\end{proof}

\subsection{Lookahead denoising}
\label{subsec:look_ahead_denoising}

\begin{wrapfigure}{r}{0.5\linewidth}
    \centering
%    \vspace{-5mm}
    \includegraphics[width=0.9\linewidth]{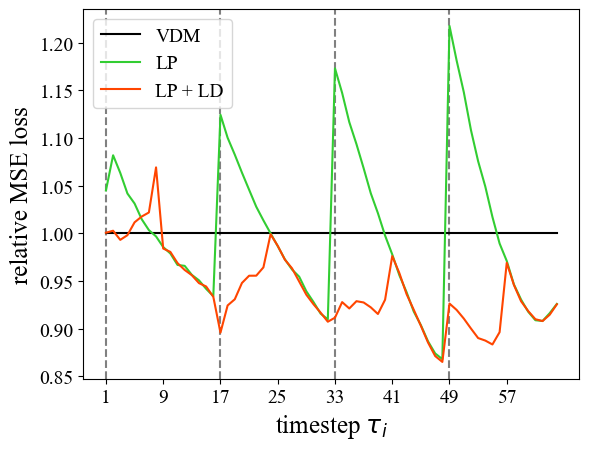} 
    \caption{
    	The relative MSE losses of the noise prediction of $\gz_{\tau_i}^i$~(see \cref{eq:mse_loss}) when $n=4$.
        `VDM' indicates the original denoising strategy as a reference line. `LP' and `LD' denote latent partitioning and lookahead denoising, respectively.}
        \vspace{-3mm}
\label{fig:mse_loss}
\end{wrapfigure}

Although our diagonal denoising introduces training-inference gap, it is advantageous in another respect because noisier frames benefit from observing cleaner ones, leading to more accurate denoising.
As empirical evidence, \cref{fig:mse_loss} shows the relative MSE losses in noise prediction of diagonal denoising  with respect to the original denoising strategy.
The formal definition of the relative MSE is given by
\begin{align} 
    \cfrac{||\bm\epsilon_\theta(\z^{\text{diag}}, \bm{\tau}^\text{diag})^i - \bm\epsilon(\z^{\text{vdm}}_{\tau_i}, \tau_i\cdot\bm{1})^i||_2}
    {||\bm\epsilon_\theta(\z^{\text{vdm}}_{\tau_i}, \tau_i\cdot\bm{1})^i - \bm\epsilon(\z^{\text{vdm}}_{\tau_i}, \tau_i\cdot\bm{1})^i||_2}.
    \label{eq:mse_loss}
\end{align}

As depicted in \cref{fig:lp_ld}~(b), we estimate noise only for the benefited later half of the frames.
In other words, we perform diagonal denoising with a stride of $f' = \lfloor \frac{f}{2} \rfloor$, updating only the last $f'$ frames to ensure that each frame is denoised with reference to a sufficient number---at least $f'$---of clearer frames.
Precisely, for $k=0,\text{\mycdots},2n-1$, each denoising step updates the queue as
\begin{align} \label{eq:ld}
    \Q_k^{f'+1:f} \leftarrow \Phi(\Q_k,\bm{\tau}_k,\bm{c};\epsilon_\theta)^{f'+1:f}.
\end{align}
\cref{alg:fifo_algorithm_full} in \cref{app:algorithm} outlines the detailed procedure of FIFO-Diffusion with lookahead denoising.
We illustrate the effectiveness of lookahead denoising with the red line in~\cref{fig:mse_loss}.
Except for a few early time steps, lookahead denoising enhances the baseline models noise prediction performance, nearly eliminating the training-inference gap described in~\cref{subsec:latent_partitioning}.
Note that, this approach requires twice the computation of the original diagonal denoising since we only update the half of the queue each step.
However, the concerns about the additional computational overhead are easily addressed via parallelization in the same manner as latent partitioning~(see \cref{tab:memory_time}).

% !TEX root = ./../main.tex

\section{Experiment}
\label{sec:experiment}

This section presents the examples generated by existing long video generation methods including FIFO-Diffusion, and evaluates their performance qualitatively and quantitatively.
We also perform the ablation study to verify the benefit of latent partitioning and lookahead denoising introduced in FIFO-Diffusion.

\subsection{Implementation details}
\label{subsec:implementation}

We implement FIFO-Diffusion based on existing open-source text-to-video diffusion models trained on short video clips, including three U-Net-based models, VideoCrafter1~\citep{chen2023videocrafter1},  VideoCrafter2~\citep{chen2024videocrafter2}, and zeroscope\footnote{\url{https://huggingface.co/cerspense/zeroscope\_v2\_576w}}, as well as a DiT-based model, Open-Sora Plan\footnote{\url{https://github.com/PKU-YuanGroup/Open-Sora-Plan}}.
We employ the DDIM sampling~\citep{song2021denoising} with $\eta\in\{0.5, 1\}$.
\cref{app:implementation_details} provides more details about our implementations.

For quantitative evaluation, we measure $\text{FVD}_{128}$~\citep{unterthiner2018towards} and IS~\citep{salimans2016is} scores using Latte~\citep{ma2024latte} as a base model, which is a DiT-based video model trained on UCF-101~\citep{soomro2012ucf}.
We generate 2,048 videos with 128 frames each to calculate $\text{FVD}_{128}$, and randomly sample a 16-frame clip from each video to measure IS score, following evaluation guidelines in \citep{skorokhodov2022styleganv}. To calculate computational cost, we adopt VideoCrafter2 as the baseline model, using a DDPM scheduler with 64 inference steps on A6000 GPUs.

\subsection{Qualitative results}
\label{subsec:qual}

\begin{figure}[!t]
    \centering
    \renewcommand{\arraystretch}{0.7}
\scalebox{0.98}{
    \setlength{\tabcolsep}{1pt} \hspace{-0.5mm}
        \hspace{-3mm}
        \begin{tabular}{ccccc}
	        \includegraphics[width=0.2\linewidth]{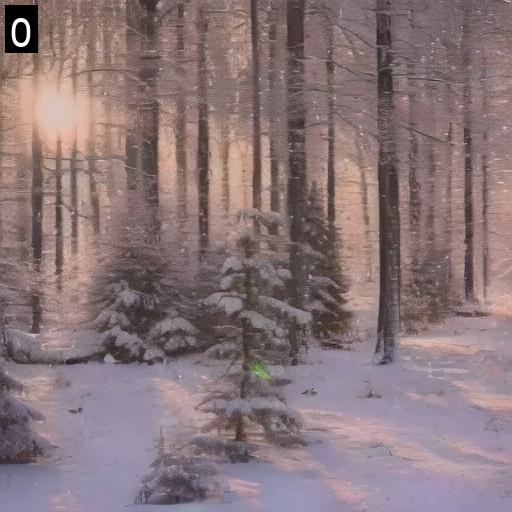} &
            \includegraphics[width=0.2\linewidth]{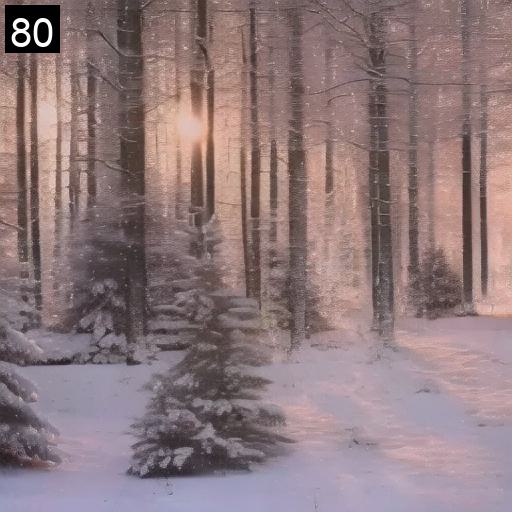} &
            \includegraphics[width=0.2\linewidth]{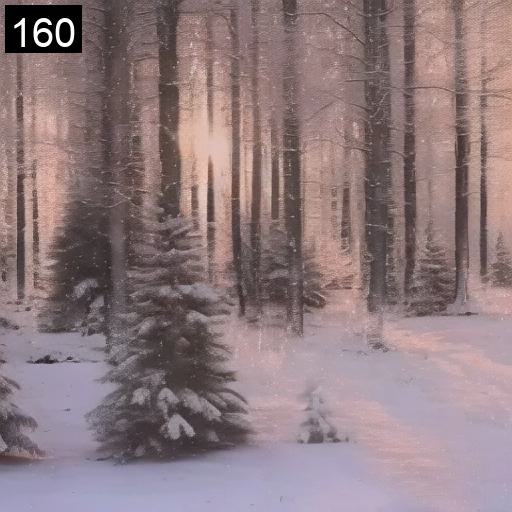} &
            \includegraphics[width=0.2\linewidth]{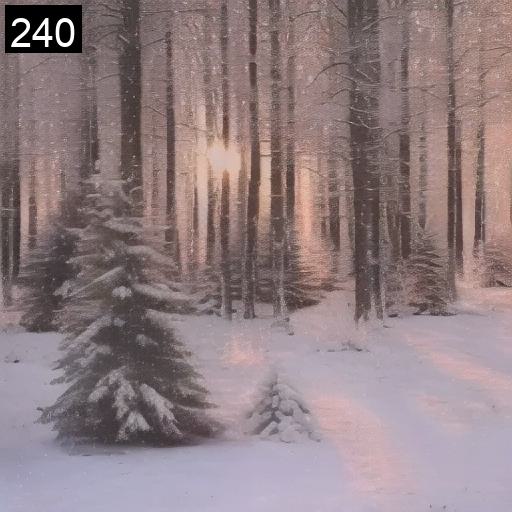} &
            \includegraphics[width=0.2\linewidth]{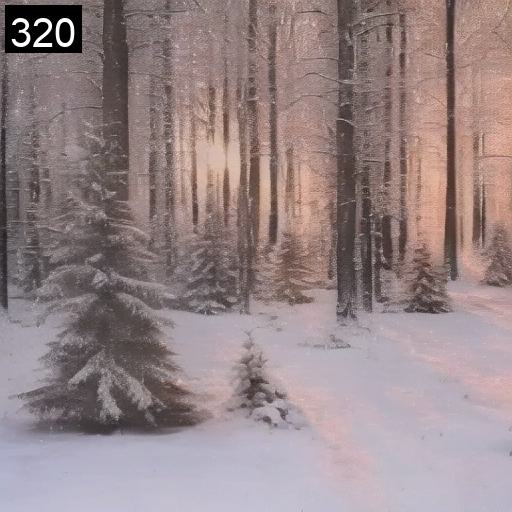} \\
            \multicolumn{5}{c}{\small (a) \textsf{"a serene winter scene in a forest. The forest is blanketed in a thick layer of snow, which~$\text{\myldots}$"}}\\
            \includegraphics[width=0.2\linewidth]{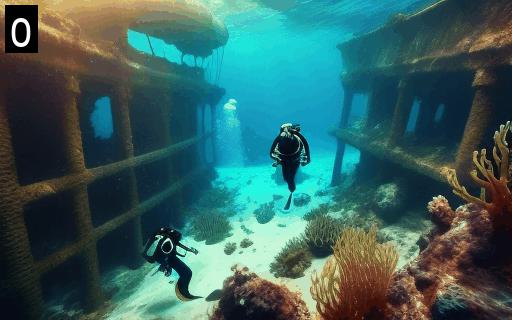} &
            \includegraphics[width=0.2\linewidth]{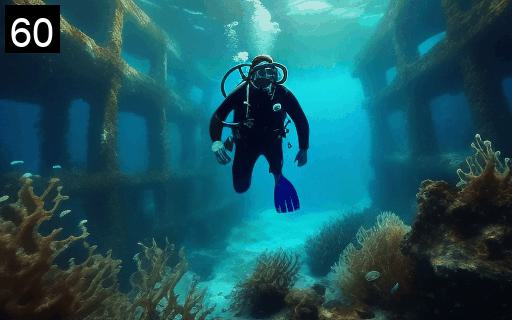} &
            \includegraphics[width=0.2\linewidth]{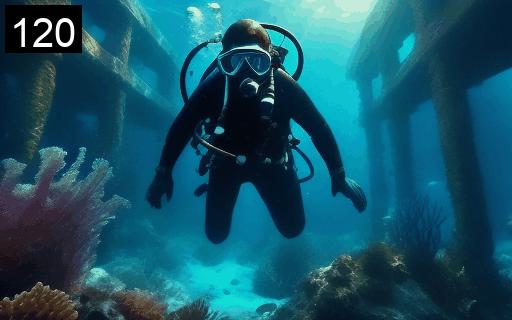} &
            \includegraphics[width=0.2\linewidth]{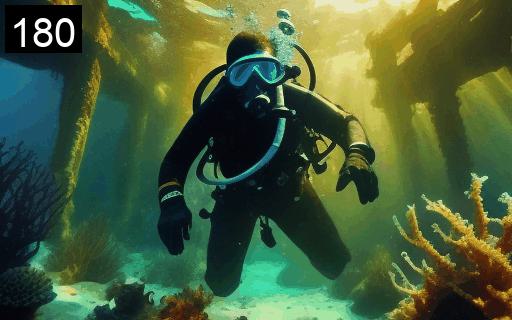} &
            \includegraphics[width=0.2\linewidth]{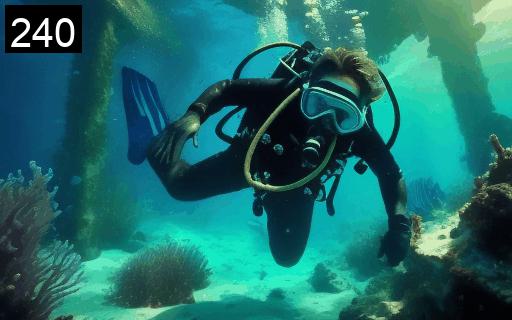} \\
            \multicolumn{5}{c}{\small (b) \textsf{"A vibrant underwater scene of a scuba diver exploring a shipwreck, 2K, photorealistic."}} \vspace{1mm}\\
            %
            \iffalse
            \includegraphics[width=0.2\linewidth]{fig/mp_jpgs/6/0.jpg} &
            \includegraphics[width=0.2\linewidth]{fig/mp_jpgs/6/4.jpg} &
            \includegraphics[width=0.2\linewidth]{fig/mp_jpgs/6/8.jpg} &
            \includegraphics[width=0.2\linewidth]{fig/mp_jpgs/6/12.jpg} &
            \includegraphics[width=0.2\linewidth]{fig/mp_jpgs/6/15.jpg} \\
            \multicolumn{5}{c}{\small (c) \textsf{"A tiger \textbf{\textit{walking}} $\rightarrow$ \textbf{\textit{standing}} $\rightarrow$  \textbf{\textit{resting}} on the grassland, photorealistic, 4k, high definition"}} \vspace{1mm}\\
            \fi
%            \iffalse
            \includegraphics[width=0.2\linewidth]{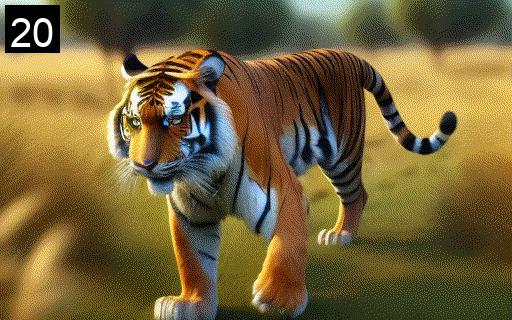} &
        \includegraphics[width=0.2\linewidth]{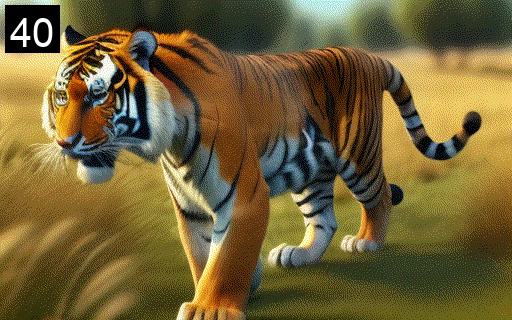} &
        \includegraphics[width=0.2\linewidth]{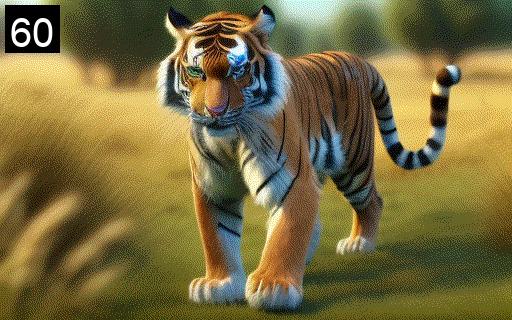} &
        \includegraphics[width=0.2\linewidth]{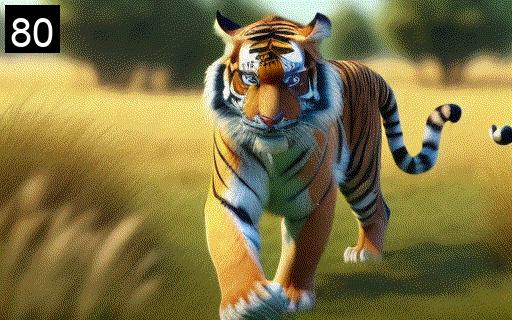} &
        \includegraphics[width=0.2\linewidth]{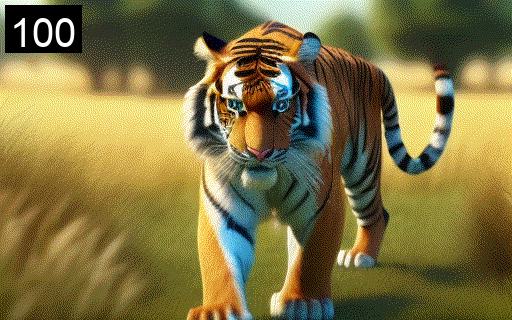} \\
        \includegraphics[width=0.2\linewidth]{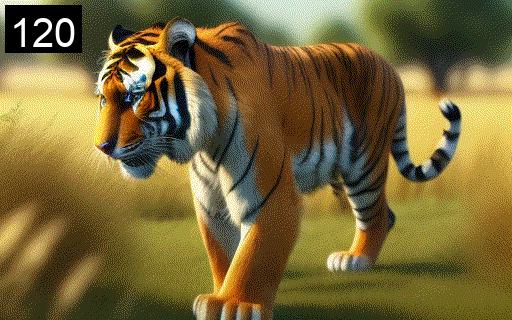} &
        \includegraphics[width=0.2\linewidth]{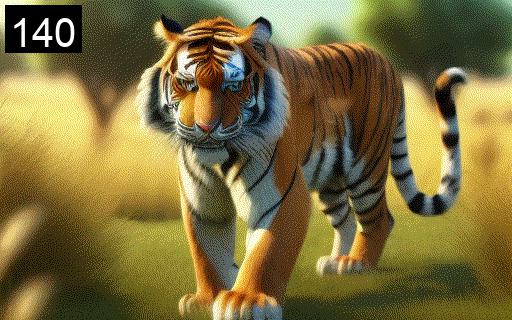} &
        \includegraphics[width=0.2\linewidth]{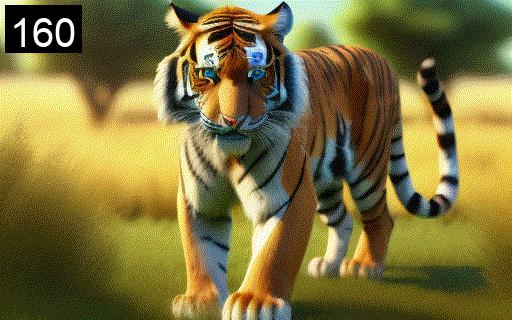} &
        \includegraphics[width=0.2\linewidth]{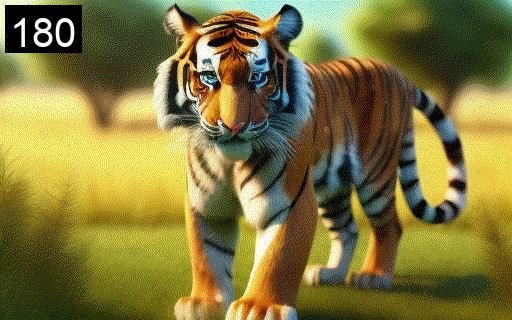} &
         \includegraphics[width=0.2\linewidth]{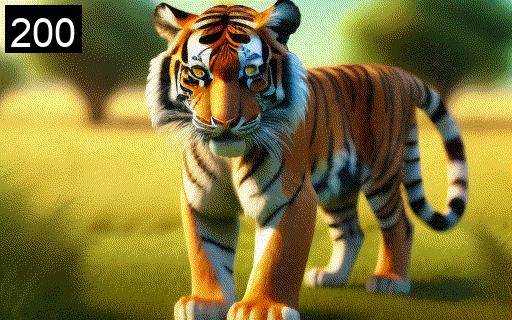} \\
        \includegraphics[width=0.2\linewidth]{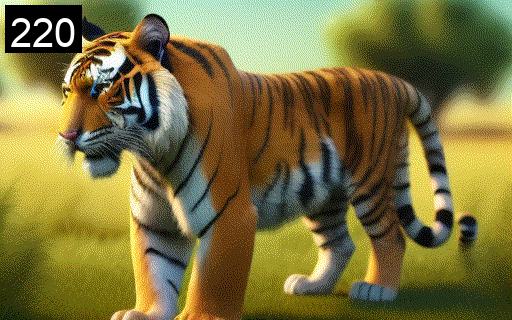} &
        \includegraphics[width=0.2\linewidth]{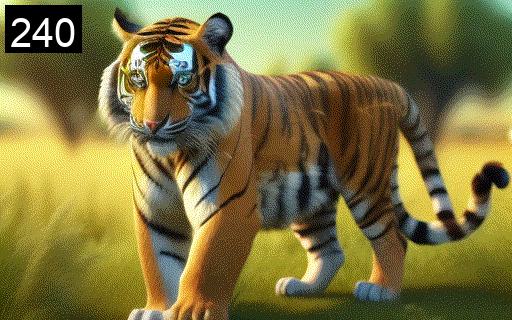} &
        \includegraphics[width=0.2\linewidth]{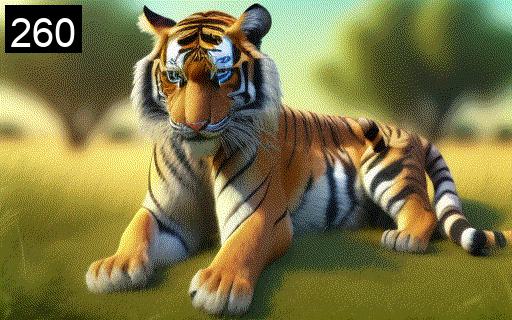} &
        \includegraphics[width=0.2\linewidth]{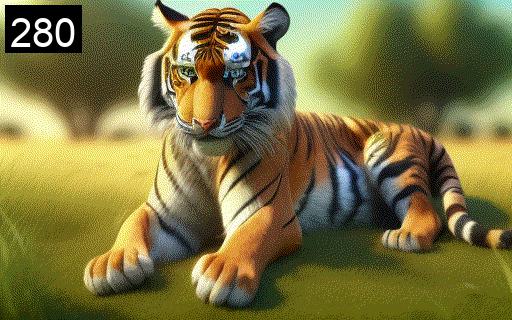} &
        \includegraphics[width=0.2\linewidth]{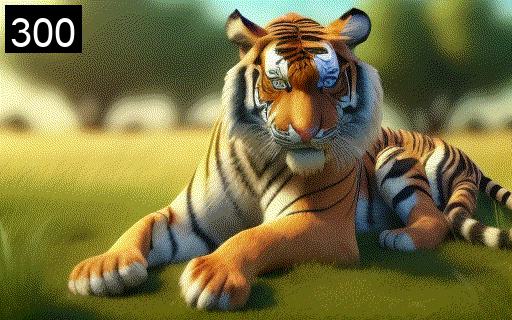} \\
         \multicolumn{5}{c}{(c) \small \textsf{"A tiger  \textbf{\textit{walking}} $\rightarrow$  \textbf{\textit{standing}} $\rightarrow$  \textbf{\textit{resting}} on the grassland, photorealistic, 4k, high definition"}} \vspace{4pt} \\
%         \fi
        \end{tabular}
    }
        \caption{
            Illustrations of long videos generated by FIFO-Diffusion based on (a) Open-Sora Plan and (b) VideoCrafter2, as well as (c) multiple prompts based on VideoCrafter2.
            The number on the top-left corner of each frame indicates the frame index.
        } \label{fig:qual_short}
%       \vspace{-4mm}
\end{figure}

\begin{figure}[!h]
    \centering
    \renewcommand{\arraystretch}{0.7}
    \scalebox{0.98}{
    \setlength{\tabcolsep}{1pt}
    \hspace{-3mm}
    \begin{tabular}{cccccc}
    \rotatebox[origin=c]{90}{\small Ours \hspace{-16mm}} &
        \includegraphics[width=0.2\linewidth]{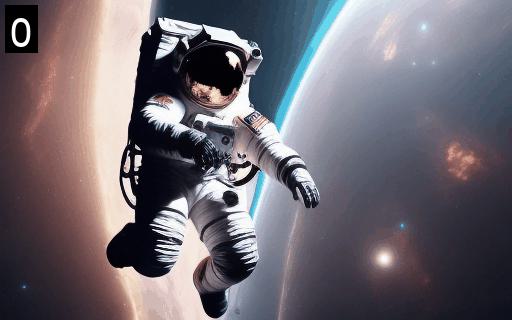} &
        \includegraphics[width=0.2\linewidth]{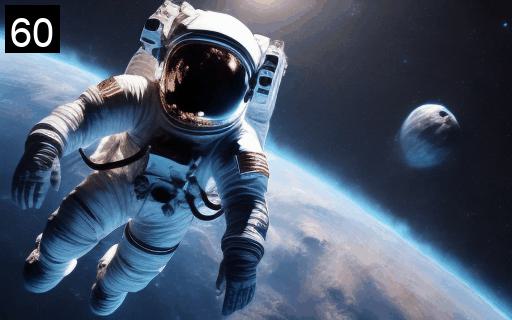} &
        \includegraphics[width=0.2\linewidth]{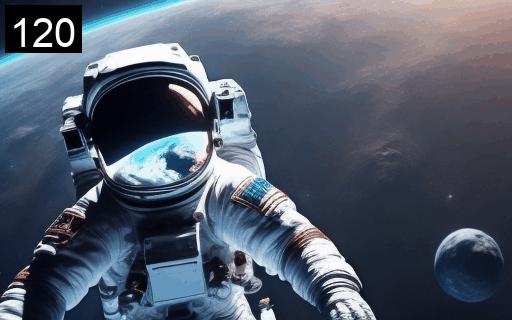} &
        \includegraphics[width=0.2\linewidth]{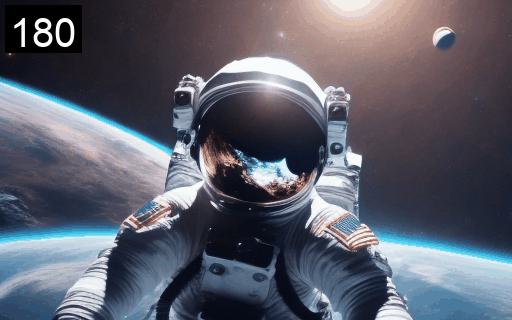} &
        \includegraphics[width=0.2\linewidth]{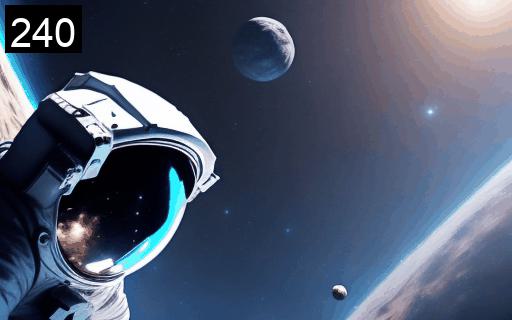} \vspace{1mm} \\
        \rotatebox[origin=c]{90}{\small FreeNoise \hspace{-16mm}} &
        \includegraphics[width=0.2\linewidth]{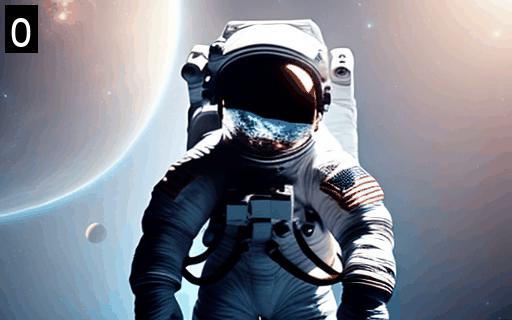} &
        \includegraphics[width=0.2\linewidth]{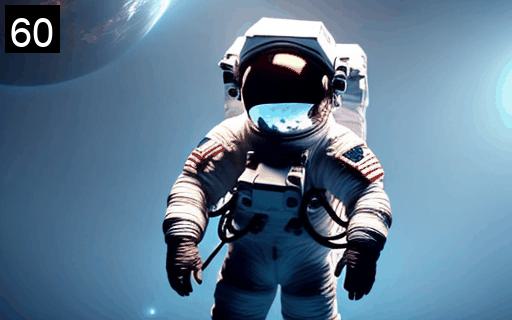} &
        \includegraphics[width=0.2\linewidth]{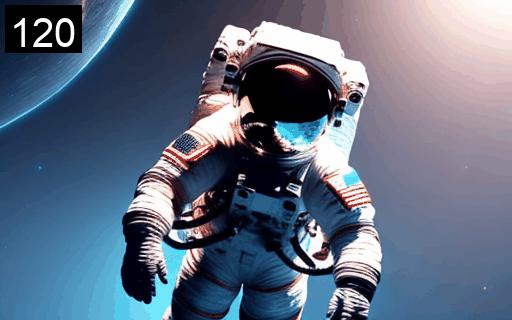} &
        \includegraphics[width=0.2\linewidth]{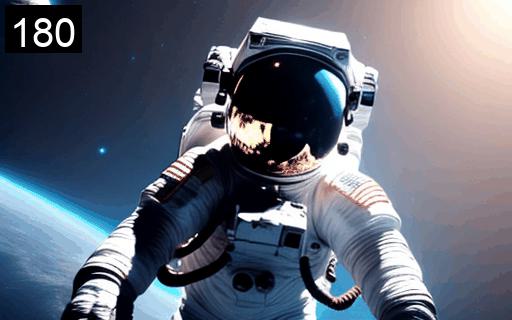} &
        \includegraphics[width=0.2\linewidth]{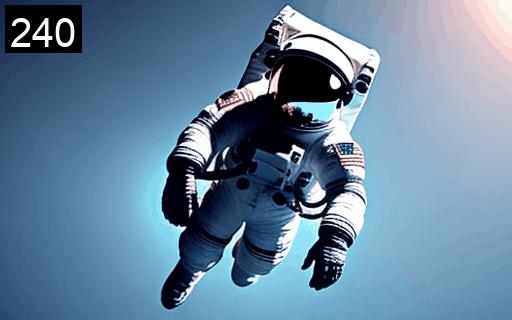} \vspace{1mm}\\
        \rotatebox[origin=c]{90}{\small Gen-L-Video \hspace{-16mm}} &
        \includegraphics[width=0.2\linewidth]{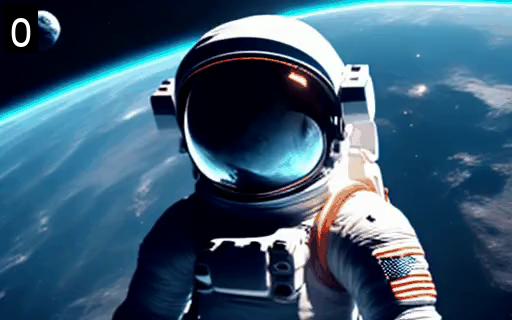} &
        \includegraphics[width=0.2\linewidth]{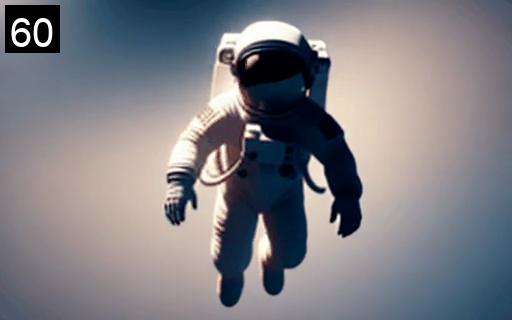} &
        \includegraphics[width=0.2\linewidth]{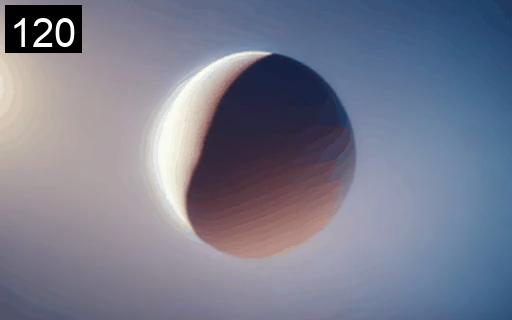} &
        \includegraphics[width=0.2\linewidth]{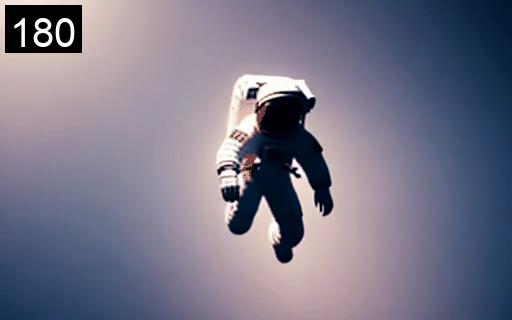} &
        \includegraphics[width=0.2\linewidth]{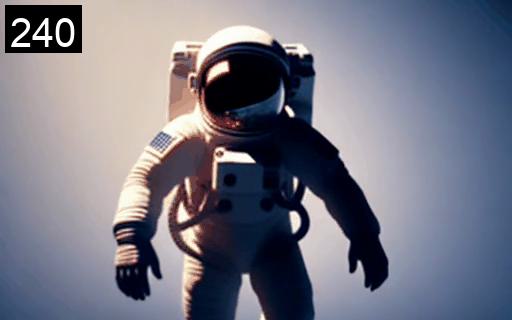} \vspace{1mm} \\
        \rotatebox[origin=c]{90}{\small LaVie+SEINE \hspace{-16mm}} &
        \includegraphics[width=0.2\linewidth]{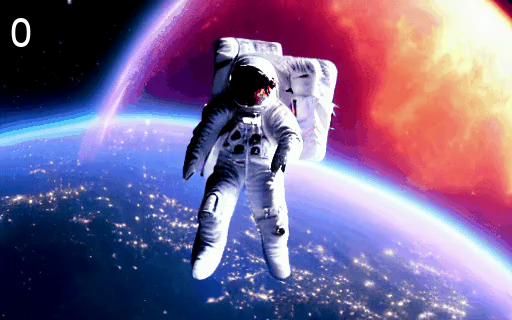} &
        \includegraphics[width=0.2\linewidth]{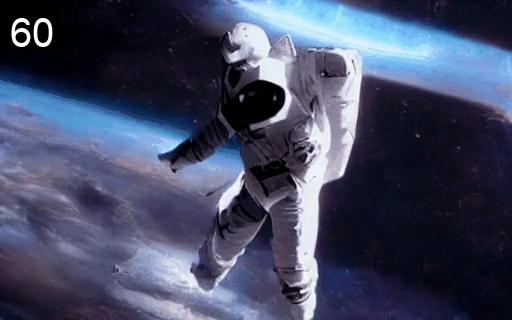} &
        \includegraphics[width=0.2\linewidth]{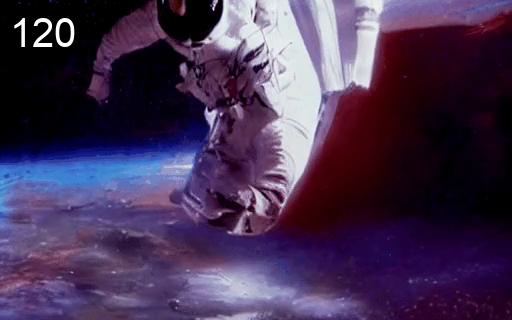} &
        \includegraphics[width=0.2\linewidth]{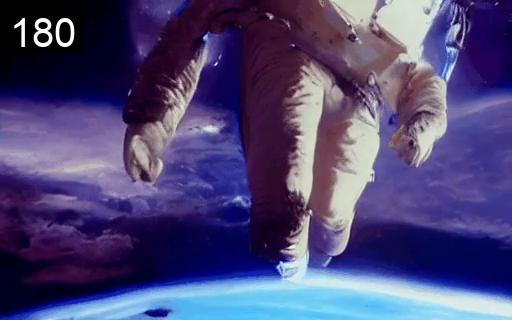} &
        \includegraphics[width=0.2\linewidth]{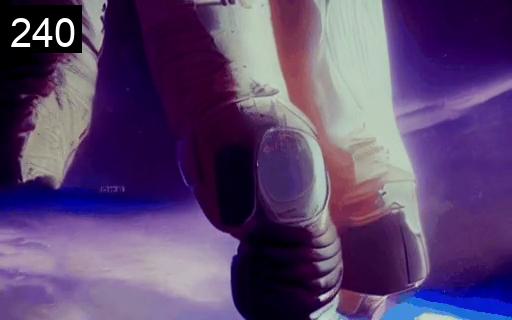} \vspace{1mm} \\
        & \multicolumn{5}{c}{\small \textsf{"An astronaut floating in space, high quality, 4K resolution."}} \vspace{3pt}\\
    
    \end{tabular}
    }
    \caption{
        Sample videos generated by 
        (first) FIFO-DIffusion on VideoCrafter2, 
        (second) FreeNoise on VideoCrafter2,
        (third) Gen-L-Video on VideoCrafter2, and 
        (last) LaVie + SEINE.
        The number on the top-left corner of each frame indicates the frame index.
    }\label{fig:qual_comparison}
    \vspace{-3mm}
\end{figure} 

We first evaluate the performance of the proposed approach qualitatively.
\cref{fig:qual_long} illustrates examples of long videos~(longer than 10K frames) generated by FIFO-Diffusion based on VideoCrafter2.
It demonstrates the ability of FIFO-Diffusion to generate significantly longer videos than the target length of pretrained baseline models---16 frames in this case.
The individual frames exhibit outstanding visual quality with no perceptual quality degradation even in the later part of the videos while preserving semantic information across all frames.
\cref{fig:qual_short} (a) and (b) present the generated videos with natural motion of scenes and cameras; the consistency of motion is effectively maintained by referencing earlier frames through the generation process.

Furthermore, \cref{fig:qual_short} (c) illustrates that FIFO-Diffusion can generate videos with extensive motion driven by a sequence of changing prompts. 
The capability to generate multiple motions and seamless transitions between scenes highlight the practicality of our method.
Please refer to \cref{app:qual,app:mp} for more examples and our project page\footref{fn:project} for video demos, in comparisons with the videos from other baselines. 

\begin{wrapfigure}{r}{0.5\linewidth}
    \centering
%    \vspace{-4mm}
    \includegraphics[width=\linewidth]{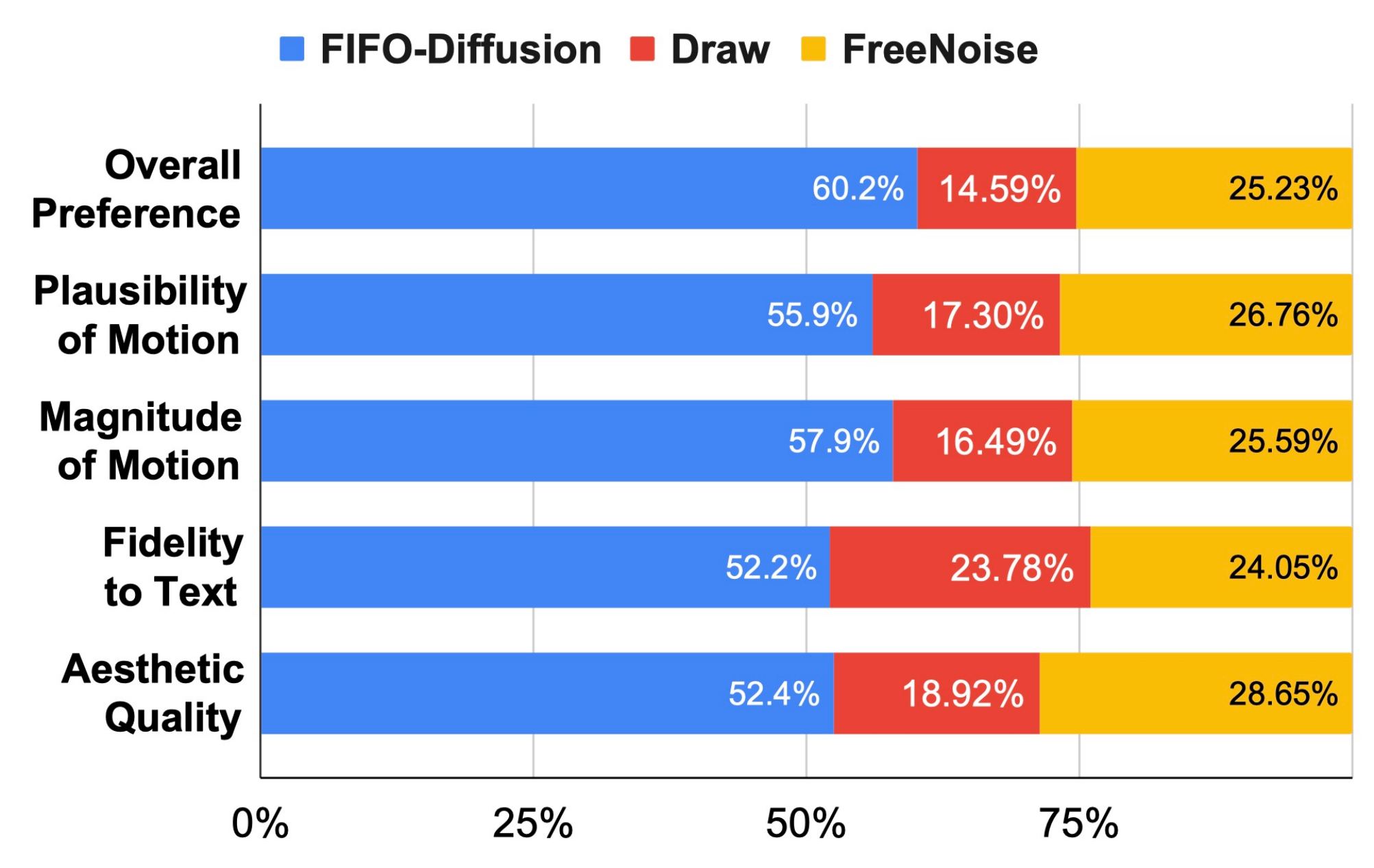} 
    \caption{The results of user study between FIFO-Diffusion and FreeNoise for five criteria.}\label{fig:user_study}
    \vspace{-2mm}
\end{wrapfigure}
\cref{fig:qual_comparison} compares the results from FIFO-Diffusion with two training-free techniques, FreeNoise~\citep{qiu2023freenoise} and Gen-L-Video~\citep{wang2023genl} based on VideoCrafter2, as well as a training-based chunked autoregressive method, LaVie~\citep{wang2023lavie} + SEINE~\citep{chen2023seine}.
Note that the chunked autoregressive method requires two models: LaVie for T2V and SEINE for I2V.
We observe that our method significantly outperforms the others in terms of motion smoothness, frame quality, and scene diversity.

Among the training-free methods, Gen-L-Video often produces videos with blurred background while FreeNoise struggles to generate dynamic scenes.\footnote{We provide qauantitative evaluation on the magnitude of motion in \cref{app:motion}.}
The videos from LaVie + SEINE gradually degrade and diverge from text prompts due to error accumulation in their autoregressive generation processes.
Additionally, they often exhibit discontinuities between adjacent chunks, as only the last frame of each chunk is employed to transfer contextual information to the next.
\cref{fig:qual_comparison_app1,fig:qual_comparison_app2} in \cref{app:qual_comparison} provide further examples comparing these methods.

We also conduct a user study to evaluate the long video generation performance of FIFO-Diffusion compared to an existing approach, FreeNoise. 
As shown in \cref{fig:user_study}, users expressed a strong preference for FIFO-Diffusion across all criteria, particularly those related to motion. 
Given that motion is one of the most defining characteristics of videos as opposed to images, the strong performance of FIFO-Diffusion in these criteria is promising and highlights its potential to generate more natural, dynamic videos. 
Details about the user study are provided in \cref{app:user_study_details}.

\begin{table}[t]
    \caption{Comparisons of $\text{FVD}_{128}$ and IS scores on UCF-101. FIFO-Diffusion with latent partitioning and lookahead denoising utilizes Latte~\cite{ma2024latte} as its baseline, where the number of partitions is four ($n=4$).
    The FVD and IS scores of the other algorithms are obtained from their respective papers, and PVDM~\citep{yu2023pvdm} denotes PVDM-L~(400-400s).\vspace{2mm}}
    \centering
    \renewcommand{\arraystretch}{1}
    \setlength{\tabcolsep}{8mm}
    \scalebox{0.8}{
    \begin{tabular}{ccccccc}
        \toprule
        & $\text{FVD}_{128}~(\downarrow)$ & IS~$(\uparrow)$ \\
                \hline
                StyleGAN-V~\citep{skorokhodov2022styleganv} & 1773.4 & 23.94{$\pm$0.73} \\
                VIDM~\citep{mei2022vidm} & 1531.9 & -- \\
                PVDM~\citep{yu2023pvdm} & \ \ 648.4 & 74.40{$\pm$1.25} \\
                %\hline
                %Latte~\citep{ma2024latte} & -- & 73.31 \quad\quad \ \   \\
                %Latte~(64 steps) & -- & 69.42{$\pm$3.90}\\
                \hline
                FIFO-Diffusion (ours) & \ \  \textbf{596.64} & \textbf{74.44}{\textbf{$\pm$1.17}} \\
        \bottomrule
    \end{tabular}
    }
    \label{tab:fvd128}
%    \vspace{-4mm}
\end{table}

\vspace{4mm}
\subsection{Quantitative results}
\label{subsec:auto_eval}

We compare FIFO-Diffusion with the baselines trained for long video generation~\citep{skorokhodov2022styleganv,mei2022vidm,yu2023pvdm} in terms of the $\text{FVD}_{128}$ and IS scores.
As shown in~\cref{tab:fvd128}, our approach outperforms all the compared methods including PVDM-L (400-400s)~\citep{yu2023pvdm}, which employs a chunked autoregressive generation strategy.
Note that PVDM-L (400-400s) iteratively generates 16 frames conditioned on the previous outputs over 400 diffusion steps while our approach only requires 64 inference steps (with lookahead denoising) without need for additional training.

\begin{table}[t]
    \centering
    \caption{Memory usages and inference times of long video generation methods. FIFO-Diffusion utilizes latent partitioning with $n=4$ and lookahead denoising.}
    \vspace{2mm}
    \label{tab:memory_time}
    \renewcommand{\arraystretch}{1.0}
    \setlength{\tabcolsep}{2.5mm}
    \scalebox{0.8}{
        \begin{tabular}{lrccccccc}
            \toprule
            & & \multicolumn{3}{c}{Memory usage~[MB]~($\downarrow$)} & & Inference time~[s/frame]~($\downarrow$) & \\
            \cmidrule{3-5}
            Method  & & 128 frames & 256 frames & 512 frames & & \\
            \hline
            FreeNoise~\citep{qiu2023freenoise} & & 26,163 & 44,683 & out of memory & & \ \ 6.09 \\
            Gen-L-Video~\citep{wang2023genl} & & 10,913 & 10,937 & 10,965 & & 22.07 \\
            \hline
            %FIFO-Diffusion~(1 GPU) & & 11245 & 11245 & 11245 & 6.20 \\
            %FIFO-Diffusion~(4 GPUs) & & 12210 & 12210 & 12210 & 1.62 \\
            FIFO-Diffusion~(1 GPU) & & 11,245 & 11,245 & 11,245 & & 12.37 \\
            FIFO-Diffusion~(8 GPUs) & & 13,496 & 13,496 & 13,496 & & \ \ 1.84 \\
            \bottomrule
    \end{tabular}}
\end{table}

\subsection{Computational cost}
\label{subsec:memory_time}

To evaluate computational efficiency, we assess memory usage and inference time per frame for training-free, long video generation methods. 
As shown in \cref{tab:memory_time}, FIFO-Diffusion generates videos of arbitrary lengths with a constant memory allocation, while FreeNoise requires memory proportional to the target video length. 
Although Gen-L-Video maintains nearly constant memory usage, it exhibits the slowest inference speed due to redundant computations. 
Notably, FIFO-Diffusion leverages parallel computation; while incorporating lookahead denoising increases computational demand, utilizing multiple GPUs for parallel processing significantly reduces sampling time.

\subsection{Ablation study}
\label{subsec:ablation}

\begin{figure}[t]
    \centering
    \renewcommand{\arraystretch}{0.7}
    \scalebox{0.98}{
    \setlength{\tabcolsep}{1pt}
    \hspace{-3mm}
    \begin{tabular}{cccccc}
%        \vspace{-3mm}\\
        \rotatebox[origin=c]{90}{\small DD\hspace{-14mm}} &
        \includegraphics[width=0.2\linewidth]{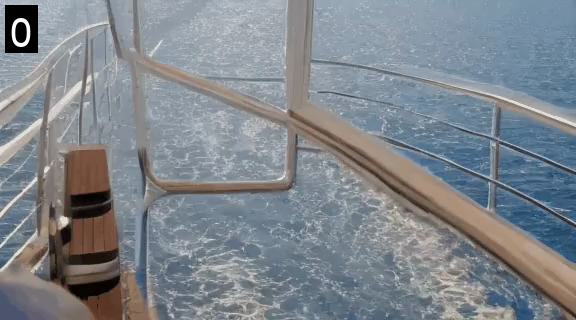} &
        \includegraphics[width=0.2\linewidth]{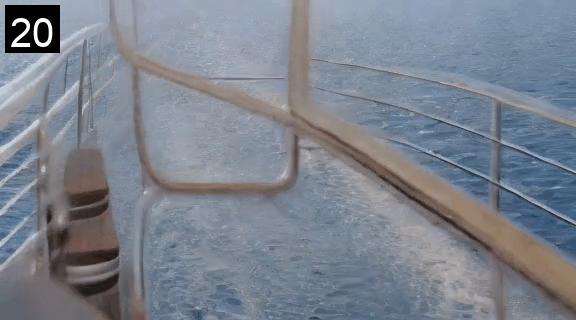} &
        \includegraphics[width=0.2\linewidth]{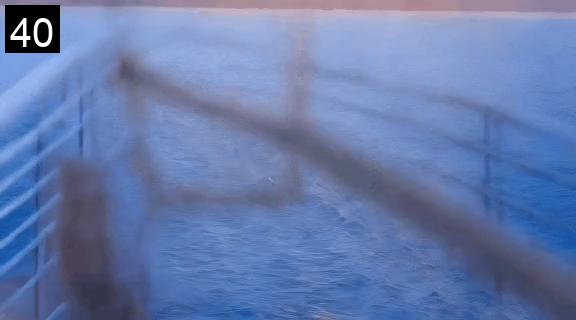} &
        \includegraphics[width=0.2\linewidth]{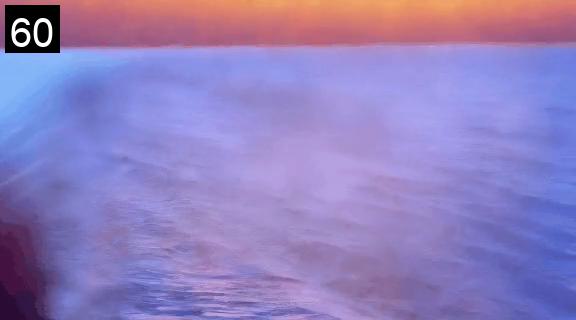} &
        \includegraphics[width=0.2\linewidth]{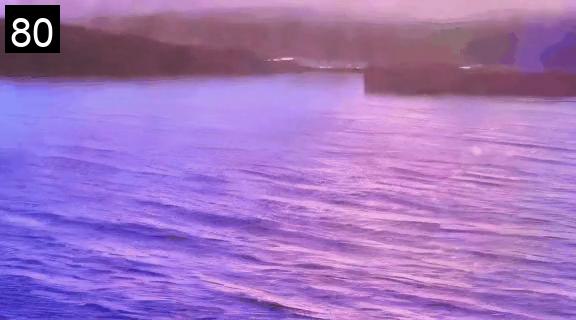} \vspace{1mm} \\
        % \shortstack[]{FreeNoise\\\vspace{8mm}} &
        \rotatebox[origin=c]{90}{\small DD+LP\hspace{-14mm}} &
        \includegraphics[width=0.2\linewidth]{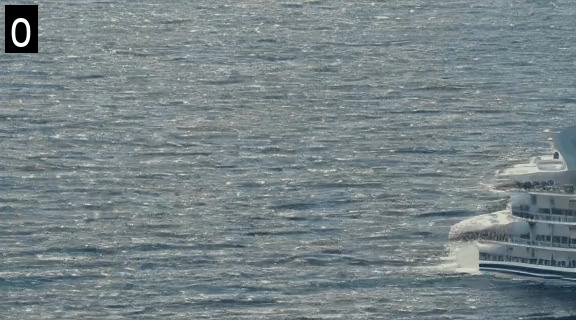} &
        \includegraphics[width=0.2\linewidth]{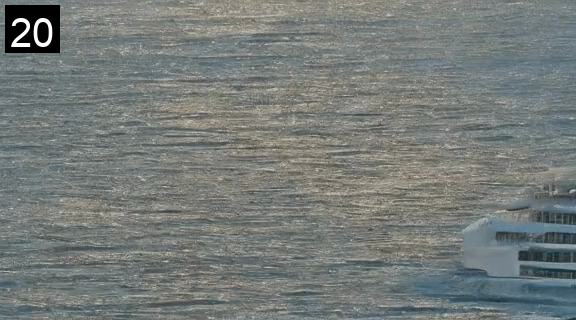} &
        \includegraphics[width=0.2\linewidth]{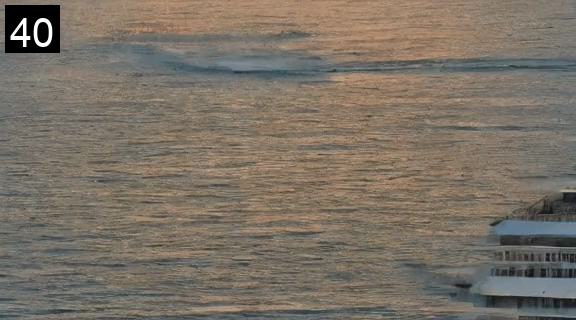} &
        \includegraphics[width=0.2\linewidth]{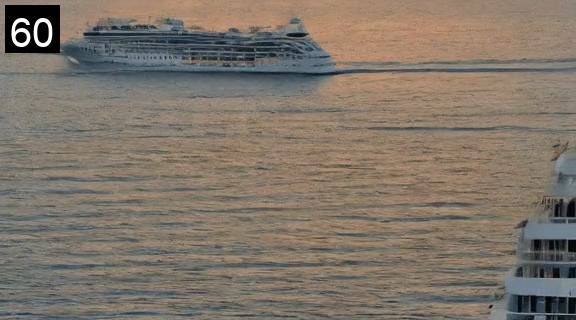} &
        \includegraphics[width=0.2\linewidth]{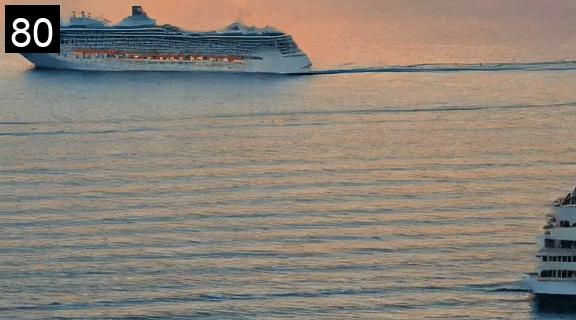} \vspace{1mm}\\
        % \shortstack[]{FIFO-\\Diffusion\vspace{8mm}} &
        \rotatebox[origin=c]{90}{\small DD+LP+LD\hspace{-14mm}} &
        \includegraphics[width=0.2\linewidth]{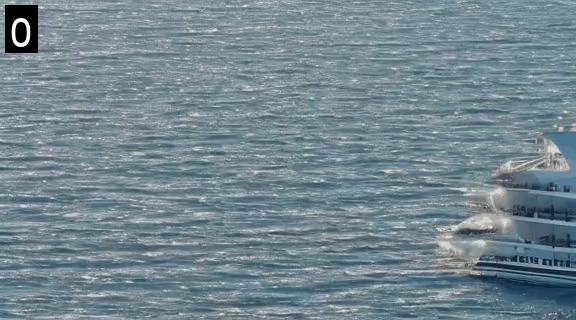} &
        \includegraphics[width=0.2\linewidth]{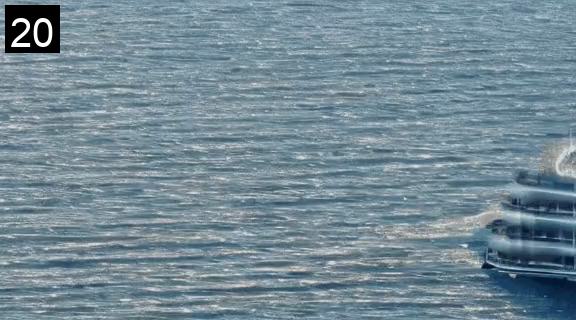} &
        \includegraphics[width=0.2\linewidth]{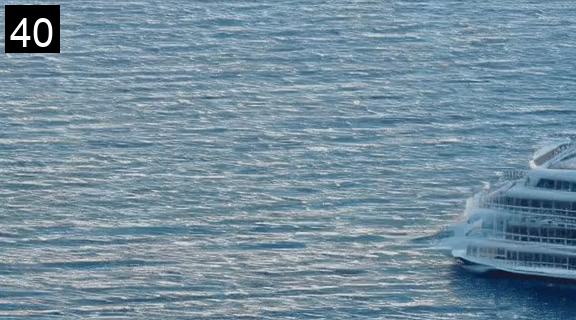} &
        \includegraphics[width=0.2\linewidth]{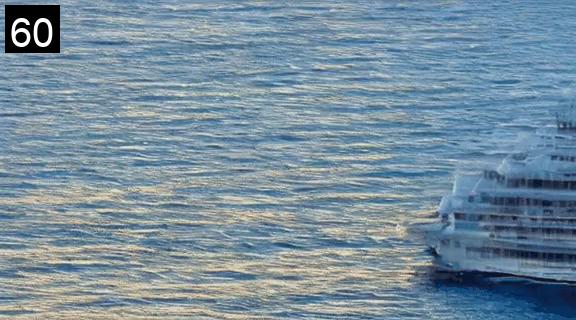} &
        \includegraphics[width=0.2\linewidth]{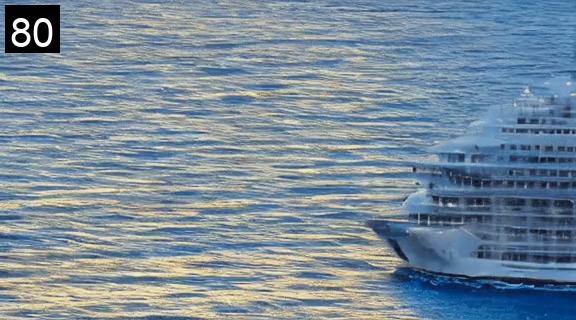} \vspace{1mm} \\
        % \shortstack[]{FIFO-\\Diffusion\vspace{8mm}} &
        & \multicolumn{5}{c}{\textsf{"A scenic cruise ship journey at sunset, ultra HD."}} \vspace{1mm} \\
    \end{tabular}
    }
    \caption{
        Ablation study.
        DD, LP, and LD signifies diagonal denoising, latent partitioning, and lookahead denoising, respectively.
        The number on the top-left corner of each frame indicates the frame index.
    }\label{fig:ablation}
%    \vspace{-3mm}
\end{figure}

We conduct ablation study to analyze the effect of latent partitioning and lookahead denoising on the performance of FIFO-Diffusion.
\cref{fig:ablation} shows that latent partitioning significantly improves both visual quality and temporal consistency of the generated videos.
Moreover, lookahead denoising further refines the quality of generated videos by facilitating temporal coherency and reducing flickering effects.
The videos on our project page\footnote{\url{https://jjihwan.github.io/projects/FIFO-Diffusion}} clearly demonstrate the benefit of FIFO-Diffusion.

Additionally, \cref{tab:ablation} compares the relative MSE loss~(see \cref{eq:mse_loss}) averaged over all time steps across different ablation settings.
The results show that latent partitioning effectively reduces the training-inference gap caused by diagonal denoising as the number of partitions increases.
Furthermore, lookahead denoising enhances the model's noise prediction accuracy, achieving low relative MSE losses (below 1.0) when used in conjunction with latent partitioning.

\begin{table}[t]
%\vspace{-5mm}
    \caption{
        Relative MSE losses of ablations.
        `LP' and `LD' denote latent partitioning and lookahead denosing, respectively.
        }
    \label{tab:ablation}
    \vspace{2mm}
    \setlength{\tabcolsep}{5mm}
    \centering
    \scalebox{0.8}{
        \begin{tabular}{cc|ccccc}
            \toprule
             & \# of partitions & without LD & with LD \\
            \hline
            without LP & 1& 1.09 & 1.01 \\
            with LP & 2 & 1.04 & 0.99 \\
            with LP & 4 & 1.02 & \textbf{0.98} \\
            \bottomrule
        \end{tabular}
        }
% \vspace{-4mm}
\end{table}

% !TEX root = ./../main.tex

\section{Related work}
\label{sec:related_work}
This section discusses existing diffusion-based generative models for videos including long video generation techniques.

\subsection{Video diffusion models}
\label{subsec:vdm}
Diffusion models, originally developed for high-quality image synthesis, have become a prominent approach in video generation~\citep{chen2023videocrafter1,ho2022video,singer2022make,zhou2022magic,wang2023modelscope}.
VDM~\citep{ho2022video} modifies the structure of U-Net~\citep{ronneberger2015unet} and proposes a 3D U-Net architecture to incorporate temporal information for denoising. 
On the other hand, Make-A-Video~\citep{singer2022make} employs a 1D temporal convolution layer following a 2D spatial convolutional layer to approximate 3D convolution.
This design enables the model to capture visual-textual relationships by training spatial layers with image-text pairs before incorporating temporal context through 1D temporal layers. 
Recently, \citep{peebles2023dit} introduce a transformer architecture,  known as DiT, for diffusion models.
Additionally, several open-sourced text-to-video models have emerged~\citep{wang2023modelscope,chen2023videocrafter1,wang2023lavie,chen2024videocrafter2}, trained on large-scale text-image and text-video datasets.

\subsection{Long video generation}
\label{subsec:long_vdm}

Long video generation approaches typically involve training models to predict future frames sequentially~\citep{voleti2022mcvd,harvey2022flexible,blattmann2023align, chen2023seine}. or generate a set of frames in a hierarchical manner~\citep{he2022latent,yin2023nuwa}.
For instance, Video LDM~\citep{blattmann2023align} and MCVD~\citep{voleti2022mcvd} employ autoregressive techniques to sequentially predict frames given several preceding ones, while FDM~\citep{harvey2022flexible} and SEINE~\citep{chen2023seine} generalize masked learning strategies for both prediction and interpolation.
Autoregressive methods are capable of producing indefinitely long videos in theory, but they often suffer from quality degradation due to error accumulation and limited temporal consistency across frames.
Alternatively, NUWA-XL~\citep{yin2023nuwa} adopts a hierarchical approach, where a global diffusion model generates sparse key frames with local diffusion models filling in frames using the key frames as references. 
However, this hierarchical setup requires batch processing, making it unsuitable for generating infinitely long videos. 

There are a few training-free long video generation techniques.
Gen-L-Video~\citep{wang2023genl} treats a video as overlapped short clips and introduces temporal co-denoising, which averages multiple predictions for one frame.
FreeNoise~\citep{qiu2023freenoise} employs window-based attention fusion to sidestep the limited attention scope issue and proposes local noise shuffle units for the initialization of long video.
FreeNoise requires memory proportional to the video length for the computation of cross, limiting its scalability for generating infinitely long videos.

\subsection{Diffusion models with latents of different noise levels}
Recent studies have adopted diffusion models for sequence generation by leveraging a sliding window approach with temporally varying noise levels~\citep{zhang2024tedi,ruhe2024rolling}.
These methods train diffusion models from scratch to accommodate latents with different noise levels, addressing tasks such as motion generation~\citep{zhang2024tedi} and video prediction~\citep{ruhe2024rolling}. 
However, training diffusion models from scratch introduces significant computational costs, especially for text-to-video generation tasks. 
In contrast, our approach is a training-free inference technique based on the standard diffusion models, trained on latents with uniform noise, for sequence generation within the sliding window framework. 
While \citep{ruhe2024rolling} is implemented with a nested loop to deal with two different axes corresponding to video frame index and diffusion time step, FIFO-Diffusion combines these two dimensions using a 1D queue, improving efficiency with a single loop.

% !TEX root = ./../main.tex

%\section{\bhr{Limitations}}
%\label{sec:limitation}

\section{Conclusion}
\label{sec:conclusion}
We introduced FIFO-Diffusion, a novel inference algorithm that enables the generation of infinitely long videos from text without tuning video diffusion models pretrained on short clips. 
Our approach achieves this by introducing diagonal denoising, which processes latents with increasing noise levels using a queue in a first-in-first-out fashion.
While diagonal denoising presents a trade-off, we addressed its limitations with latent partitioning and leveraged its strengths with lookahead denoising.
Together, these techniques allow FIFO-Diffusion to generate high-quality, long videos that maintain strong scene consistency and expressive dynamic motion.
Although latent partitioning reduces the training-inference gap of diagonal denoising, the gap persists due to changes in the model's input distribution.
However, we believe that this gap could be addressed by integrating the diagonal denoising paradigm into the training phase, and the benefits of FIFO-Diffusion remains for training as well. 
We leave this integration as future work; aligning the training and inference environments can significantly enhance FIFO-Diffusion's performance.

% We introduced FIFO-Diffusion, a novel inference algorithm that enables the generation of infinitely long videos from text without tuning video diffusion models pretrained on short clips. 
% Our approach achieves this by introducing diagonal denoising, which processes latents with increasing noise levels using a queue in a first-in-first-out fashion. 
% %At each step, a fully denoised frame is dequeued, while a new random noise frame is enqueued. 
% While diagonal denoising presents a trade-off, we addressed its limitations with latent partitioning and leveraged its strengths with lookahead denoising. 
% Together, these techniques allow FIFO-Diffusion to generate high-quality, long videos that maintain strong scene consistency and expressive dynamic motion.

% Although latent partitioning reduces the training-inference gap of diagonal denoising and lookahead denoising improves denoising accuracy, as shown in \cref{tab:ablation}, the gap persists due to changes in the model's input distribution. 
% However, we believe that the benefits of diagonal denoising are promising for training as well, and that this gap could be addressed by integrating the diagonal denoising paradigm into the training phase. We leave this integration as future work; aligning the training and inference environments can significantly enhance FIFO-Diffusion's performance.

\section*{Acknowledgements}
\label{sec:acknowledgements}
This work was partly supported by LG AI Research, and the Institute of Information \& communications Technology Planning \& Evaluation (IITP) grant funded by the Korea government (MSIT) [RS-2022-II220959 (No.2022-0-00959), (Part 2) Few-Shot Learning of Causal Inference in Vision and Language for Decision Making); NO.RS-2021- II211343, Artificial Intelligence Graduate School Program (Seoul National University)].
\bibliography{bibilography}
\bibliographystyle{plain}

\newpage
\appendix
% !TEX root = ./../main.tex

\clearpage

\section*{\Large{Appendix}}

\section{Details for \cref{lemma1,thm1}}

\subsection{Proof of \cref{lemma1}} 
\label{app:proof}   

\newtheorem*{L1}{{\normalfont\textbf{Lemma~\ref{lemma1}}}}

\begin{L1}
    If $\bm{\epsilon}(\cdot)$ is bounded, then
    \begin{align*}
        ||\gz_{t}^i - \gz_{s}^i|| = O(|t - s|) ~~\text{for any} ~~i.
    \end{align*}
\end{L1}  

\begin{proof}
    Since $\bm\epsilon(\cdot)$ is bounded, there exists some $M>0$ satisfying $||\bm\epsilon(\cdot)|| \leq M$.
    \begin{align*}
        ||\gz_{t}^i - \gz_{s}^i|| 
        &\leq ||\z_{t}^\text{vdm} - \z_{s}^\text{vdm}|| \\
        &= ||\int_s^t c\cdot \bm\epsilon(\z_{u}^\text{vdm},u\cdot\bm{1})du|| \\
        &\leq |\int_s^t c \cdot ||\bm\epsilon(\z_{u}^\text{vdm},u\cdot\bm{1})||du| \\
        &\leq c\cdot M\cdot|t-s|.
    \end{align*}
\end{proof}

\subsection{Justification on \textit{(Hypothesis 2)} of \cref{thm1}} 
\label{app:justification}
We provide justification for the hypothesis, which the diffusion model is K-Lipschitz continuous.
At inference, we can consider $z\in[0,B]^{f\times c\times h \times w}$ and $\sigma \in [\sigma_\text{min}, \sigma_\text{max}]$, where $\sigma_\text{min}>0$ since $z$ is pixel values and we inference for such $\sigma$.
In appendix B.3 of \citep{karras2022elucidating}, $\epsilon(z,\sigma)$ is given as the following:
$$\epsilon(z,\sigma)= -\sigma \cfrac{\nabla_{z} \sum_i \mathcal{N}(z;y_i,\sigma^2 \mathbf{I})}{\sum_i \mathcal{N}(z;y_i,\sigma^2 \mathbf{I})},$$
where ${ y_1, y_2, \ldots, y_n}$ are data points. Note that $\mathcal{N}(z;y_i, \sigma^2 \mathbf{I})$ is twice differentiable and continuous,
and $\sum_i \mathcal{N}(z;y_i,\sigma^2 \mathbf{I}) \geq c$ for $\exists c>0$.
Therefore, the differential function of $\epsilon(z,\sigma)$ is bounded and is Lipschitz continuous. Since $\epsilon_\theta(\cdot)$ estimates $\epsilon(\cdot)$, assuming Lipschitz continuity can be justified.

%%%%%%%%%%%%%%%%%%%%%%%%%%%%%%%%%%%%%%%%%%%%%%%%%%%%%%%%%%%%%%%%%%%%%%%%%%%%%%%%%%%%%%%%%%%%%%%%%%
\clearpage
\section{Implementation details}
\label{app:implementation_details}

We provide the implementation details of the experiments in \cref{tab:implementation}.
We use VideoCrafter1~\citep{chen2023videocrafter1}, VideoCrafter2~\citep{chen2024videocrafter2}, zeroscope\footnote{\url{https://huggingface.co/cerspense/zeroscope\_v2\_576w}}, Open-Sora Plan\footnote{\url{https://github.com/PKU-YuanGroup/Open-Sora-Plan}}, LaVie~\citep{wang2023lavie}, and SEINE~\citep{chen2023seine} as pre-trained models.
zeroscope, VideoCrafter, and Open-Sora Plan are under CC BY-NC 4.0, Apache License 2.0, and MIT License, respectively.
Except for automated results, all prompts used in experiments are randomly generated by ChatGPT-4~\citep{openai2023gpt}.
We empirically choose $n=4$ for the number of partitions in latent partitioning and lookahead denoising.
Also, stochasticity $\eta$, introduced by DDIM~\citep{song2021denoising}, is chosen to achieve good results from the baseline video generation models.

\begin{table*}[h!]
    \centering
    \caption{Implementation details regarding experiments}
    \label{tab:implementation}
        \setlength{\tabcolsep}{1mm}
        \renewcommand{\arraystretch}{1.1}
    \scalebox{0.85}{
        \begin{tabular}{ccccccccccc}
            \toprule
            Experiment & Model & $f$ & Sampling Method & $n$ & $\eta$ & \# Prompts & \# Frames & Resolution \\
            \hline
            \multirow{1}{*}{MSE loss} & \multirow{2}{*}{VideoCrafter1} & \multirow{2}{*}{16} & \multirow{2}{*}{FIFO-Diffusion} & \multirow{2}{*}{4} & \multirow{2}{*}{0.5} & \multirow{2}{*}{200} & \multirow{2}{*}{-} & \multirow{2}{*}{$320\times512$} \\
            (\cref{fig:mse_loss,tab:ablation}) \\
            \hline
            \multirow{7}{*}{Qualitative Result} 
            & zeroscope & 24 & FIFO-Diffusion & 4 & 0.5 & - & 100 & $320\times576$ \\
            & VideoCrafter1 & 16 & FIFO-Diffusion & 4 & 0.5 & - & 100 & $320\times512$ \\
            & VideoCrafter2 & 16 & FIFO-Diffusion & 4 & 1 & - & 100$\sim$10k & $320\times512$ \\ 
            & Open-Sora Plan & 17 & FIFO-Diffusion & 4 & 1 & - & 385 & $512\times512$ \\ 
            & VideoCrafter2 & 16 & FreeNoise & - & 1 & - & 100 & $320\times512$ \\ 
            & VideoCrafter2 & 16 & Gen-L-Video & - & 1 & - & 100 & $320\times512$ \\ 
            & LaVie + SEINE & 16 & chunked autoregressive & - & 1 & - & 100 & $320\times512$ \\ 
            \hline
            \multirow{2}{*}{User Study}
            & VideoCrafter2 & 16 & FIFO-Diffusion & 4 & 1 & 30 & 100 & $320\times512$ \\ 
            & LaVie & 16 & FreeNoise & - & 1 & 30 & 100 & $320\times512$ \\ 
            \hline
            \multirow{2}{*}{Motion Evaluation} 
            & VideoCrafter1 & 16 & FIFO-Diffusion & 4 & 0.5 & 512 & 100 & $256\times256$ \\
            & VideoCrafter1 & 16 & FreeNoise & - & 0.5 & 512 & 100 & $256\times256$ \\ 
            \hline
            
            Ablation study & zeroscope & 24 & FIFO-Diffusion & $\{1, 4\}$ & 0.5 & - & 100 & $320\times576$ \\
            \bottomrule 
        \end{tabular}}
\end{table*}

\subsection{Details for user study}
\label{app:user_study_details}
We randomly generated 30 prompts from ChatGPT-4 without cherry-picking, and generated a video for each prompt with 100 frames using each method.
The evaluators were asked to choose their preference (A is better, draw, or B is better) between the two videos generated by FIFO-Diffusion and FreeNoise with the same prompts, on five criteria: overall preference, plausibility of motion, magnitude of motion, fidelity to text, and aesthetic quality.
A total of 70 users submitted 111 sets of ratings, where each set consists of 20 videos from 10 prompts.
We used LaVie as the baseline for FreeNoise, since it was the latest model officially implemented at that time.

%%%%%%%%%%%%%%%%%%%%%%%%%%%%%%%%%%%%%%%%%%%%%%%%%%%%%%%%%%%%%%%%%%%%%%%%%%%%%%%%%%%%%%%%%%%%%%%%%%
\clearpage
\section{Algorithms of FIFO-Diffusion}
\label{app:algorithm}
This section illustrates pseudo-code for FIFO-Diffusion with and without latent partitioning and lookahead denoising.

\begin{algorithm}[h]
    \caption{FIFO-Diffusion with diagonal denoising~(\cref{subsec:diagonal})}
    \label{alg:fifo_algorithm}
    \begin{algorithmic}
        \Require~$N$, $f$, $\epsilon_\theta(\cdot)$, $\text{Dec}(\cdot)$, $\Phi(\cdot)$
        \State \textbf{Input:} $[\gz_{\tau_1}^1;\text{\myldots};\gz_{\tau_{f}}^f]$, $[\tau_1;\text{\myldots};\tau_f]$, $\bm{c}$
        \State \textbf{Output:} $\bm{v}$
        \State $\bm{v} \leftarrow [ ]$
        \State $\bm{\tau} \leftarrow [\tau_1;\text{\myldots};\tau_f]$ 
        \State $\Q \leftarrow [\gz_{\tau_1}^1;\text{\myldots};\gz_{\tau_f}^{f}]$ 
        \For{$i=1$~\textbf{to}~$N$}
        \State $\Q \leftarrow \Phi(\Q,\bm{\tau},\bm{c};\epsilon_\theta)$ \Comment{\cref{eq:diagonal_denoising_step}}
        \vspace{1mm}
        \State $\gz_{\tau_0}^i \leftarrow \Q\text{.dequeue}()$ \Comment{Fully denoised frame}
        \State $\bm{v}\text{.append}(\text{Dec}(\gz_{\tau_0}^i))$
        \vspace{1mm}
        \State $\gz_{\tau_f}^{i+f}\sim \mathcal{N}(\mathbf{0},\mathbf{I})$ \Comment{New random noise}
        \State $\Q\text{.enqueue}(\gz_{\tau_f}^{i+f})$ 
        \EndFor
        \State \textbf{return} $\bm{v}$
    \end{algorithmic}
\end{algorithm}

\begin{algorithm}[h]
    \caption{Initial latent construction~(\cref{subsec:diagonal})}
    \label{alg:fifo_init}
    \begin{algorithmic}
        \Require~$N$, $f$, $\epsilon_\theta(\cdot)$, $\text{Dec}(\cdot)$, $\Phi(\cdot)$
        \State \textbf{Input:} $\gz_{\tau_{f}}^{1:f}\sim\mathcal{N}(0, \bm{I}), \{\tau_i\}_{i=0}^f, \bm{c}$
        \State \textbf{Output:} $[\gz_{\tau_1}^{1};\text{\myldots};\gz_{\tau_f}^{f}]$
        \State $\bm{\tau} \leftarrow [\tau_f;\text{\myldots};\tau_f]$ 
        \State $\Q \leftarrow [\gz_{\tau_f}^{1};\text{\myldots};\gz_{\tau_f}^{f}]$ 
        \For{$i=1$~\textbf{to}~$f$}
        \State $\Q \leftarrow \Phi(\Q,\bm{\tau},\bm{c};\epsilon_\theta)$
        \vspace{1mm}
        \State $\Q\text{.dequeue}()$
        \vspace{1mm}
        \State $\gz_{\tau_f}^{i}\sim \mathcal{N}(\mathbf{0},\mathbf{I})$ \Comment{New random noise}
        \vspace{1mm}
        \State $\Q\text{.enqueue}(\gz_{\tau_f}^{i})$ 
        \State $\bm\tau \leftarrow [\overbrace{\tau_{f-i};\text{\myldots};\tau_{f-i}}^{f-i};\overbrace{\tau_{f-i+1}\text{\myldots};\tau_{f}}^{i}]$ \Comment{Varying timestep}
        \EndFor
        \State \textbf{return} $\Q=[\gz_{\tau_1}^{1};\text{\myldots};\gz_{\tau_f}^{f}]$
    \end{algorithmic}
\end{algorithm}

\begin{algorithm}[h]
    \caption{FIFO-Diffusion with latent partitioning~(\cref{subsec:latent_partitioning})}
    \label{alg:fifo_algorithm_lp}
    \begin{algorithmic} 
        \Require~$N$, $f$, $\epsilon_\theta(\cdot), \text{Dec}(\cdot), \Phi(\cdot), n$ \Comment{$n\ge 2$ if latent partitioning}
        \State \textbf{Input:} $[\gz_{\tau_1}^1;\text{\myldots};\gz_{\tau_{nf}}^{nf}]$, $[\tau_1;\text{\myldots};\tau_{nf}]$, $\bm{c}$
        \State \textbf{Output:} $\bm{v}$
        \State $\bm{v} \leftarrow [~]$
        \State $\bm{\tau} \leftarrow [\tau_{1};\text{\myldots};\tau_{nf}]$
        \State $\Q \leftarrow [\gz_{\tau_{1}}^{1};\text{\myldots};\gz_{\tau_{nf}}^{nf}]$
        \vspace{2mm}
        \For {$i=1$~\textbf{to}~$N$} 
            \For{$k=0$~\textbf{to}~$n-1$ }\Comment{Parallelizable}
                \State $\bm{\tau}_k \leftarrow \bm{\tau}^{kf+1:(k+1)f}$
                \State $\Q_k \leftarrow \Q^{kf+1:(k+1)f}$
                \vspace{2mm}
                \State $\Q_k \leftarrow \Phi(\Q_k,\bm{\tau}_k,\bm{c};\epsilon_\theta)$ \Comment{\cref{eq:lp}}
            \EndFor
            \State $Q \leftarrow [Q_0;\text{\myldots};Q_{n-1}]$
            \State $\gz_{\tau_{0}}^{i} \leftarrow \Q\text{.dequeue}()$
            \State $\bm{v}\text{.append}(\text{Dec}(\gz_{\tau_0}^i))$ 
            \vspace{1mm}
            \State $\gz_{\tau_f}^{i+nf}\sim \mathcal{N}(\mathbf{0},\mathbf{I})$
            \State $\Q\text{.enqueue}(\gz_{\tau_{nf}}^{i+nf})$
        \EndFor
        \State \textbf{return} $\bm{v}$
    \end{algorithmic}
\end{algorithm}
\vspace{-4mm}

\begin{algorithm}[h!]
    \caption{FIFO-Diffusion with lookahead denoising~(\cref{subsec:look_ahead_denoising})}
    \label{alg:fifo_algorithm_full}
    \begin{algorithmic} 
        \Require~$N, \epsilon_\theta(\cdot), \text{Dec}(\cdot), \Phi(\cdot), n$ \Comment{$n\ge 2$ if latent partitioning}
        \State \textbf{Input:} $[\gz_{\tau_1}^1;\text{\myldots};\gz_{\tau_{nf}}^{nf}]$, $[\tau_1;\text{\myldots};\tau_{nf}]$, $\bm{c}$
        \State \textbf{Output:} $\bm{v}$
        \State $\bm{v} \leftarrow [~]$
        \State $\bm{\tau} \leftarrow [\overbrace{\tau_1;\text{\myldots};\tau_1}^{f'};\tau_{1};\text{\myldots};\tau_{nf}]$
        \State $\Q \leftarrow [\overbrace{\gz_{\tau_{1}}^{1};\text{\myldots};\gz_{\tau_{1}}^{1}}^{f'};\gz_{\tau_{1}}^{1};\text{\myldots};\gz_{\tau_{nf}}^{nf}]$ \Comment{dummy latents are required}
        \vspace{2mm}
        \For{$i=1$~\textbf{to}~$N$}
            \State $\gz_{\tau_{1}}^{i} \leftarrow \Q^{f'+1}$
            \For{$k=0$~\textbf{to}~$2n-1$ } \Comment{Parallelizable}
                \State $\bm{\tau}_k \leftarrow \bm{\tau}^{kf'+1:(k+2)f'}$
                \State $\Q_k \leftarrow \Q^{kf'+1:(k+2)f'}$
                \vspace{1mm}
                \State $\Q_k^{f'+1:f} \leftarrow \Phi(\Q_k,\bm{\tau}_k,\bm{c};\epsilon_\theta)^{f'+1:f}$ \Comment{\cref{eq:ld}}
            \EndFor
            \State $\gz_{\tau_{0}}^{i} \leftarrow \Q_0^{f'+1}$
            \State $\bm{v}\text{.append}(\text{Dec}(\gz_{\tau_0}^i))$ 
            \State $\Q_0^{f'+1} \leftarrow \gz_{\tau_{1}}^{i}$
            \State $Q \leftarrow [Q_0^{1:f'};Q_0^{f'+1:f};\text{\myldots};Q_{2n-1}^{f'+1:f}]$
            \vspace{1mm}
            \State $Q \leftarrow [Q_0;Q_1^{f'+1:f};\text{\myldots};Q_{2n-1}^{f'+1:f}]$

            \State \Q\text{.dequeue}()
            \vspace{1mm}
            \State $\gz_{\tau_nf}^{i+nf}\sim \mathcal{N}(\mathbf{0},\mathbf{I})$
            \State $\Q\text{.enqueue}(\gz_{\tau_{nf}}^{i+nf})$
        \EndFor
        \State \textbf{return} $\bm{v}$
    \end{algorithmic}
\end{algorithm}

%%%%%%%%%%%%%%%%%%%%%%%%%%%%%%%%%%%%%%%%%%%%%%%%%%%%%%%%%%%%%%%%%%%%%%%%%%%%%%%%%%%%%%%%%%%%%%%%%%
\clearpage
\section{Qualitative results of FIFO-Diffusion}
\label{app:qual}
In \cref{fig:qual:vc2_0,fig:qual:vc2_1,fig:qual:vc2_2,fig:qual:vc1_0,fig:qual:zero_0,fig:qual:opensoraplan_0}, we provide more qualitative results with 4 baselines, VideoCrafter2~\citep{chen2024videocrafter2}, VideoCrafter1~\citep{chen2023videocrafter1}, zeroscope\footnote{\url{https://huggingface.co/cerspense/zeroscope\_v2\_576w}}, and Open-Sora Plan\footnote{\url{https://github.com/PKU-YuanGroup/Open-Sora-Plan}}.

\subsection{VideoCrafter2}
\label{app:qual:vc2}

\scalebox{1}{
    \setlength{\tabcolsep}{1pt}
    \begin{tabular}{ccccc}
        \multicolumn{5}{c}{} \\
        \includegraphics[width=0.2\linewidth]{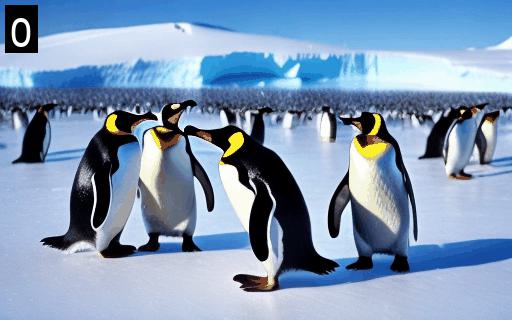} &
        \includegraphics[width=0.2\linewidth]{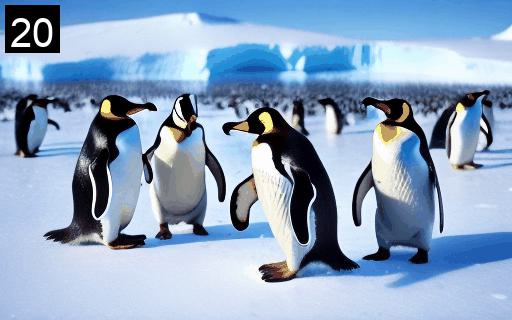} &
        \includegraphics[width=0.2\linewidth]{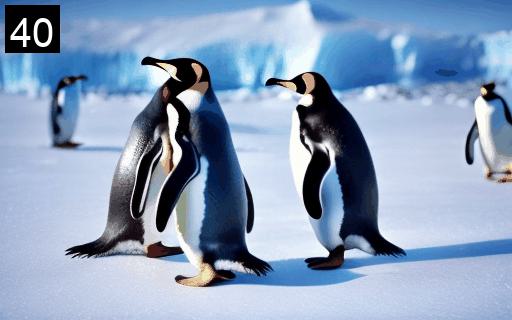} &
        \includegraphics[width=0.2\linewidth]{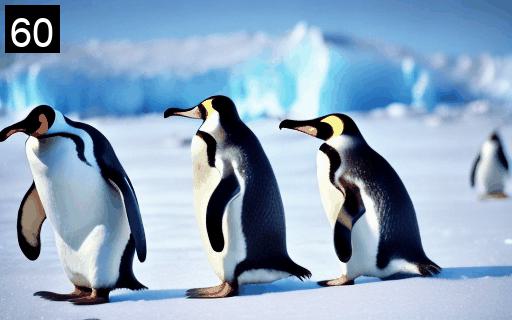} &
        \includegraphics[width=0.2\linewidth]{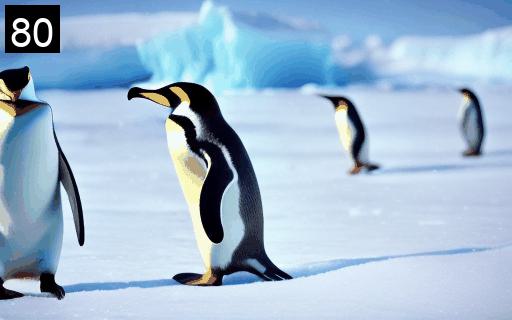} \\
        \multicolumn{5}{c}{\small (a) \textsf{"A colony of penguins waddling on an Antarctic ice sheet, 4K, ultra HD."}} \vspace{2pt} \\
        \includegraphics[width=0.2\linewidth]{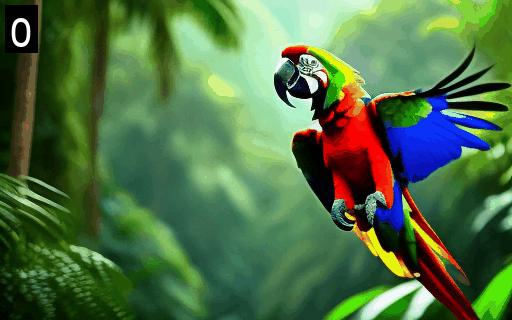} &
        \includegraphics[width=0.2\linewidth]{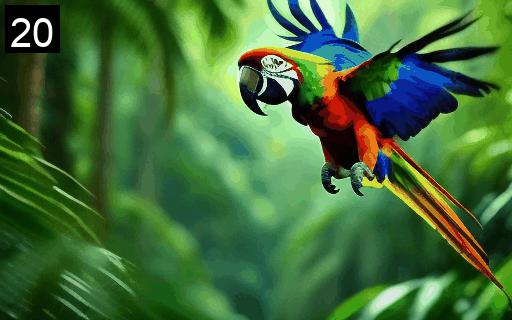} &
        \includegraphics[width=0.2\linewidth]{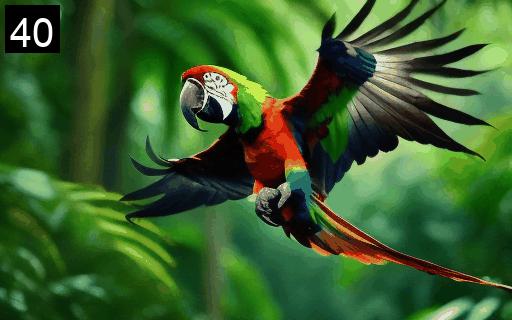} &
        \includegraphics[width=0.2\linewidth]{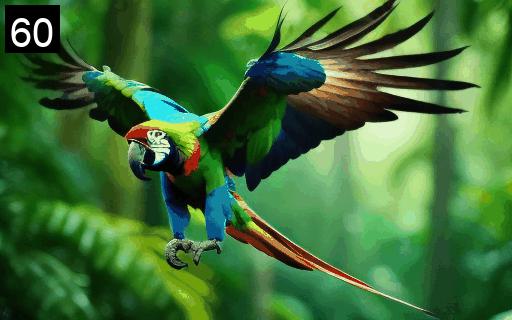} &
        \includegraphics[width=0.2\linewidth]{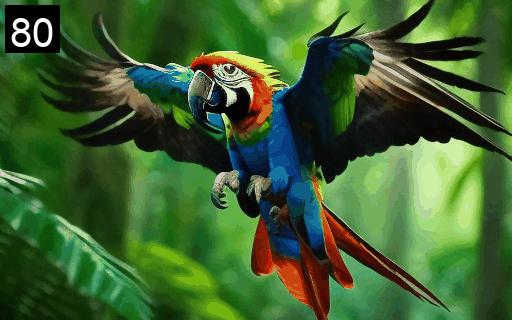} \\
        \multicolumn{5}{c}{\small (b) \textsf{"A colorful macaw flying in the rainforest, ultra HD."}}\vspace{2pt} \\
        \includegraphics[width=0.2\linewidth]{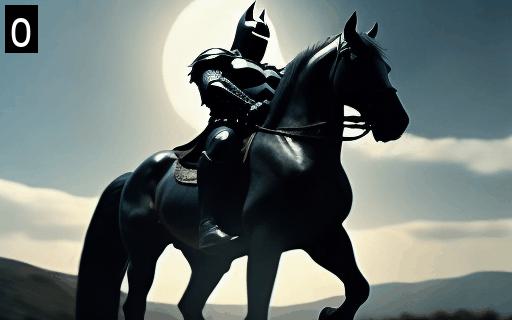} &
        \includegraphics[width=0.2\linewidth]{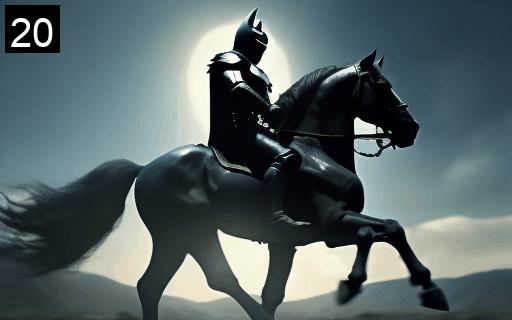} &
        \includegraphics[width=0.2\linewidth]{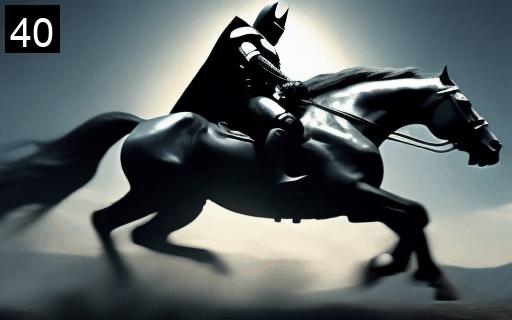} &
        \includegraphics[width=0.2\linewidth]{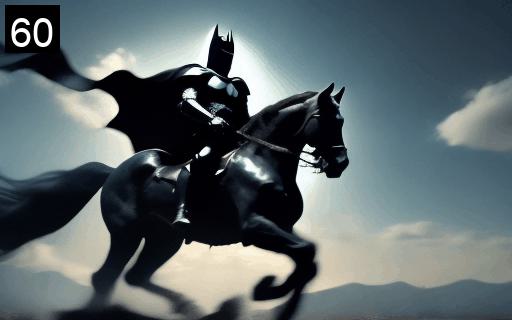} &
        \includegraphics[width=0.2\linewidth]{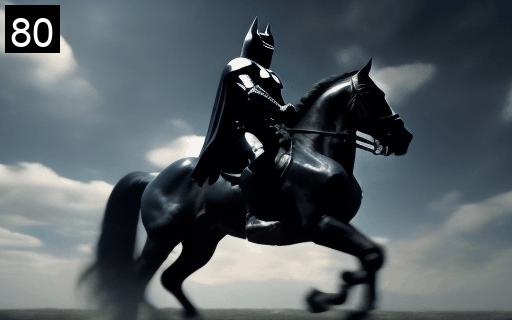} \\
        \multicolumn{5}{c}{\small (c) \textsf{"A dark knight riding on a black horse on the glassland,  photorealistic, 4k, high definition."}} \vspace{2pt}\\
        \includegraphics[width=0.2\linewidth]{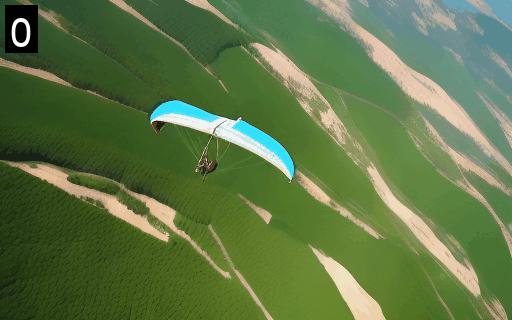} &
        \includegraphics[width=0.2\linewidth]{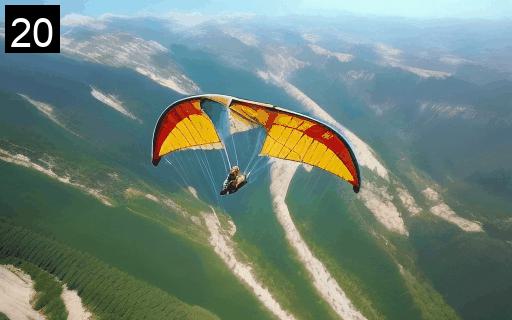} &
        \includegraphics[width=0.2\linewidth]{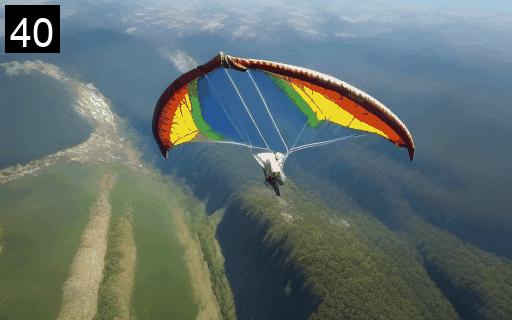} &
        \includegraphics[width=0.2\linewidth]{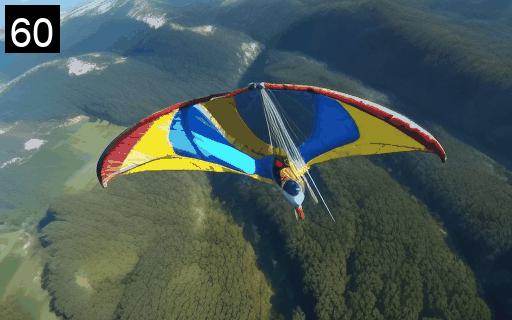} &
        \includegraphics[width=0.2\linewidth]{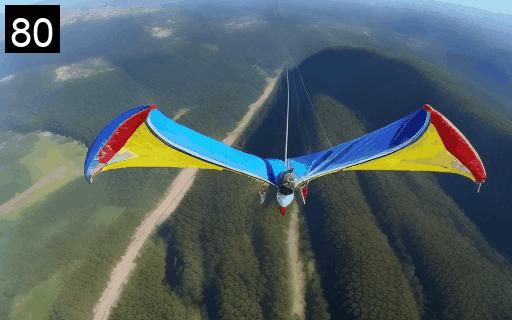} \\
        \multicolumn{5}{c}{\small (d) \textsf{"A high-altitude view of a hang glider in flight, high definition, 4K."}} \vspace{2pt}\\
        \includegraphics[width=0.2\linewidth]{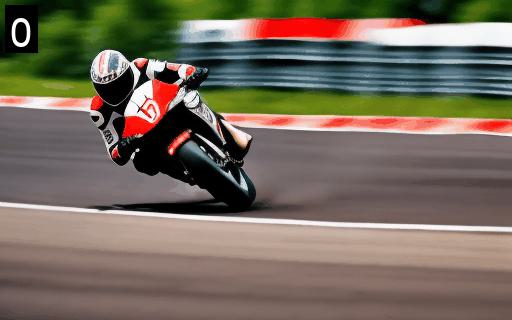} &
        \includegraphics[width=0.2\linewidth]{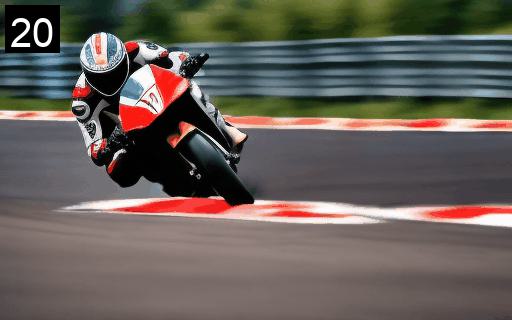} &
        \includegraphics[width=0.2\linewidth]{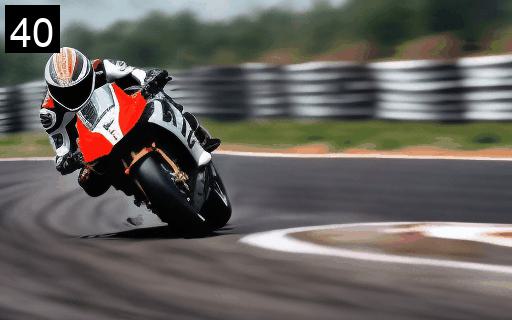} &
        \includegraphics[width=0.2\linewidth]{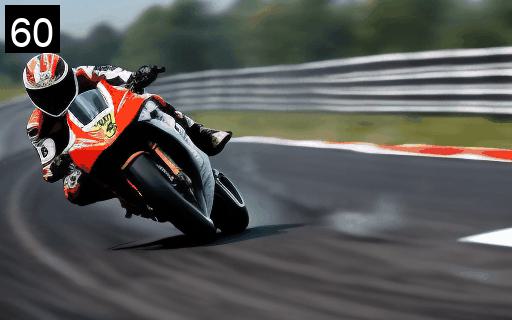} &
        \includegraphics[width=0.2\linewidth]{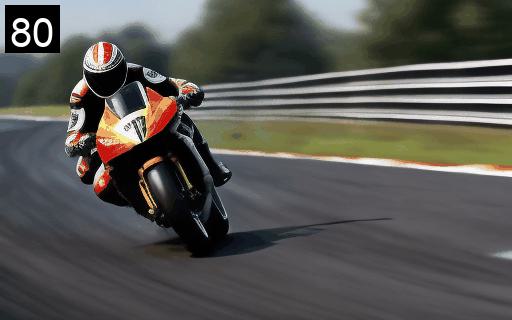} \\
        \multicolumn{5}{c}{\small (e) \textsf{"A high-speed motorcycle race on a track, ultra HD, 4K resolution."}}\vspace{2pt} \\
        \includegraphics[width=0.2\linewidth]{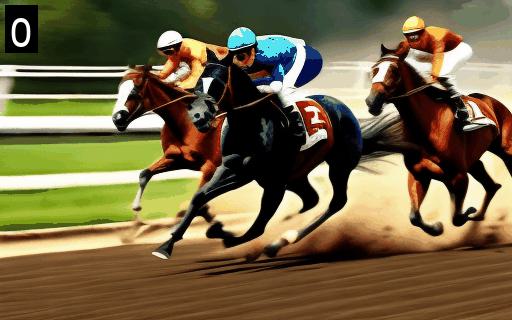} &
        \includegraphics[width=0.2\linewidth]{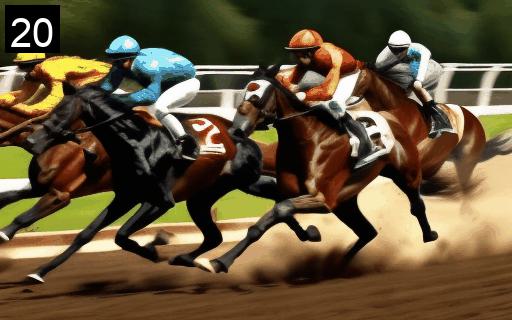} &
        \includegraphics[width=0.2\linewidth]{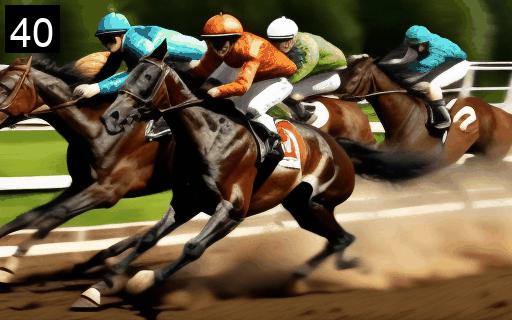} &
        \includegraphics[width=0.2\linewidth]{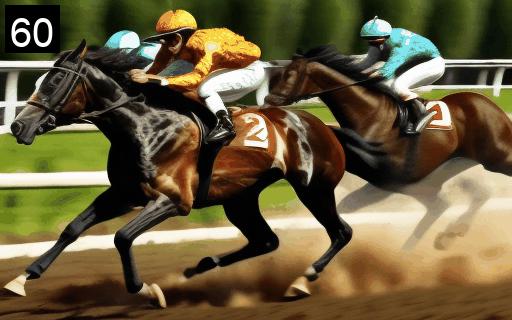} &
        \includegraphics[width=0.2\linewidth]{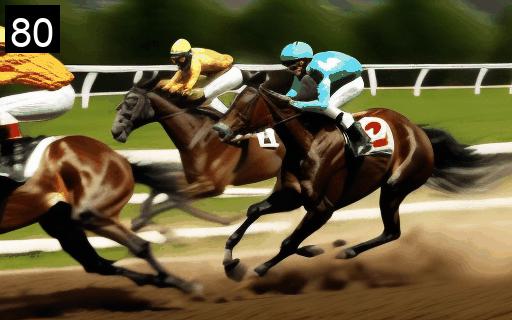} \\
        \multicolumn{5}{c}{\small (f) \textsf{"A horse race in full gallop, capturing the speed and excitement, 2K, photorealistic."}} \\

    \end{tabular}
}
\captionof{figure}{
    Videos generated by FIFO-Diffusion with VideoCrafter2.
    The number on the top left of each frame indicates the frame index.
    }
\label{fig:qual:vc2_0}

\clearpage
\scalebox{1}{
    \setlength{\tabcolsep}{1pt}
    \begin{tabular}{ccccc}
        \includegraphics[width=0.2\linewidth]{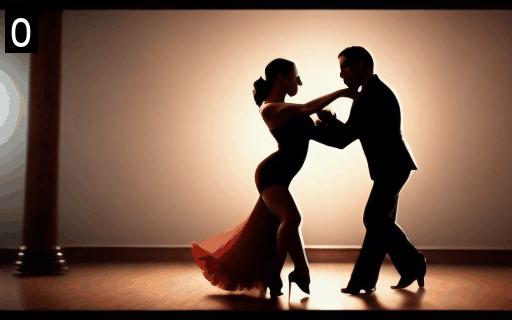} &
        \includegraphics[width=0.2\linewidth]{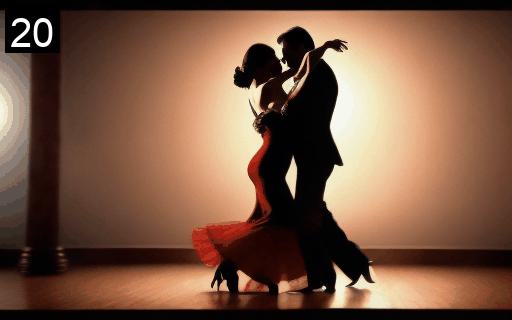} &
        \includegraphics[width=0.2\linewidth]{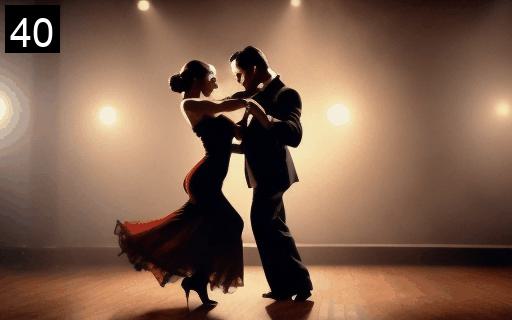} &
        \includegraphics[width=0.2\linewidth]{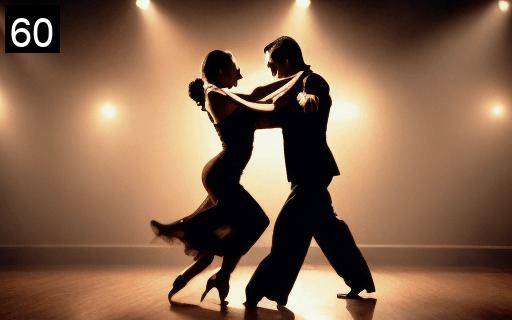} &
        \includegraphics[width=0.2\linewidth]{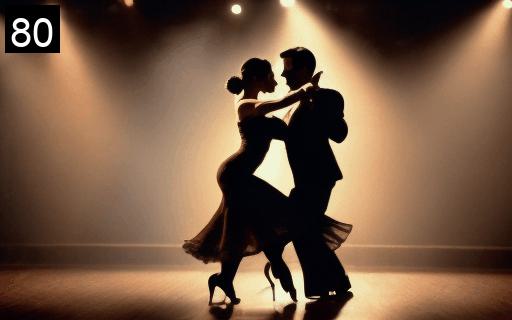} \\
        \multicolumn{5}{c}{\small (a) \textsf{"A pair of tango dancers performing in Buenos Aires, 4K, high resolution."}}\vspace{2pt} \\
        \includegraphics[width=0.2\linewidth]{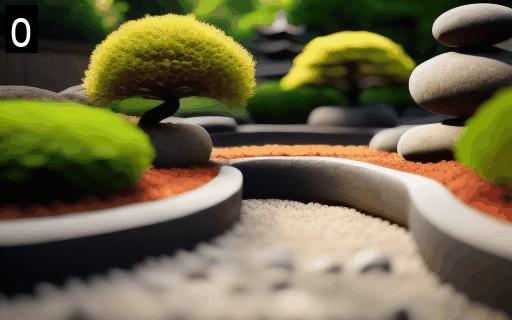} &
        \includegraphics[width=0.2\linewidth]{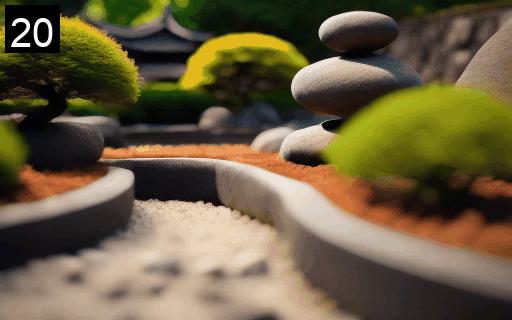} &
        \includegraphics[width=0.2\linewidth]{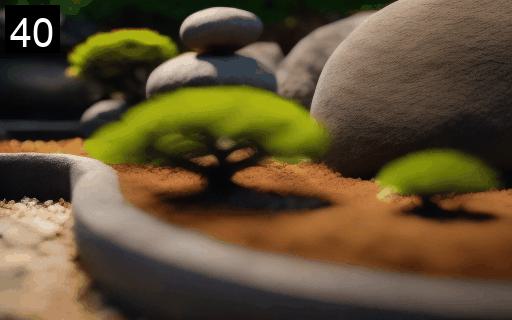} &
        \includegraphics[width=0.2\linewidth]{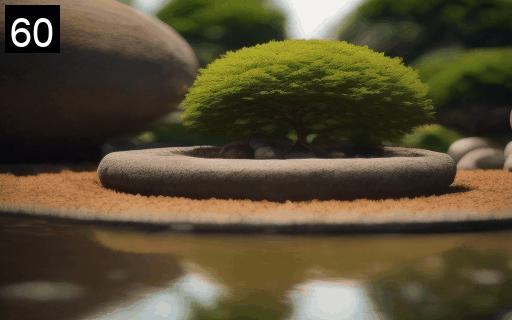} &
        \includegraphics[width=0.2\linewidth]{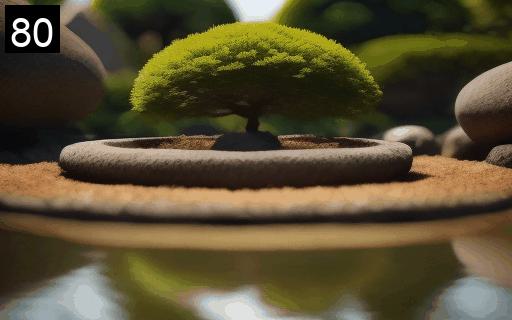} \\
        \multicolumn{5}{c}{\small (b) \textsf{"A panoramic view of a peaceful Zen garden, high-quality, 4K resolution."}}\vspace{2pt} \\
        \includegraphics[width=0.2\linewidth]{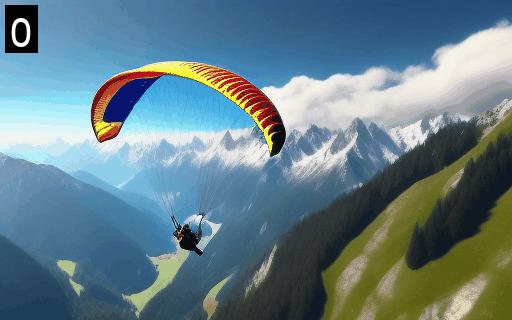} &
        \includegraphics[width=0.2\linewidth]{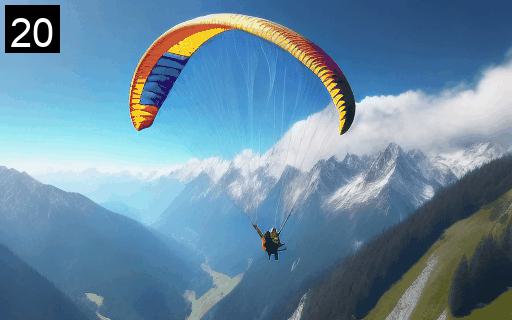} &
        \includegraphics[width=0.2\linewidth]{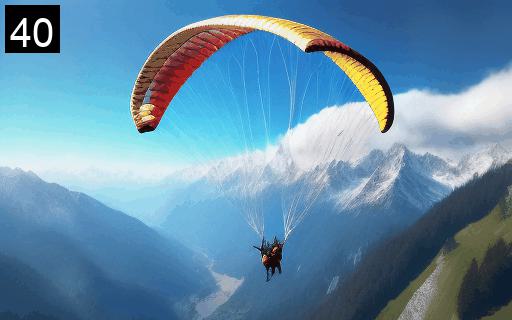} &
        \includegraphics[width=0.2\linewidth]{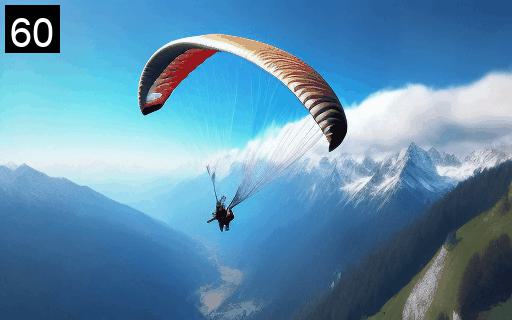} &
        \includegraphics[width=0.2\linewidth]{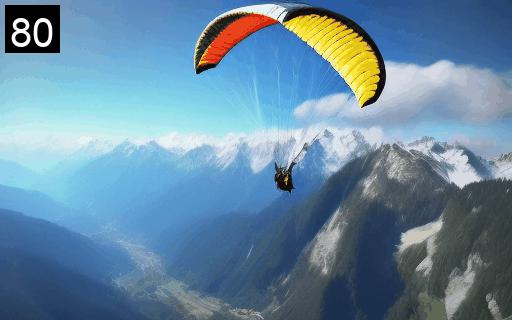} \\
        \multicolumn{5}{c}{\small (c) \textsf{"A paraglider soaring over the Alps, photorealistic, 4K, high definition."}}\vspace{2pt} \\
        \includegraphics[width=0.2\linewidth]{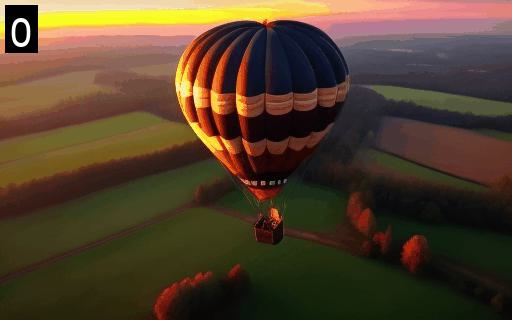} &
        \includegraphics[width=0.2\linewidth]{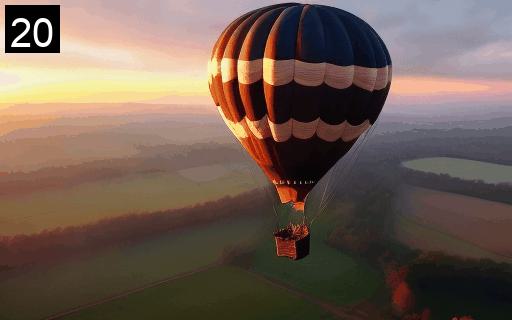} &
        \includegraphics[width=0.2\linewidth]{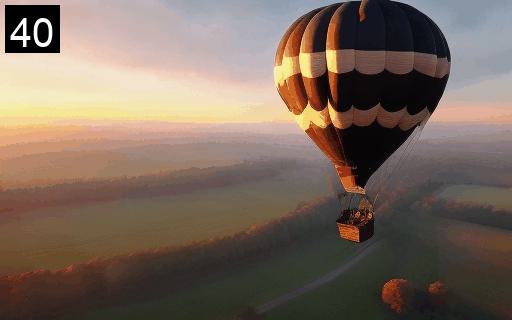} &
        \includegraphics[width=0.2\linewidth]{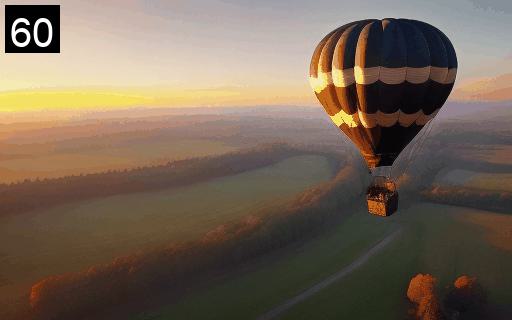} &
        \includegraphics[width=0.2\linewidth]{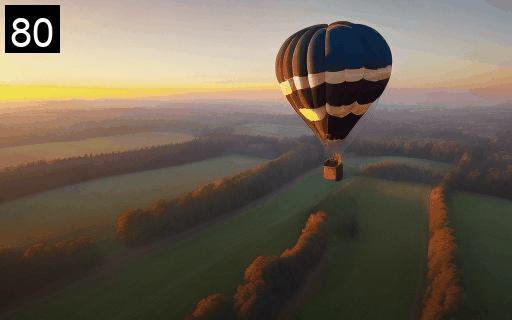} \\
        \multicolumn{5}{c}{\small (d) \textsf{"A scenic hot air balloon flight at sunrise, high quality, 4K."}}\vspace{2pt} \\
        \includegraphics[width=0.2\linewidth]{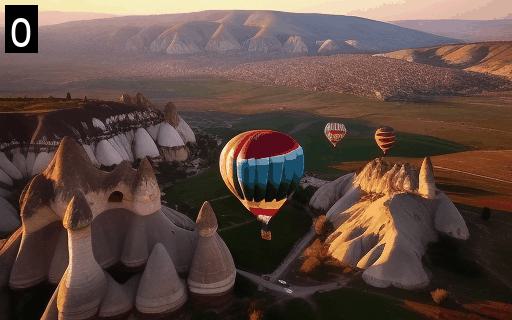} &
        \includegraphics[width=0.2\linewidth]{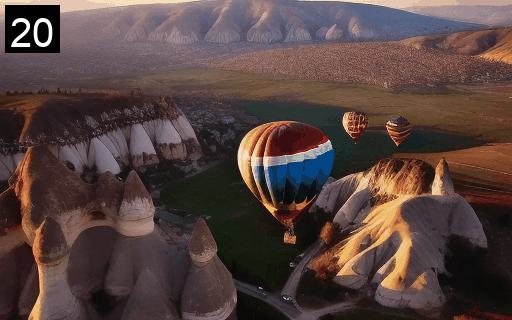} &
        \includegraphics[width=0.2\linewidth]{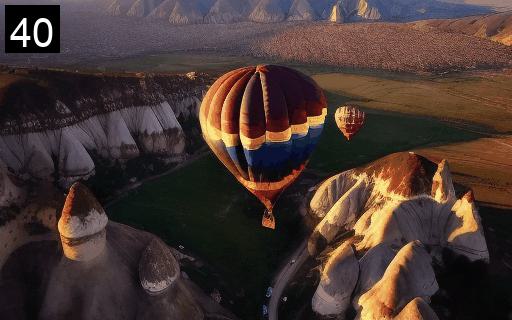} &
        \includegraphics[width=0.2\linewidth]{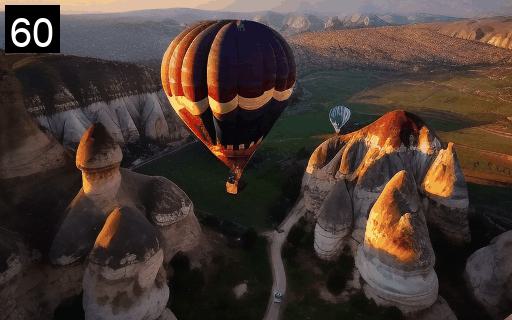} &
        \includegraphics[width=0.2\linewidth]{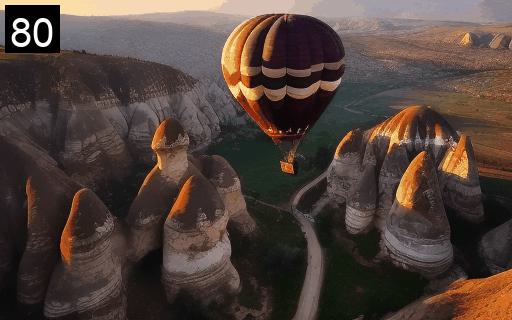} \\
        \multicolumn{5}{c}{\small (e) \textsf{"A scenic hot air balloon flight over Cappadocia, Turkey, 2K, ultra HD."}} \vspace{2pt}\\
        \includegraphics[width=0.2\linewidth]{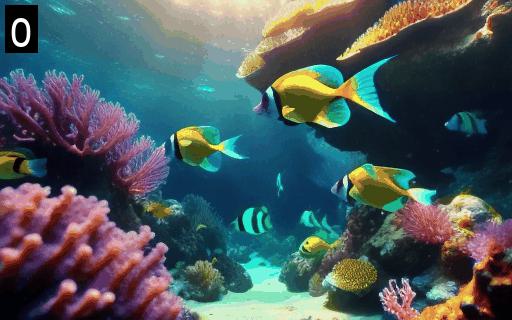} &
        \includegraphics[width=0.2\linewidth]{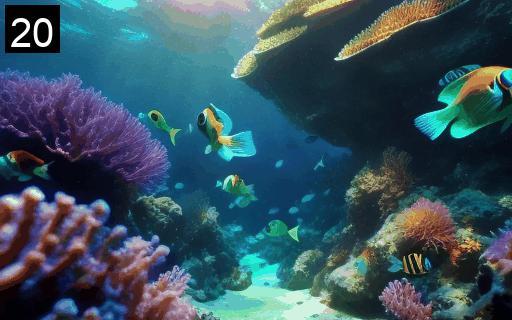} &
        \includegraphics[width=0.2\linewidth]{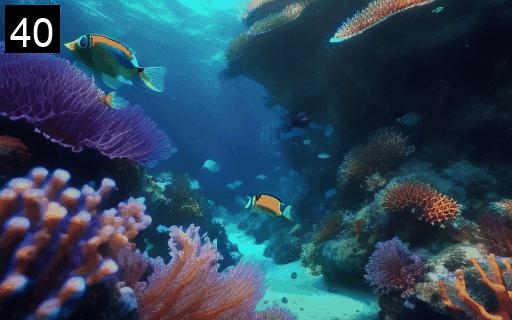} &
        \includegraphics[width=0.2\linewidth]{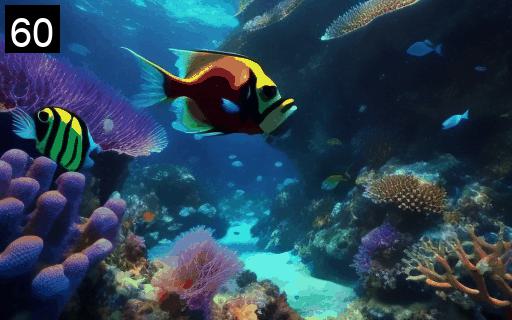} &
        \includegraphics[width=0.2\linewidth]{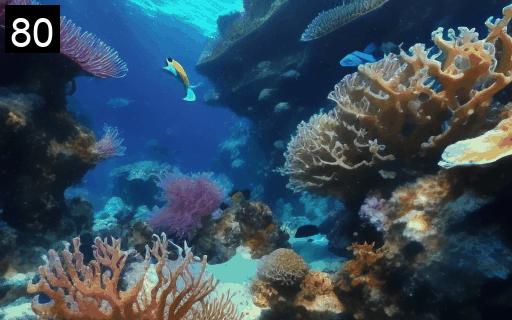} \\
        \multicolumn{5}{c}{\small (f) \textsf{"A school of colorful fish swimming in a coral reef, ultra high quality, 2K."}}\vspace{2pt} \\
        \includegraphics[width=0.2\linewidth]{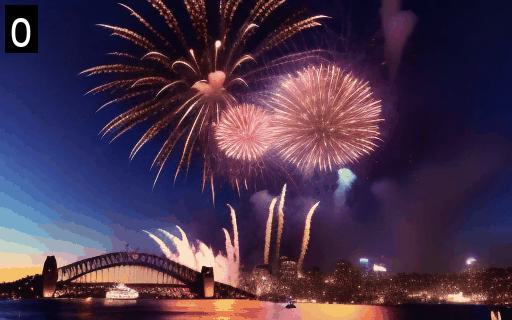} &
        \includegraphics[width=0.2\linewidth]{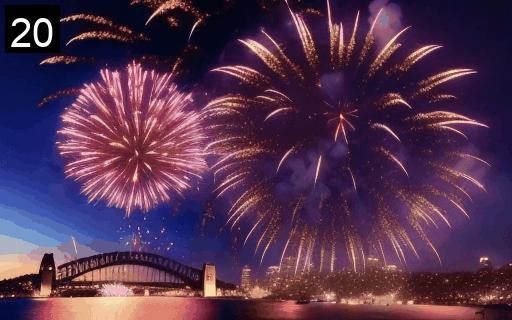} &
        \includegraphics[width=0.2\linewidth]{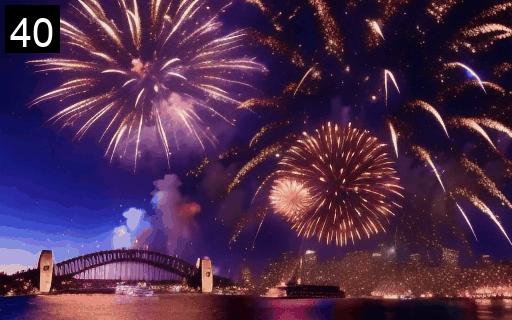} &
        \includegraphics[width=0.2\linewidth]{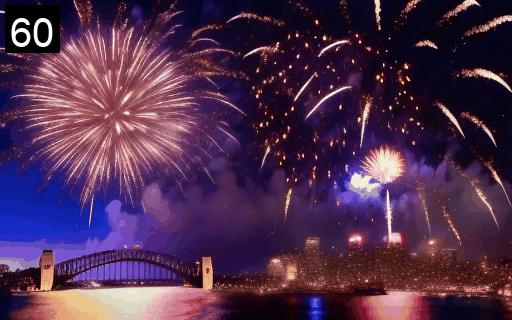} &
        \includegraphics[width=0.2\linewidth]{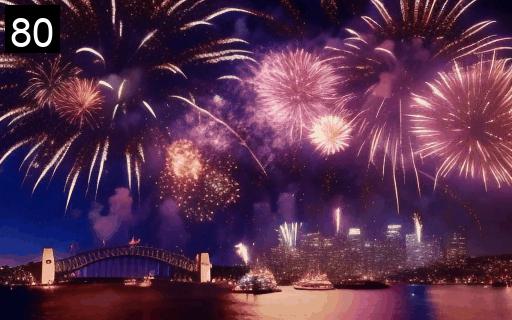} \\
        \multicolumn{5}{c}{\small (g) \textsf{"A spectacular fireworks display over Sydney Harbour, 4K, high resolution."}} \\

    \end{tabular}
}
\captionof{figure}{
    Videos generated by FIFO-Diffusion with VideoCrafter2.
    The number on the top left of each frame indicates the frame index.
    }
\label{fig:qual:vc2_1}

\clearpage
{\centering
\scalebox{1}{
    \setlength{\tabcolsep}{1pt}
    \begin{tabular}{ccccc}
        \includegraphics[width=0.2\linewidth]{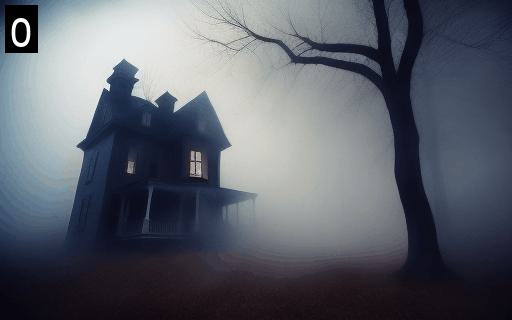} &
        \includegraphics[width=0.2\linewidth]{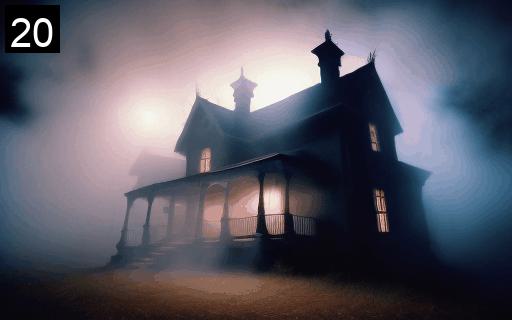} &
        \includegraphics[width=0.2\linewidth]{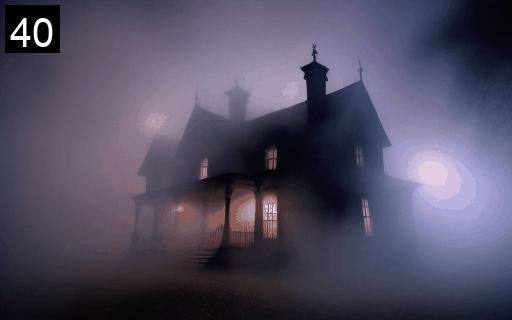} &
        \includegraphics[width=0.2\linewidth]{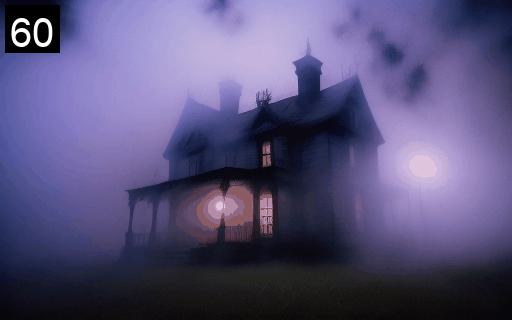} &
        \includegraphics[width=0.2\linewidth]{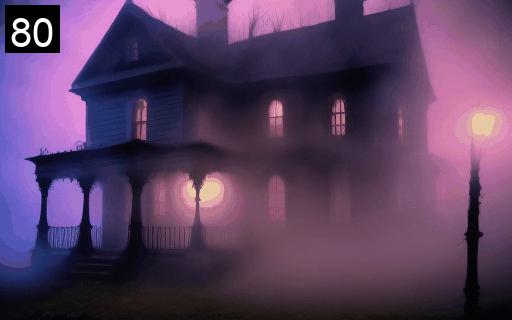} \\
        \multicolumn{5}{c}{\small (a) \textsf{"A spooky haunted house, foggy night, high definition."}}\vspace{2pt} \\
        \includegraphics[width=0.2\linewidth]{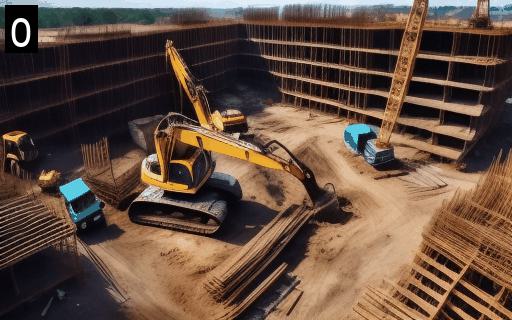} &
        \includegraphics[width=0.2\linewidth]{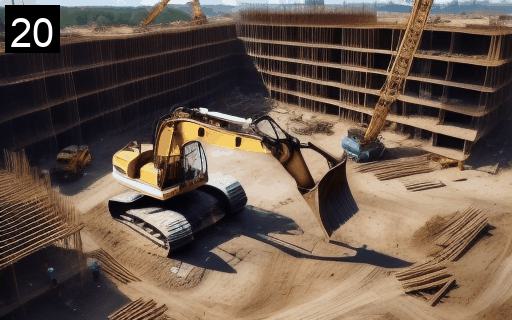} &
        \includegraphics[width=0.2\linewidth]{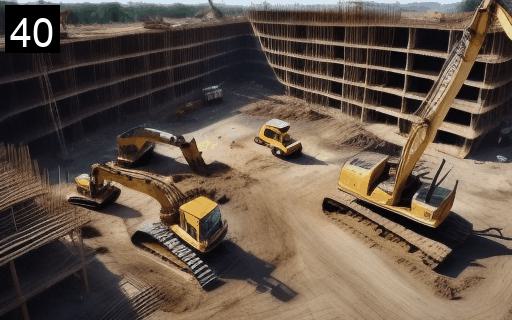} &
        \includegraphics[width=0.2\linewidth]{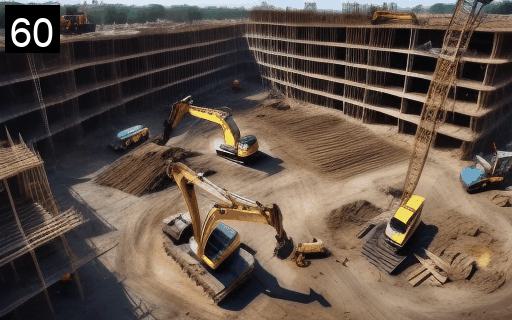} &
        \includegraphics[width=0.2\linewidth]{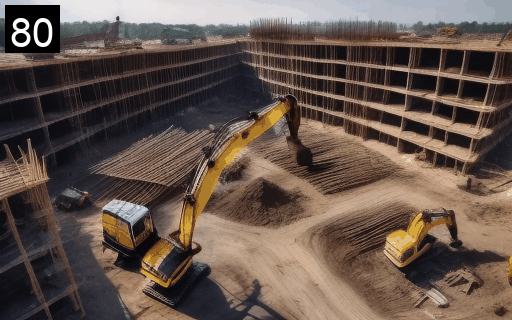} \\
        \multicolumn{5}{c}{\small (b) \textsf{"A time-lapse of a busy construction site, high definition, 4K."}} \vspace{2pt}\\
        \includegraphics[width=0.2\linewidth]{fig/videocrafter2_jpgs/17/0.jpg} &
        \includegraphics[width=0.2\linewidth]{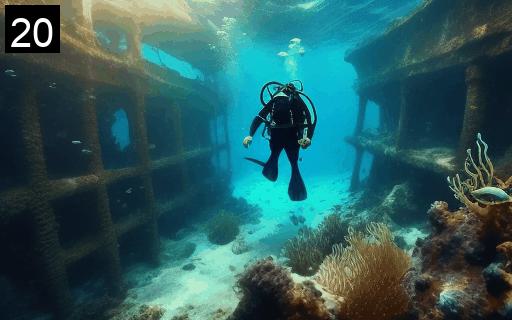} &
        \includegraphics[width=0.2\linewidth]{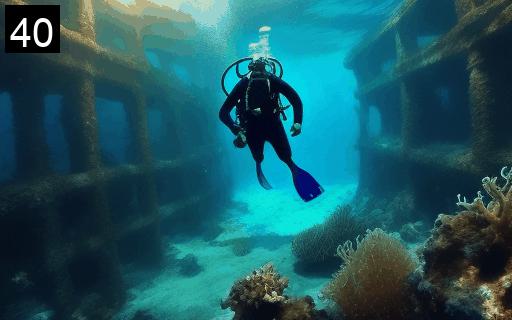} &
        \includegraphics[width=0.2\linewidth]{fig/videocrafter2_jpgs/17/3.jpg} &
        \includegraphics[width=0.2\linewidth]{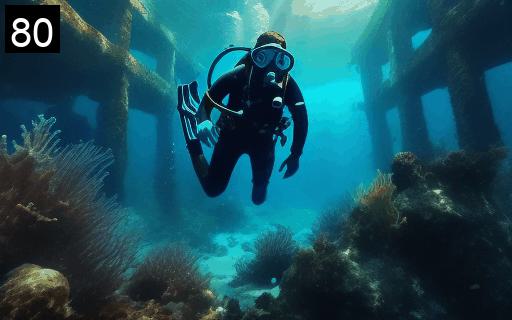} \\
        \multicolumn{5}{c}{\small (c) \textsf{"A vibrant underwater scene of a scuba diver exploring a shipwreck, 2K, photorealistic."}}\vspace{2pt} \\
        \includegraphics[width=0.2\linewidth]{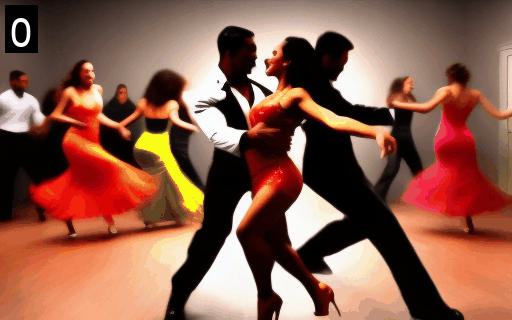} &
        \includegraphics[width=0.2\linewidth]{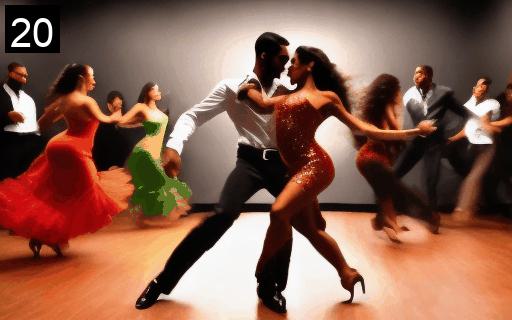} &
        \includegraphics[width=0.2\linewidth]{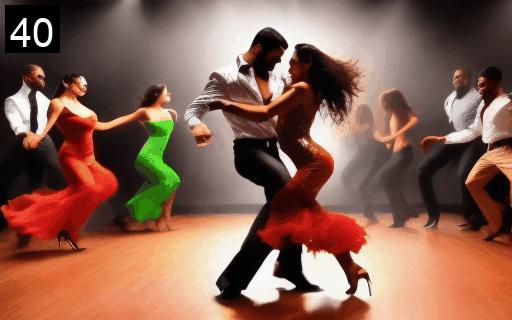} &
        \includegraphics[width=0.2\linewidth]{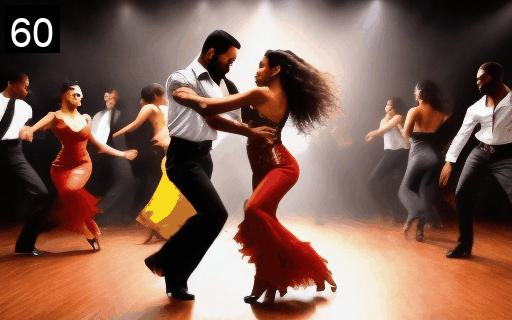} &
        \includegraphics[width=0.2\linewidth]{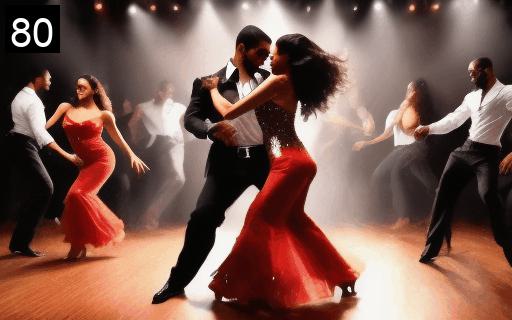} \\
        \multicolumn{5}{c}{\small (d) \textsf{"A vibrant, fast-paced salsa dance performance, ultra high quality, 2K."}}\vspace{2pt} \\
        \includegraphics[width=0.2\linewidth]{fig/videocrafter2_jpgs/19/0.jpg} &
        \includegraphics[width=0.2\linewidth]{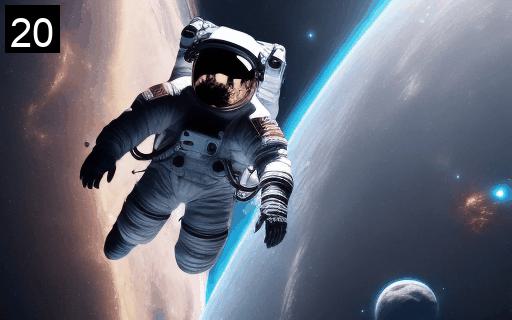} &
        \includegraphics[width=0.2\linewidth]{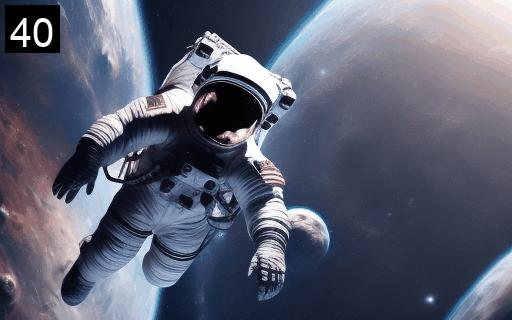} &
        \includegraphics[width=0.2\linewidth]{fig/videocrafter2_jpgs/19/3.jpg} &
        \includegraphics[width=0.2\linewidth]{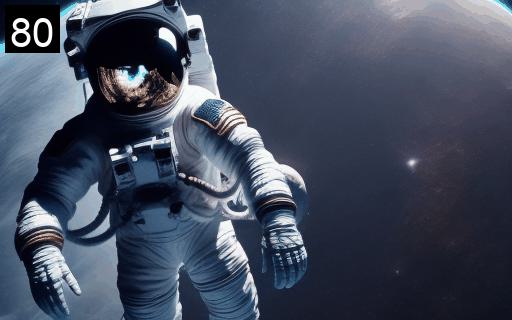} \\
        \multicolumn{5}{c}{\small (e) \textsf{"An astronaut floating in space, high quality, 4K resolution."}}\vspace{2pt} \\
        \includegraphics[width=0.2\linewidth]{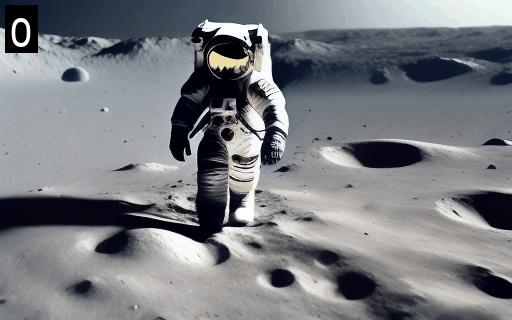} &
        \includegraphics[width=0.2\linewidth]{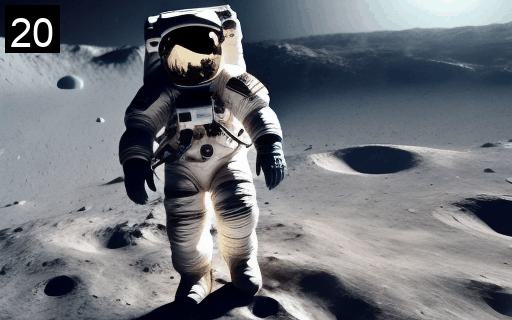} &
        \includegraphics[width=0.2\linewidth]{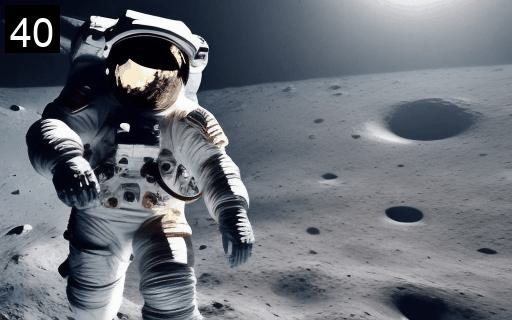} &
        \includegraphics[width=0.2\linewidth]{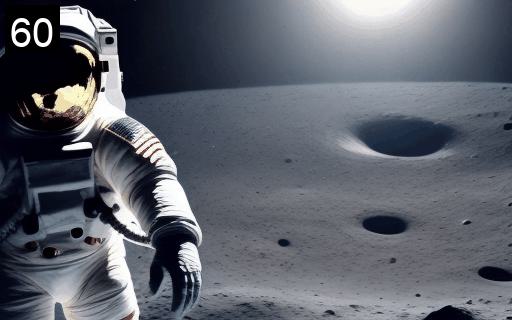} &
        \includegraphics[width=0.2\linewidth]{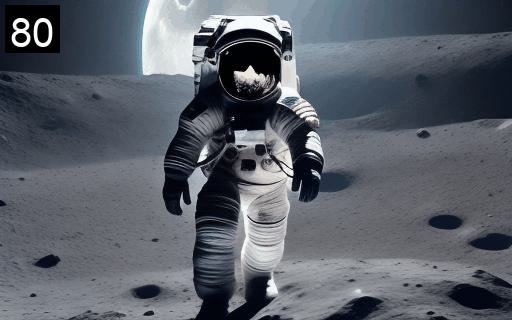} \\
        \multicolumn{5}{c}{\small (f) \textsf{"An astronaut walking on the moon's surface, high-quality, 4K resolution."}}\vspace{2pt} \\
        \includegraphics[width=0.2\linewidth]{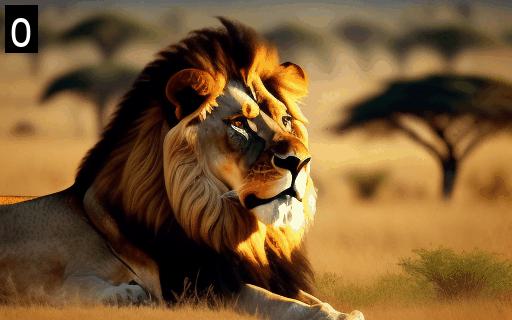} &
        \includegraphics[width=0.2\linewidth]{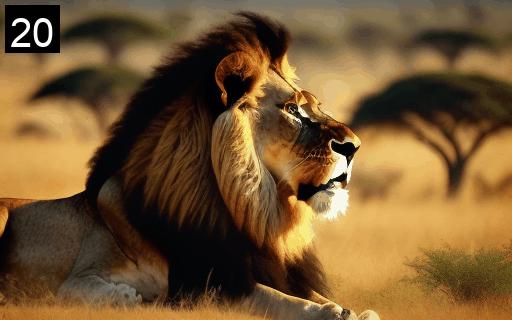} &
        \includegraphics[width=0.2\linewidth]{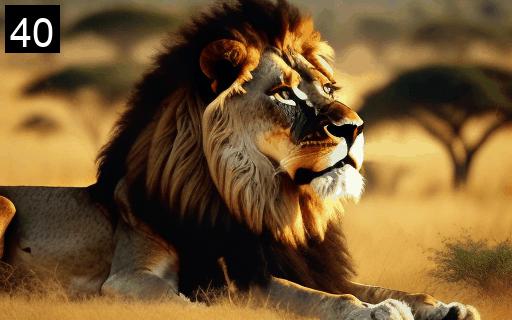} &
        \includegraphics[width=0.2\linewidth]{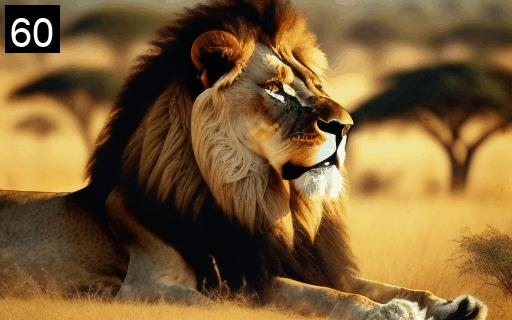} &
        \includegraphics[width=0.2\linewidth]{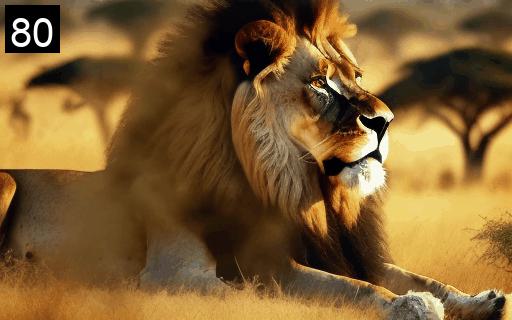} \\
        \multicolumn{5}{c}{\small (g) \textsf{"A majestic lion roaring in the African savanna, ultra HD, 4K."}} \\
    \end{tabular}
}}
\captionof{figure}{
    Videos generated by FIFO-Diffusion with VideoCrafter2.
    The number on the top left of each frame indicates the frame index.
    }
\label{fig:qual:vc2_2}

\clearpage
\subsection{VideoCrafter1}
\label{app:qual:vc1}

\scalebox{0.97}{
    \setlength{\tabcolsep}{1pt}
    \begin{tabular}{ccccc}
        \multicolumn{5}{c}{} \\
        \includegraphics[width=0.2\linewidth]{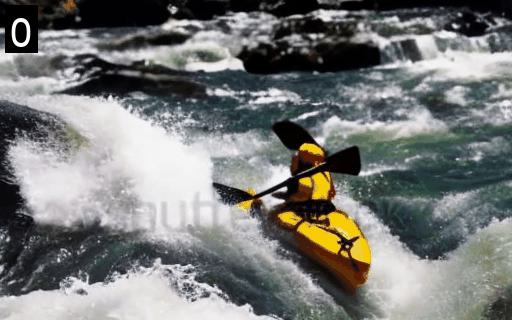} &
        \includegraphics[width=0.2\linewidth]{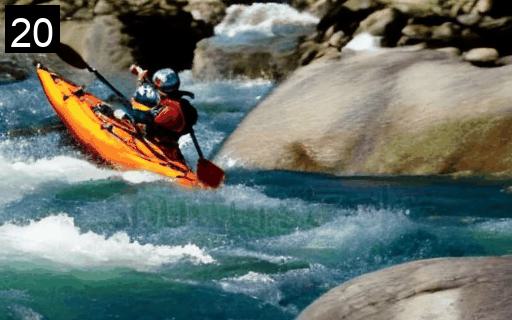} &
        \includegraphics[width=0.2\linewidth]{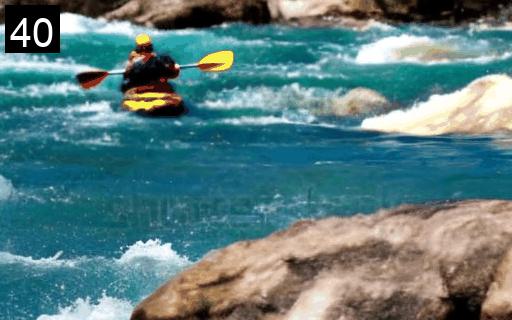} &
        \includegraphics[width=0.2\linewidth]{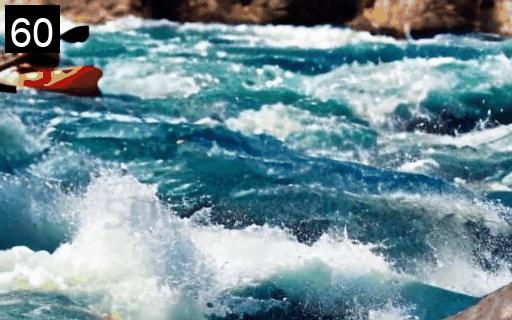} &
        \includegraphics[width=0.2\linewidth]{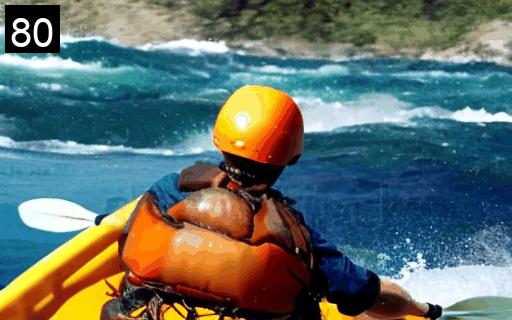} \\
        \multicolumn{5}{c}{\small (a) \textsf{"A kayaker navigating through rapids, photorealistic, 4K, high quality."}} \vspace{2pt}\\
        \includegraphics[width=0.2\linewidth]{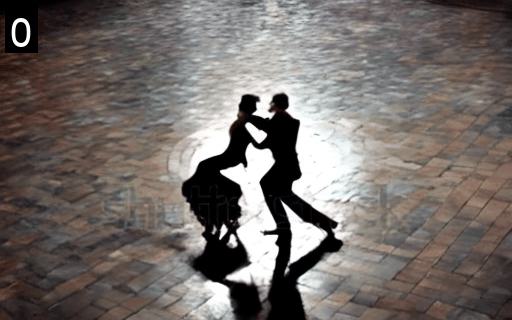} &
        \includegraphics[width=0.2\linewidth]{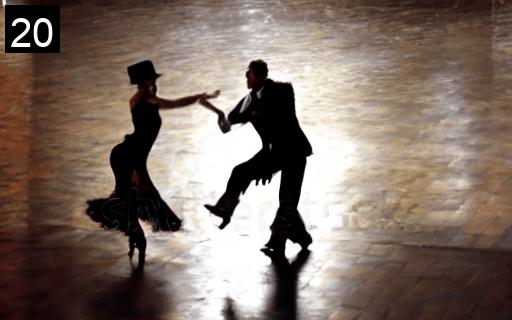} &
        \includegraphics[width=0.2\linewidth]{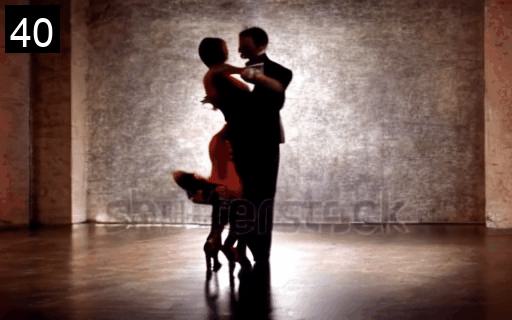} &
        \includegraphics[width=0.2\linewidth]{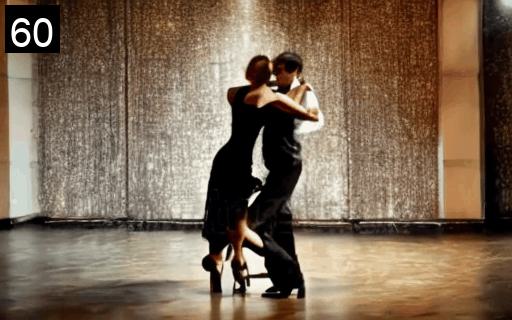} &
        \includegraphics[width=0.2\linewidth]{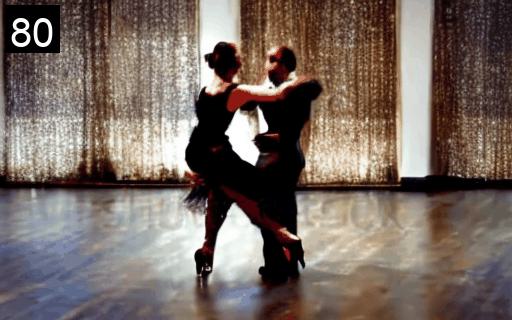} \\
        \multicolumn{5}{c}{\small (b) \textsf{"A pair of tango dancers performing in Buenos Aires, 4K, high resolution."}}\vspace{2pt} \\
        \includegraphics[width=0.2\linewidth]{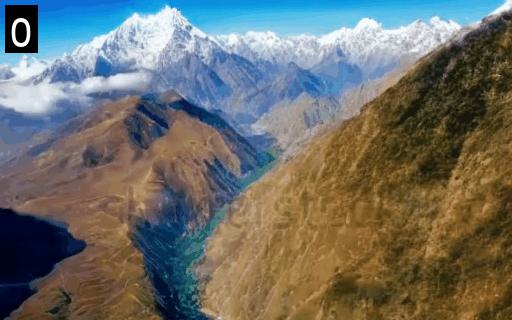} &
        \includegraphics[width=0.2\linewidth]{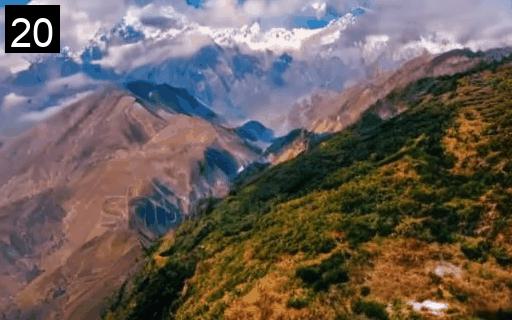} &
        \includegraphics[width=0.2\linewidth]{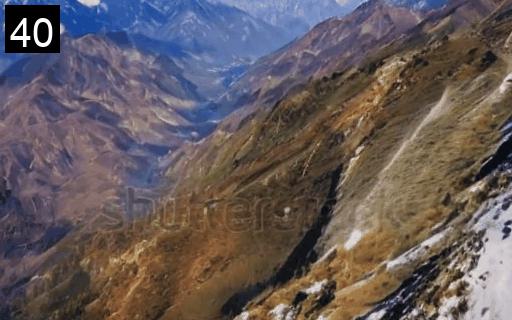} &
        \includegraphics[width=0.2\linewidth]{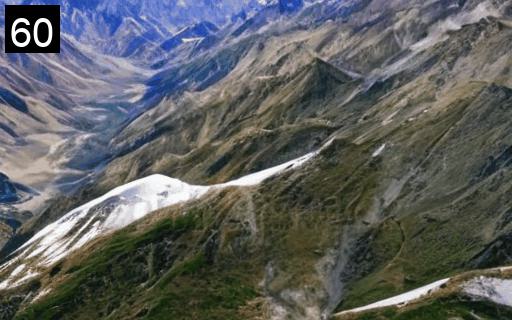} &
        \includegraphics[width=0.2\linewidth]{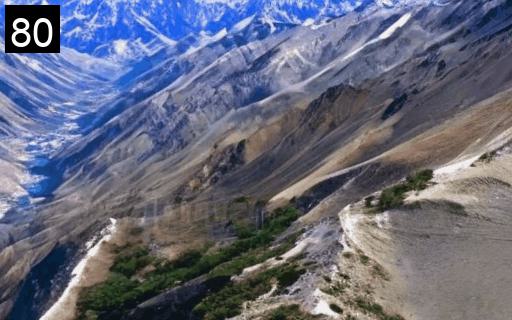} \\
        \multicolumn{5}{c}{\small (c) \textsf{"A panoramic view of the Himalayas from a drone, high definition, 4K."}} \vspace{2pt}\\
        \includegraphics[width=0.2\linewidth]{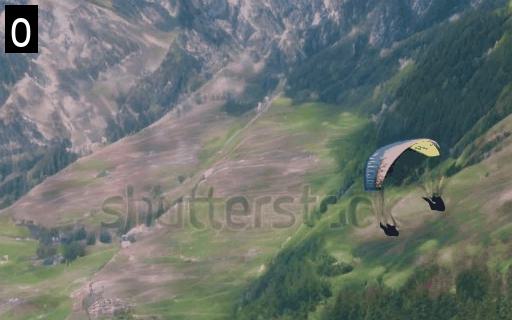} &
        \includegraphics[width=0.2\linewidth]{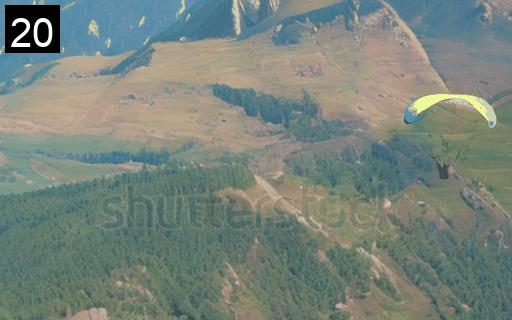} &
        \includegraphics[width=0.2\linewidth]{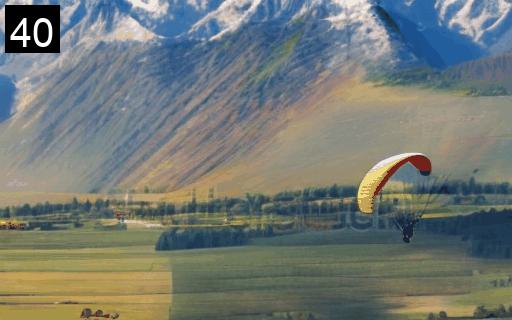} &
        \includegraphics[width=0.2\linewidth]{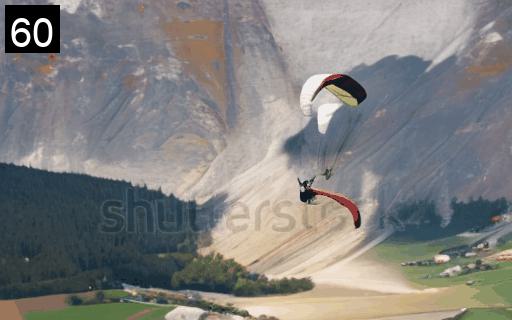} &
        \includegraphics[width=0.2\linewidth]{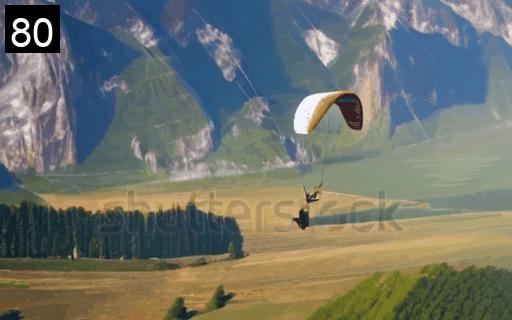} \\
        \multicolumn{5}{c}{\small (d) \textsf{"A paraglider soaring over the Alps, photorealistic, 4K, high definition."}}\vspace{2pt} \\
        \includegraphics[width=0.2\linewidth]{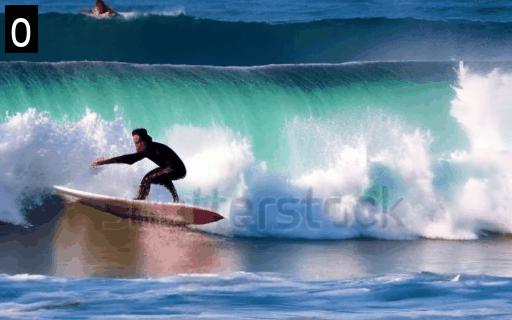} &
        \includegraphics[width=0.2\linewidth]{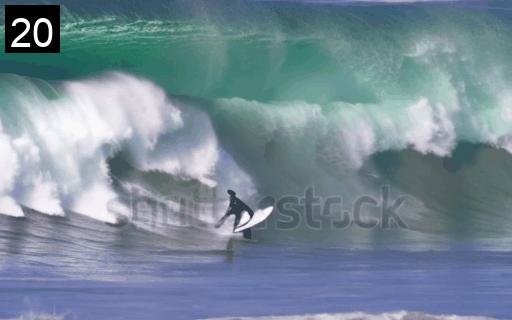} &
        \includegraphics[width=0.2\linewidth]{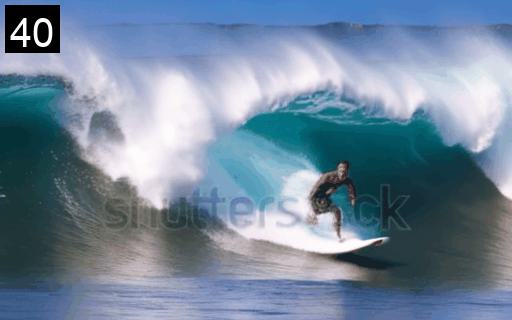} &
        \includegraphics[width=0.2\linewidth]{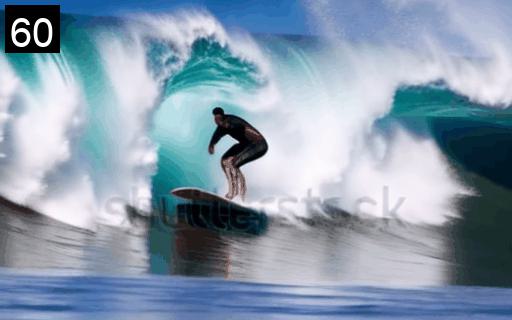} &
        \includegraphics[width=0.2\linewidth]{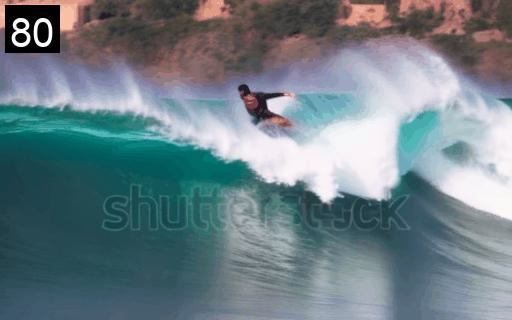} \\
        \multicolumn{5}{c}{\small (e) \textsf{"A professional surfer riding a large wave, high-quality, 4K."}} \vspace{2pt}\\
        \includegraphics[width=0.2\linewidth]{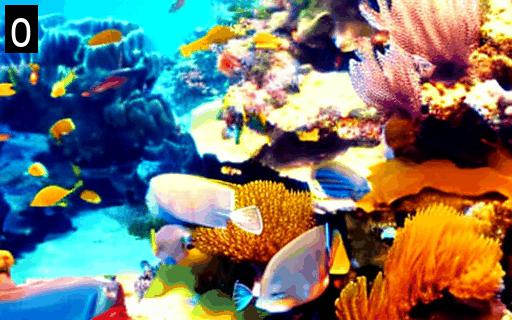} &
        \includegraphics[width=0.2\linewidth]{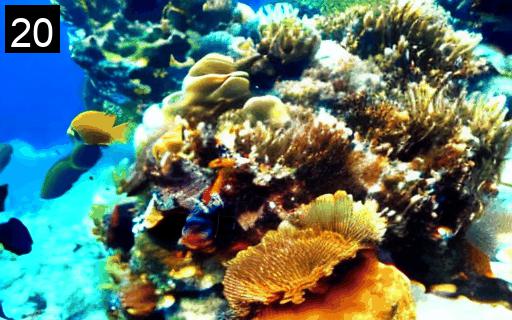} &
        \includegraphics[width=0.2\linewidth]{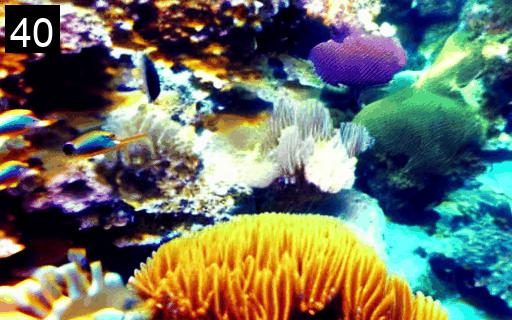} &
        \includegraphics[width=0.2\linewidth]{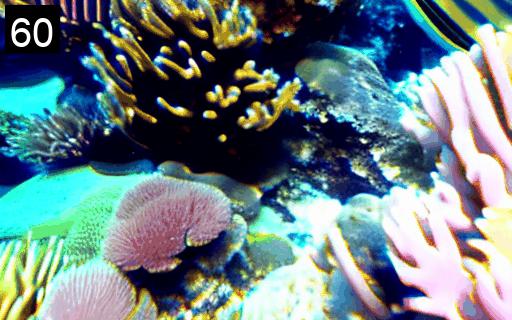} &
        \includegraphics[width=0.2\linewidth]{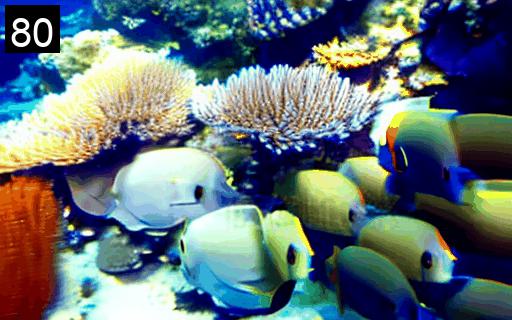} \\
        \multicolumn{5}{c}{\small (f) \textsf{"A school of colorful fish swimming in a coral reef, ultra high quality, 2K."}} \vspace{2pt}\\
        \includegraphics[width=0.2\linewidth]{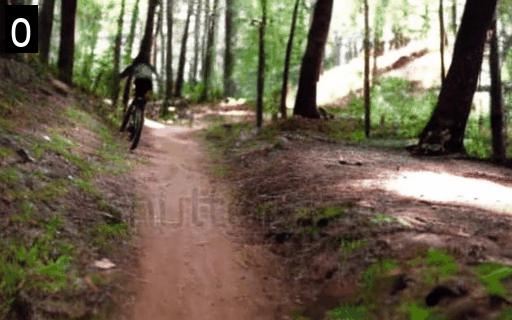} &
        \includegraphics[width=0.2\linewidth]{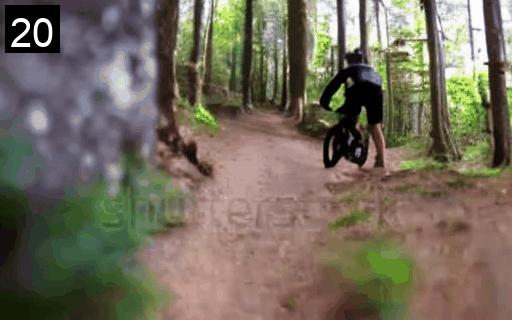} &
        \includegraphics[width=0.2\linewidth]{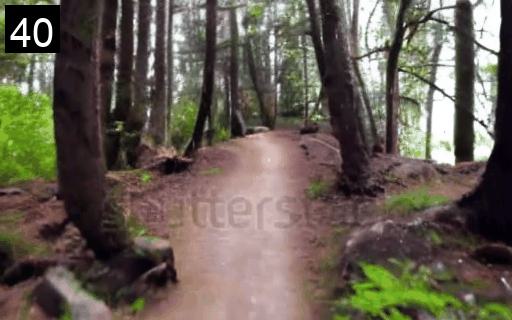} &
        \includegraphics[width=0.2\linewidth]{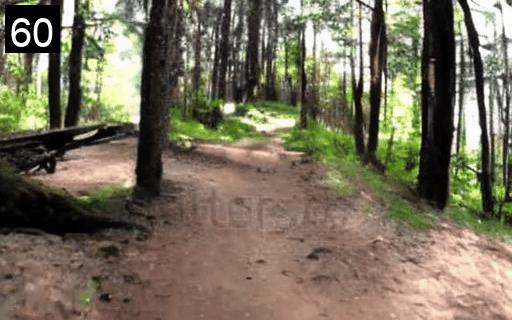} &
        \includegraphics[width=0.2\linewidth]{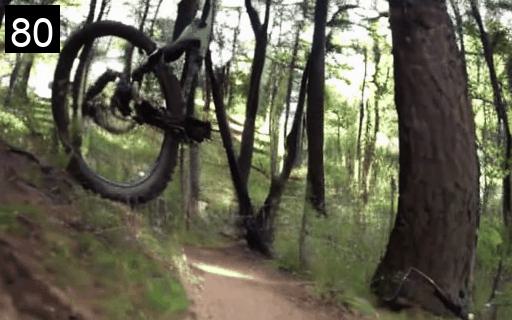} \\
        \multicolumn{5}{c}{\small (g) \textsf{"An exciting mountain bike trail ride through a forest, 2K, ultra HD."}} \vspace{2pt}\\
        \includegraphics[width=0.2\linewidth]{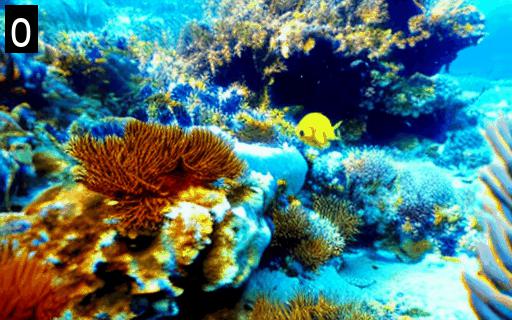} &
        \includegraphics[width=0.2\linewidth]{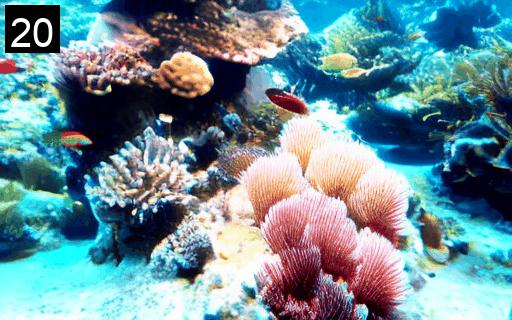} &
        \includegraphics[width=0.2\linewidth]{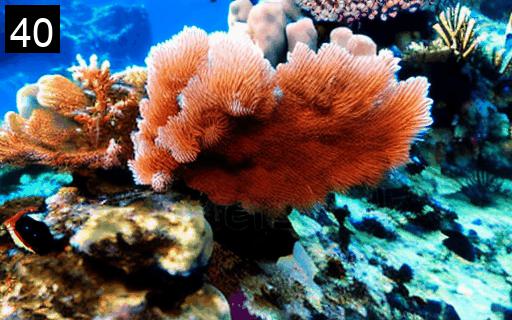} &
        \includegraphics[width=0.2\linewidth]{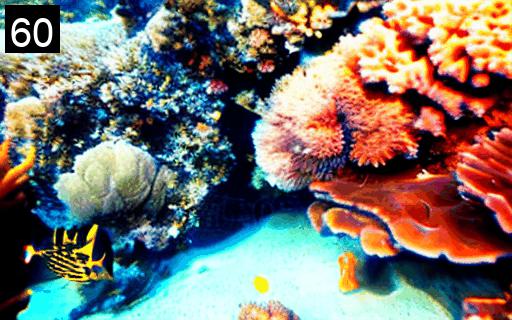} &
        \includegraphics[width=0.2\linewidth]{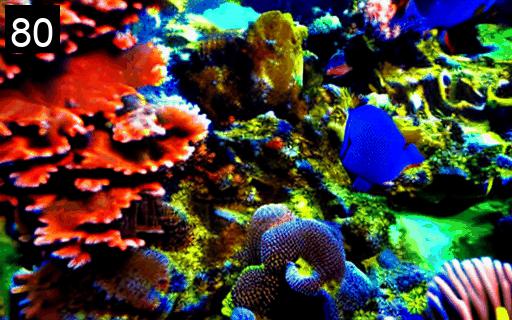} \\
        \multicolumn{5}{c}{\small (h) \textsf{"A vibrant coral reef with diverse marine life, photorealistic, 2K resolution."}} \\
    \end{tabular} 
}
\captionof{figure}{
    Videos generated by FIFO-Diffusion with VideoCrafter1.
    The number on the top left of each frame indicates the frame index.
    }
\label{fig:qual:vc1_0}

\clearpage
\subsection{zeroscope}
\label{app:qual:zeroscope}
\vspace{2pt}
{\centering
\scalebox{1}{
    \setlength{\tabcolsep}{1pt}
    \begin{tabular}{ccccc}
        \includegraphics[width=0.2\linewidth]{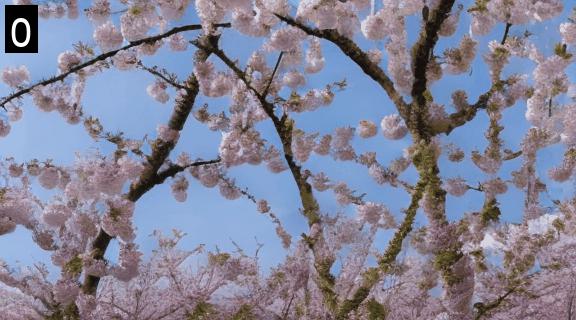} &
        \includegraphics[width=0.2\linewidth]{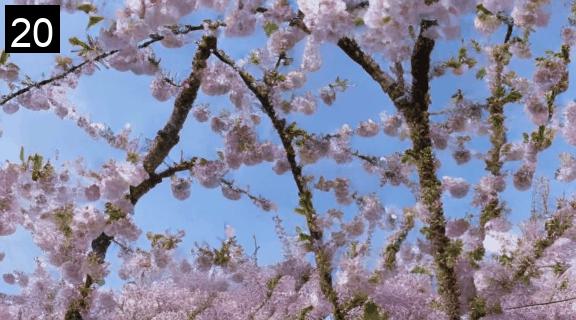} &
        \includegraphics[width=0.2\linewidth]{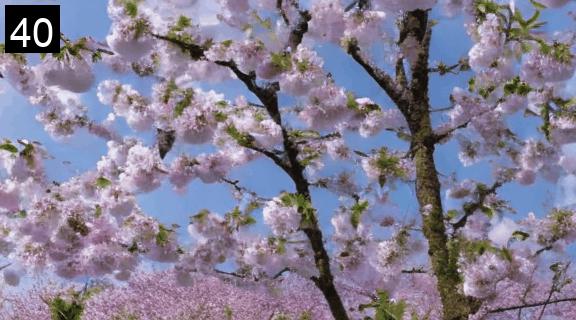} &
        \includegraphics[width=0.2\linewidth]{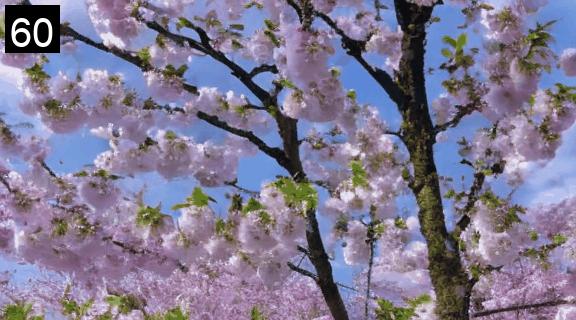} &
        \includegraphics[width=0.2\linewidth]{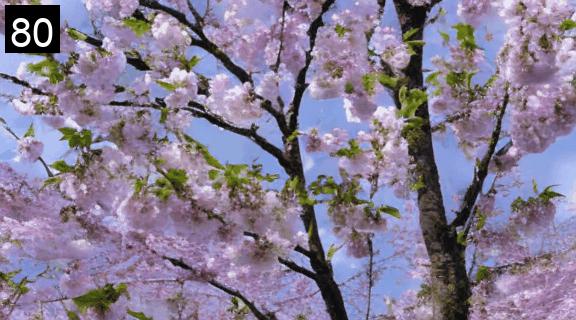} \\
        \multicolumn{5}{c}{\small (a) \textsf{"A beautiful cherry blossom festival, time-lapse, high quality."}} \vspace{2pt}\\
        \includegraphics[width=0.2\linewidth]{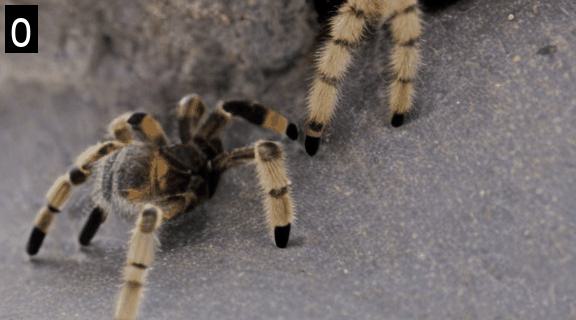} &
        \includegraphics[width=0.2\linewidth]{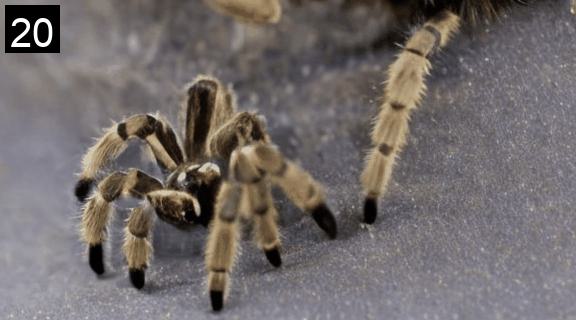} &
        \includegraphics[width=0.2\linewidth]{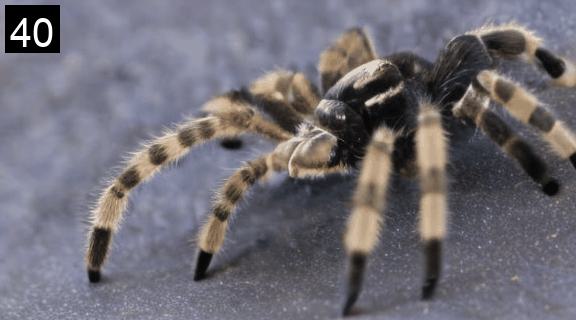} &
        \includegraphics[width=0.2\linewidth]{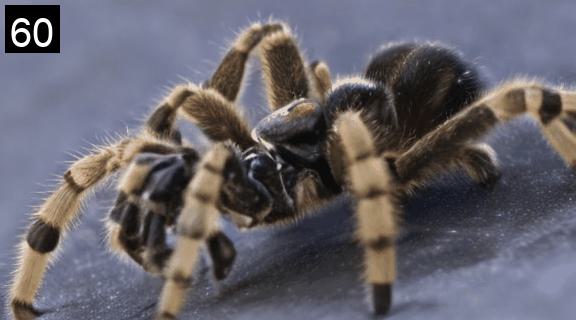} &
        \includegraphics[width=0.2\linewidth]{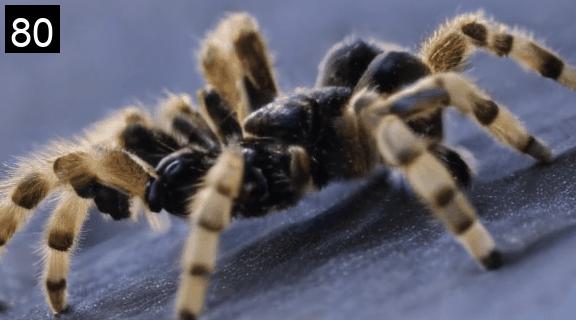} \\
        \multicolumn{5}{c}{\small (b) \textsf{"A close-up of a tarantula walking, high definition."}}\vspace{2pt} \\
        \includegraphics[width=0.2\linewidth]{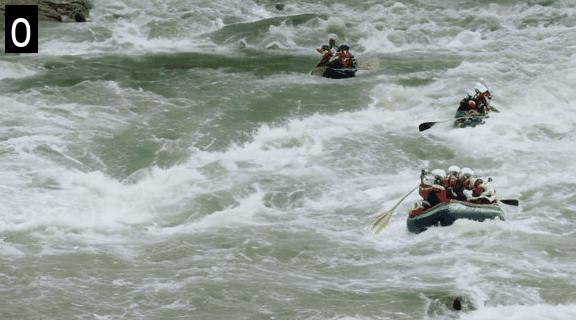} &
        \includegraphics[width=0.2\linewidth]{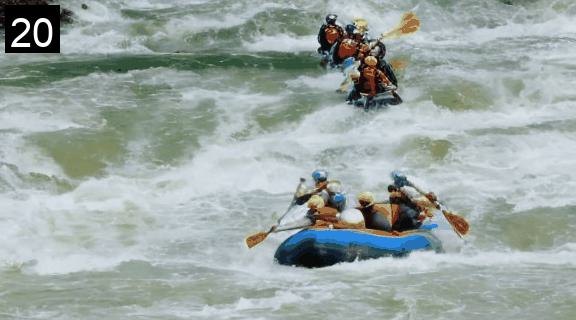} &
        \includegraphics[width=0.2\linewidth]{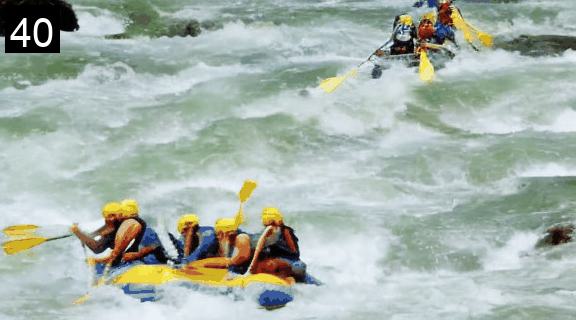} &
        \includegraphics[width=0.2\linewidth]{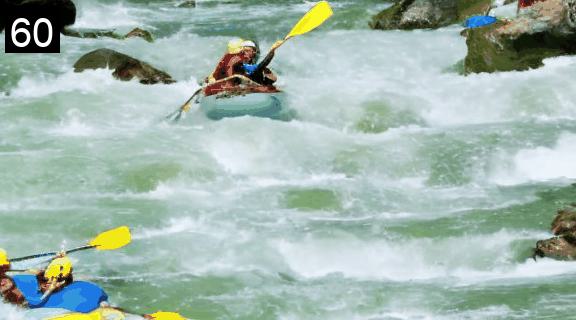} &
        \includegraphics[width=0.2\linewidth]{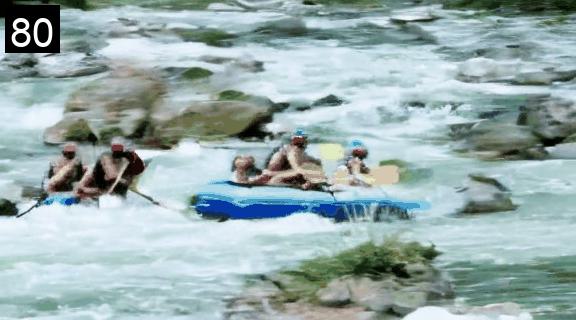} \\
        \multicolumn{5}{c}{\small (c) \textsf{"A thrilling white water rafting adventure, high definition."}} \vspace{2pt}\\
        \includegraphics[width=0.2\linewidth]{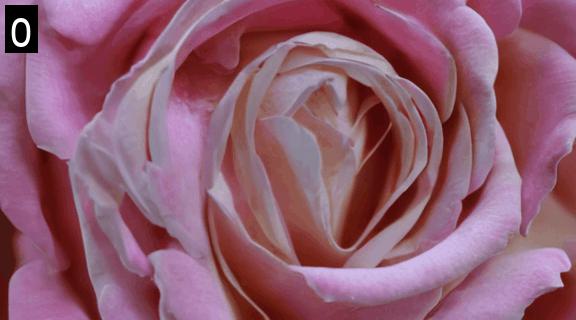} &
        \includegraphics[width=0.2\linewidth]{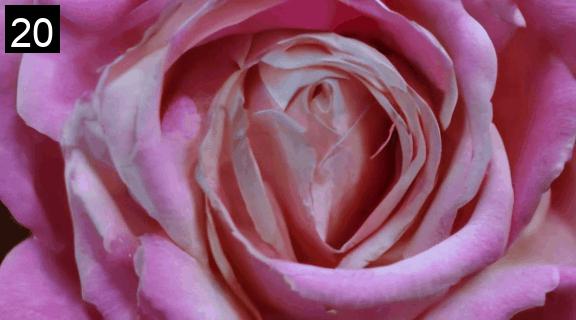} &
        \includegraphics[width=0.2\linewidth]{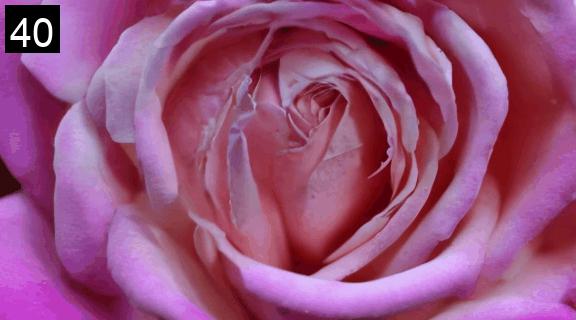} &
        \includegraphics[width=0.2\linewidth]{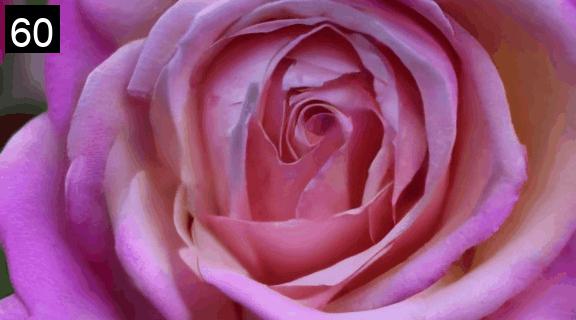} &
        \includegraphics[width=0.2\linewidth]{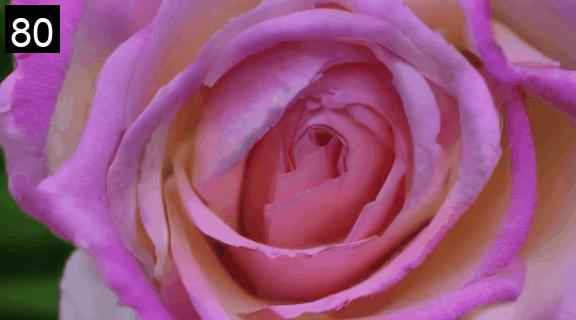} \\
        \multicolumn{5}{c}{\small  (d) \textsf{"A detailed macro shot of a blooming rose, 4K."}} \vspace{2pt}\\
        \includegraphics[width=0.2\linewidth]{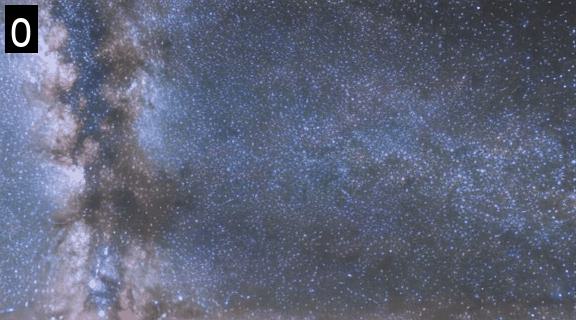} &
        \includegraphics[width=0.2\linewidth]{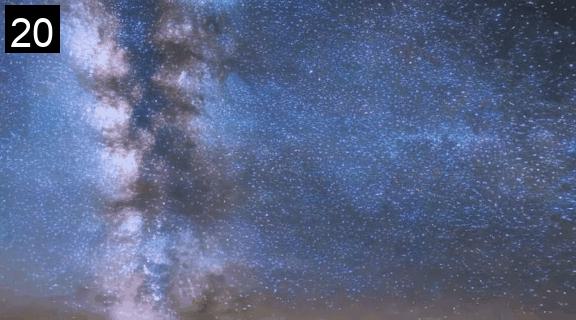} &
        \includegraphics[width=0.2\linewidth]{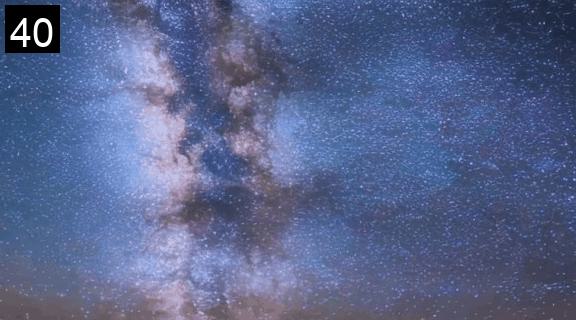} &
        \includegraphics[width=0.2\linewidth]{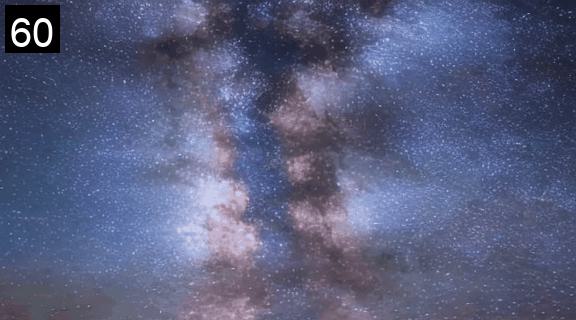} &
        \includegraphics[width=0.2\linewidth]{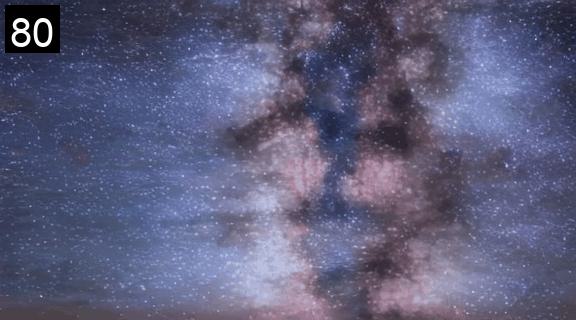} \\
        \multicolumn{5}{c}{\small (e) \textsf{"A panoramic view of the Milky Way, ultra HD."}} \vspace{2pt}\\
        \includegraphics[width=0.2\linewidth]{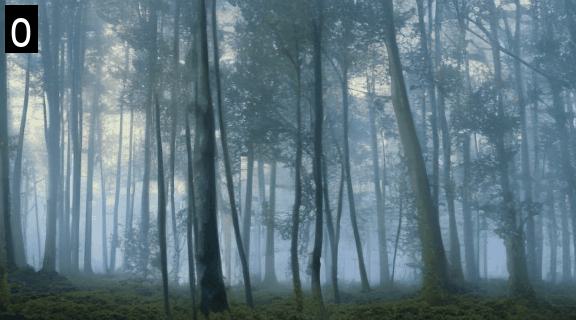} &
        \includegraphics[width=0.2\linewidth]{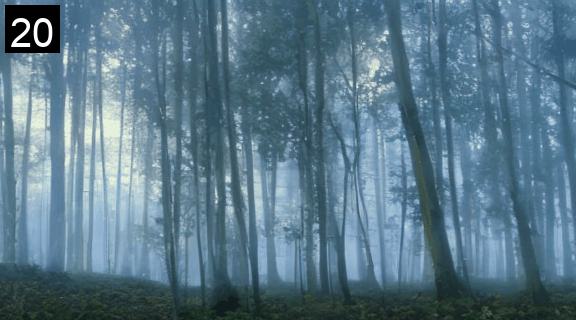} &
        \includegraphics[width=0.2\linewidth]{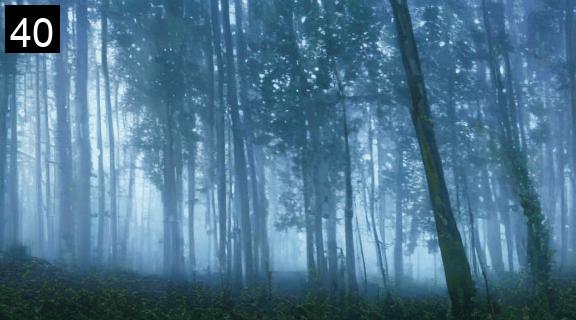} &
        \includegraphics[width=0.2\linewidth]{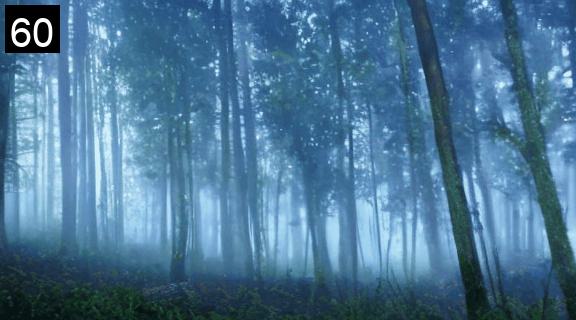} &
        \includegraphics[width=0.2\linewidth]{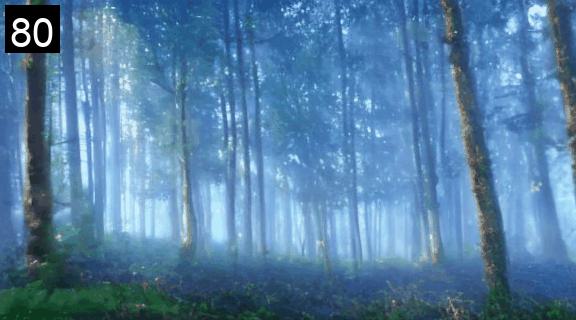} \\
        \multicolumn{5}{c}{\small(f) \textsf{"A mysterious foggy forest at dawn, high quality, 4K."}}\vspace{2pt} \\
        \includegraphics[width=0.2\linewidth]{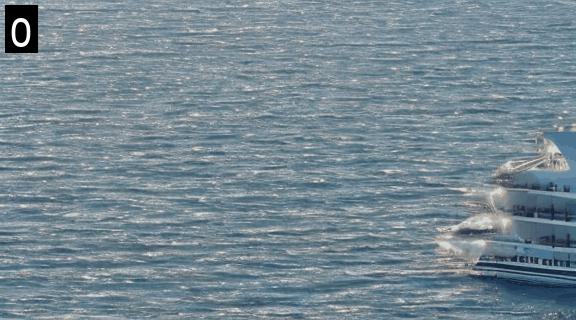} &
        \includegraphics[width=0.2\linewidth]{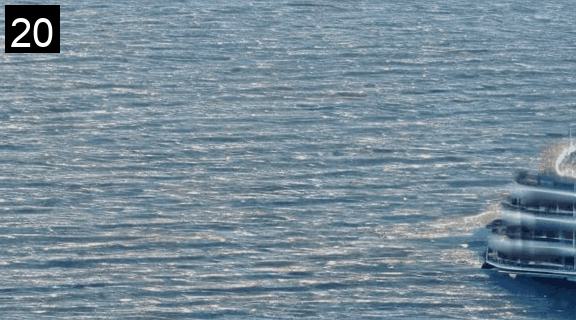} &
        \includegraphics[width=0.2\linewidth]{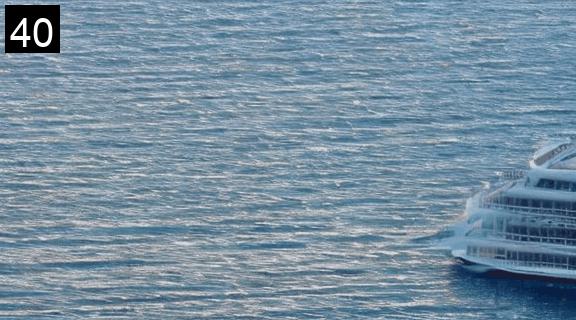} &
        \includegraphics[width=0.2\linewidth]{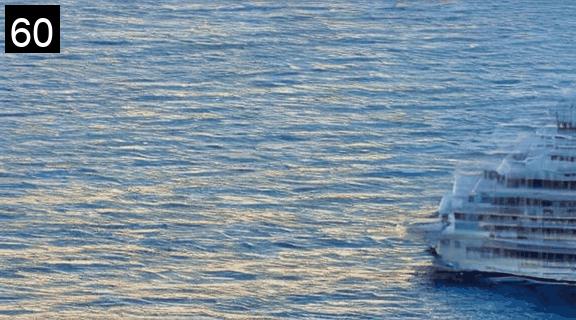} &
        \includegraphics[width=0.2\linewidth]{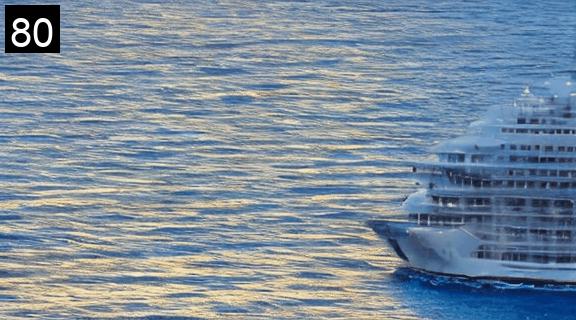} \\
        \multicolumn{5}{c}{\small (g) \textsf{"A scenic cruise ship journey at sunset, ultra HD."}}\vspace{2pt} \\
        \includegraphics[width=0.2\linewidth]{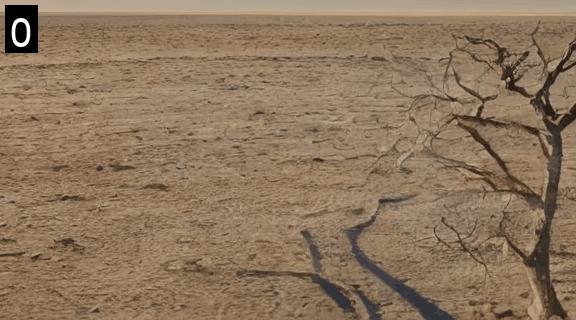} &
        \includegraphics[width=0.2\linewidth]{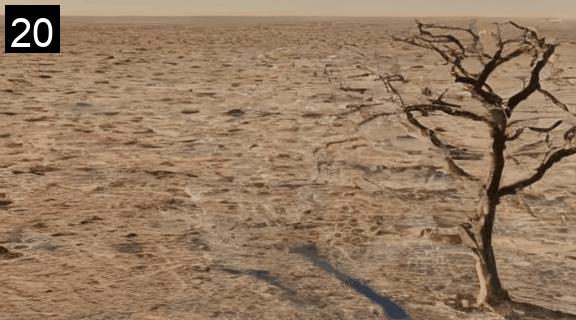} &
        \includegraphics[width=0.2\linewidth]{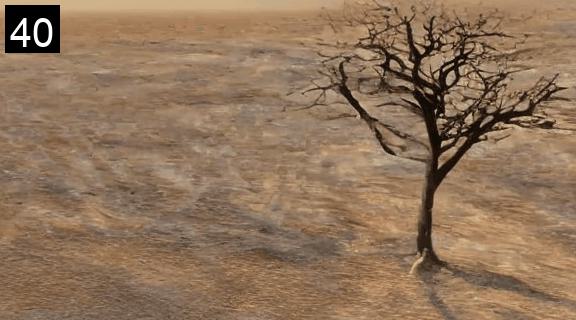} &
        \includegraphics[width=0.2\linewidth]{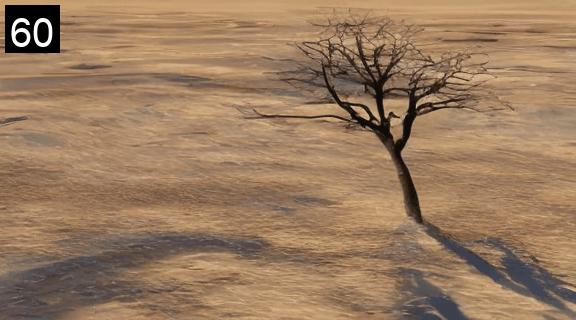} &
        \includegraphics[width=0.2\linewidth]{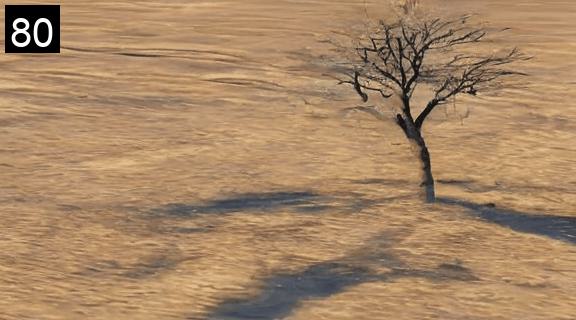} \\
        \multicolumn{5}{c}{\small (h) \textsf{"A lone tree in a vast desert, sunset, high definition."}} \\
    \end{tabular}
}
}
\captionof{figure}{
    Videos generated by FIFO-Diffusion with zeroscope.
    The number on the top left of each frame indicates the frame index.
    }
\label{fig:qual:zero_0}

\clearpage
\subsection{Open-Sora Plan}
\label{app:qual:opensoraplan}

\scalebox{0.97}{
    \setlength{\tabcolsep}{1pt}
    \begin{tabular}{ccccc}
        \multicolumn{5}{c}{} \\
        \includegraphics[width=0.2\linewidth]{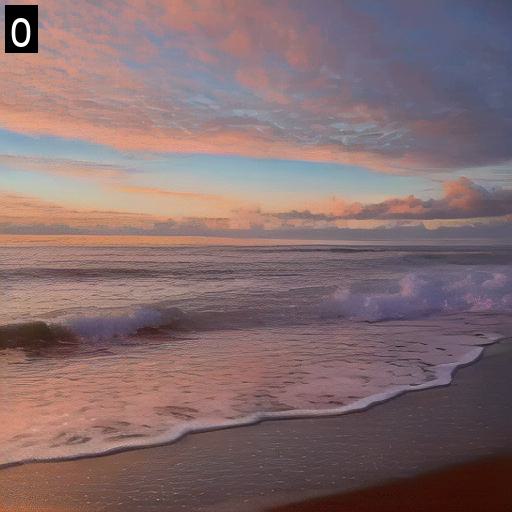} &
        \includegraphics[width=0.2\linewidth]{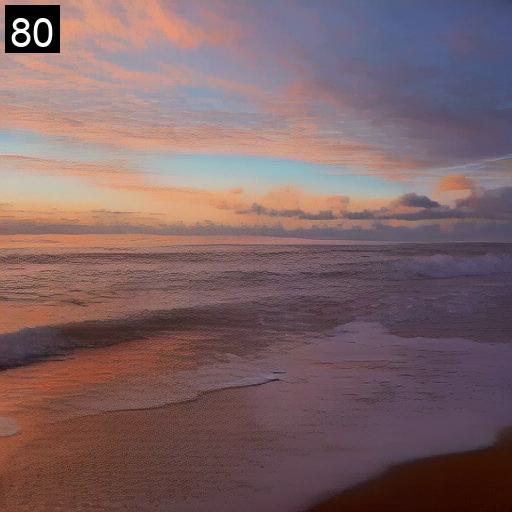} &
        \includegraphics[width=0.2\linewidth]{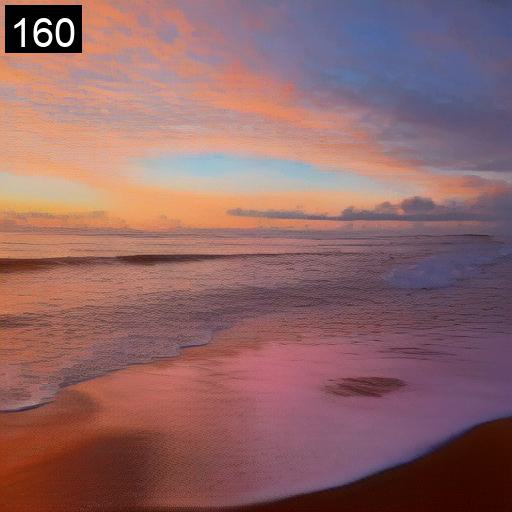} &
        \includegraphics[width=0.2\linewidth]{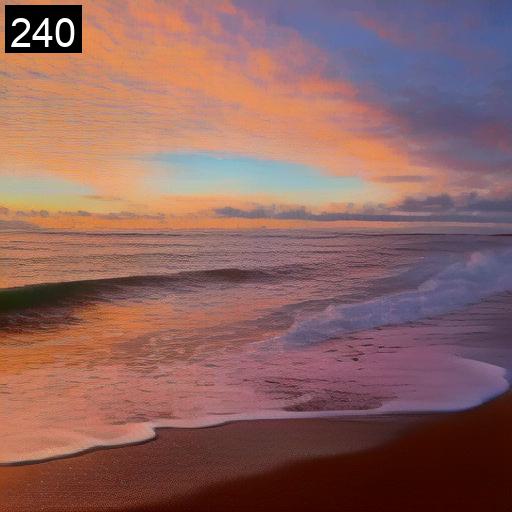} &
        \includegraphics[width=0.2\linewidth]{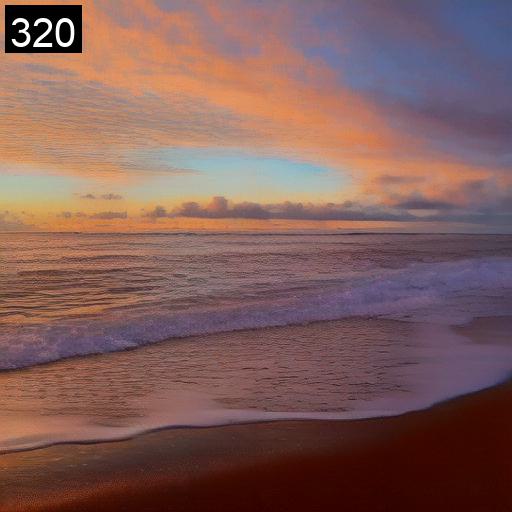} \\
        \multicolumn{5}{c}{\small(a) \textsf{"A quiet beach at dawn, the waves gently lapping at the shore and the sky painted in pastel hues."}} \vspace{2pt}\\
        \includegraphics[width=0.2\linewidth]{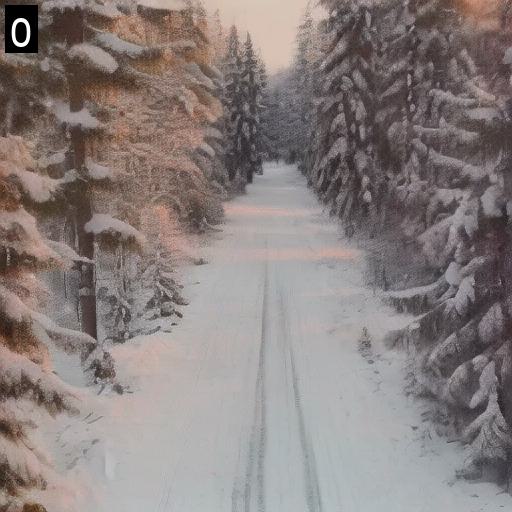} &
        \includegraphics[width=0.2\linewidth]{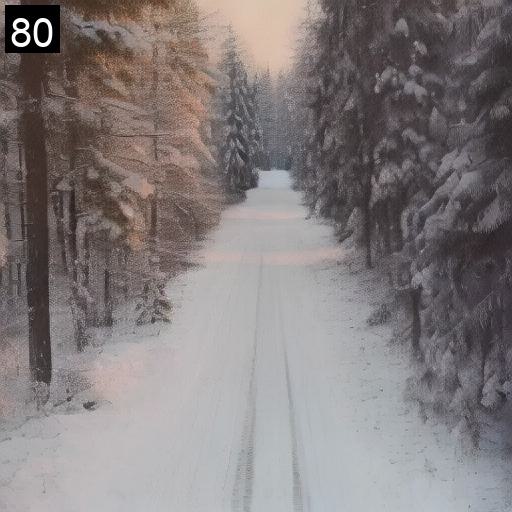} &
        \includegraphics[width=0.2\linewidth]{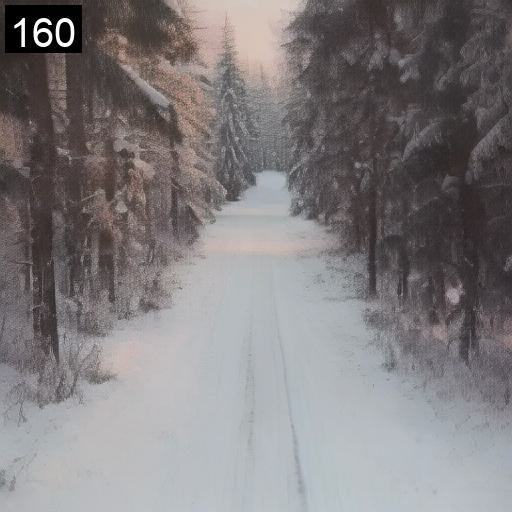} &
        \includegraphics[width=0.2\linewidth]{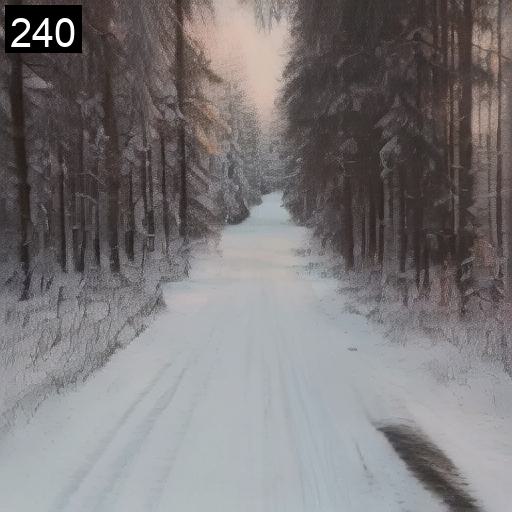} &
        \includegraphics[width=0.2\linewidth]{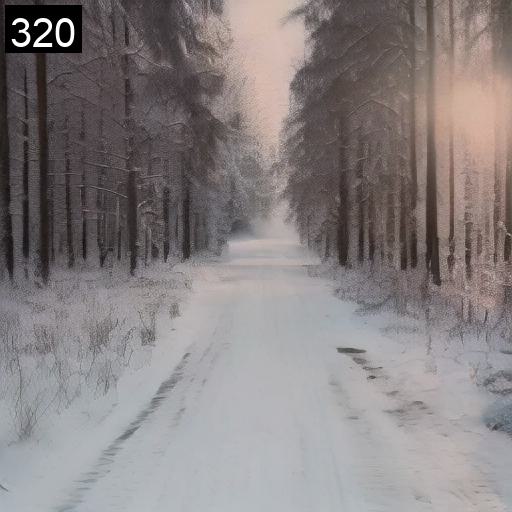} \\
        \multicolumn{5}{c}{\small (b) \textsf{"A snowy forest landscape with a dirt road running through it. The road is flanked~$\text{\myldots}$"}}\vspace{2pt} \\
        \includegraphics[width=0.2\linewidth]{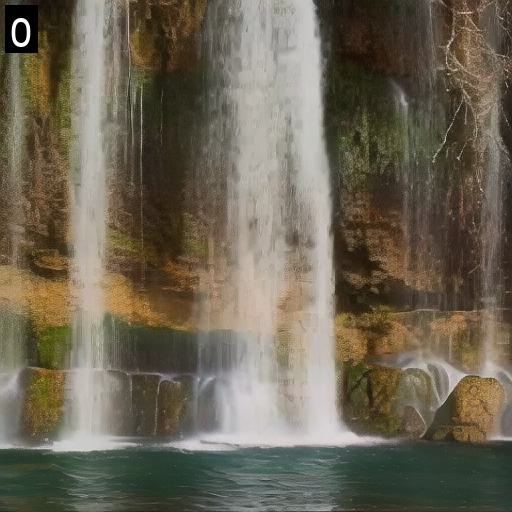} &
        \includegraphics[width=0.2\linewidth]{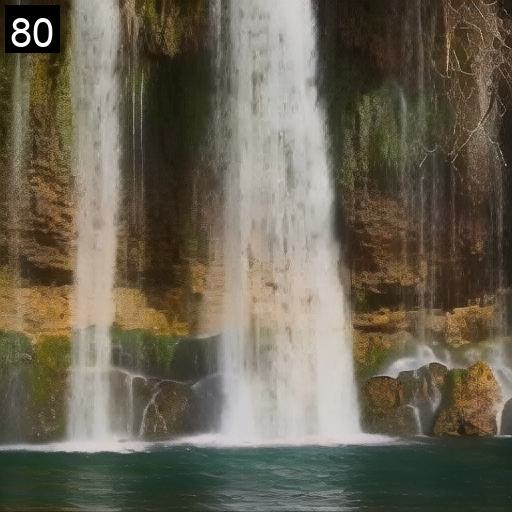} &
        \includegraphics[width=0.2\linewidth]{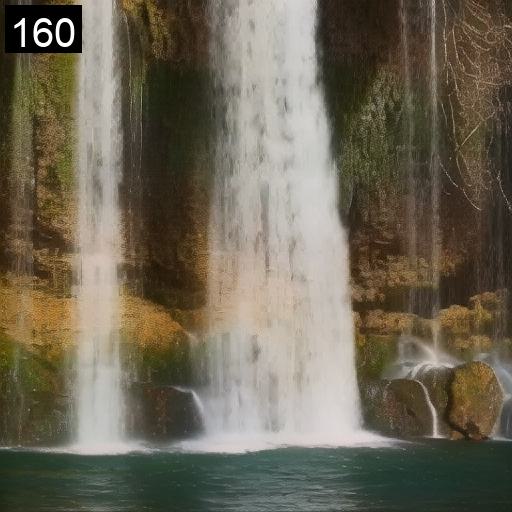} &
        \includegraphics[width=0.2\linewidth]{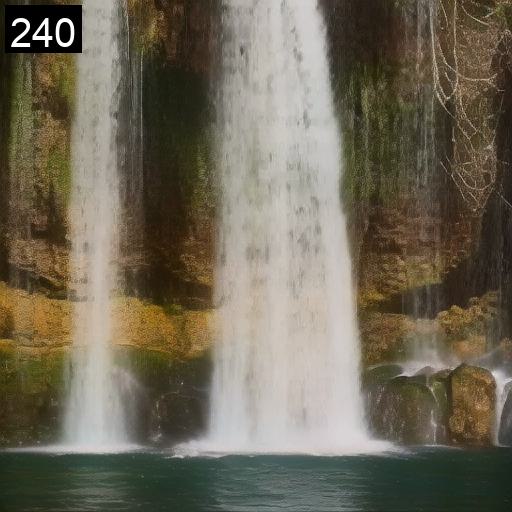} &
        \includegraphics[width=0.2\linewidth]{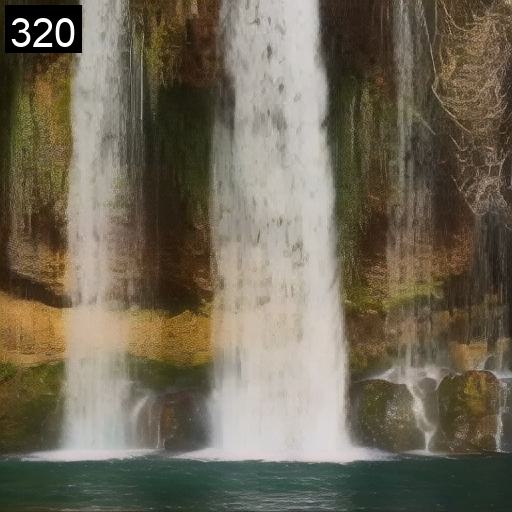} \\
        \multicolumn{5}{c}{\small (c) \textsf{"The majestic beauty of a waterfall cascading down a cliff into a serene lake."}} \vspace{2pt}\\
        \includegraphics[width=0.2\linewidth]{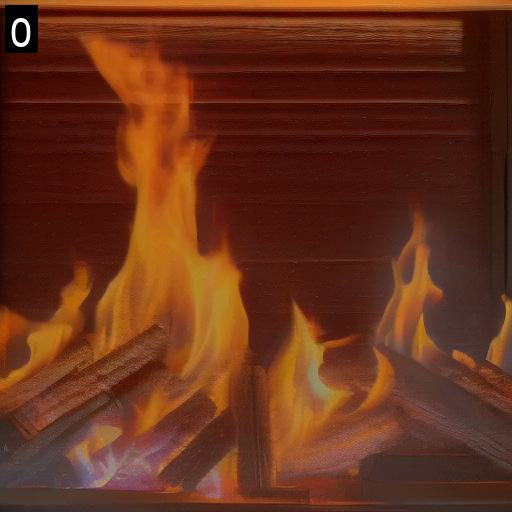} &
        \includegraphics[width=0.2\linewidth]{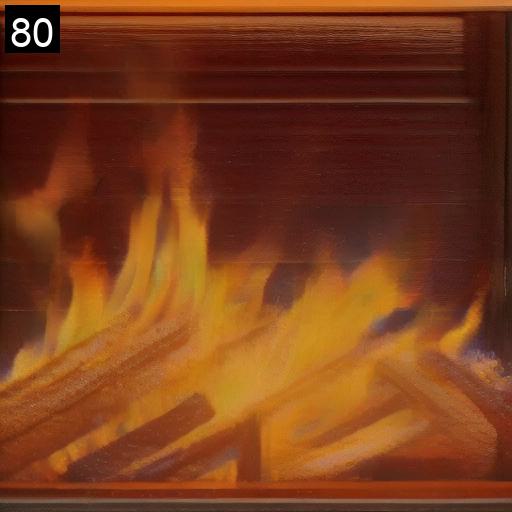} &
        \includegraphics[width=0.2\linewidth]{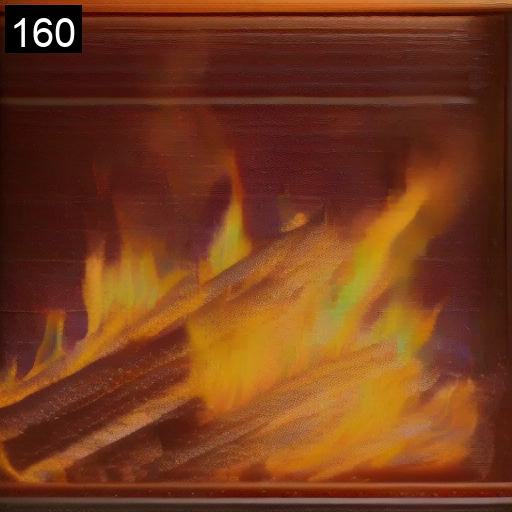} &
        \includegraphics[width=0.2\linewidth]{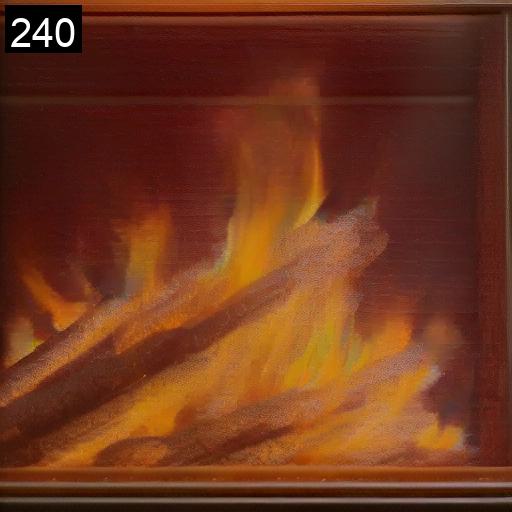} &
        \includegraphics[width=0.2\linewidth]{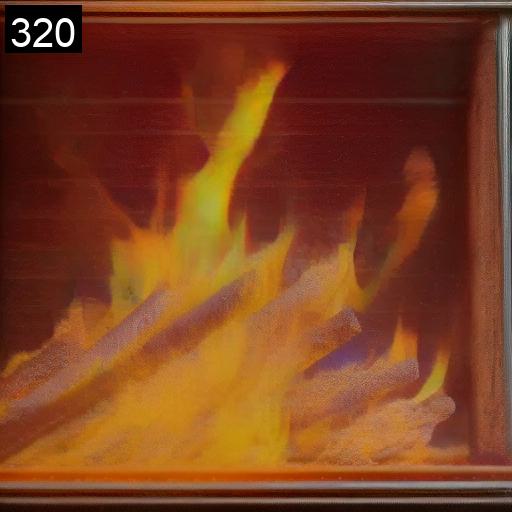} \\
        \multicolumn{5}{c}{\small (d) \textsf{"Slow pan upward of blazing oak fire in an indoor fireplace."}}\vspace{2pt} \\
        \includegraphics[width=0.2\linewidth]{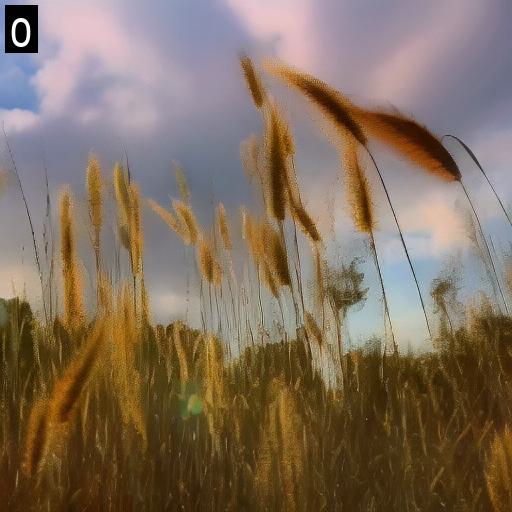} &
        \includegraphics[width=0.2\linewidth]{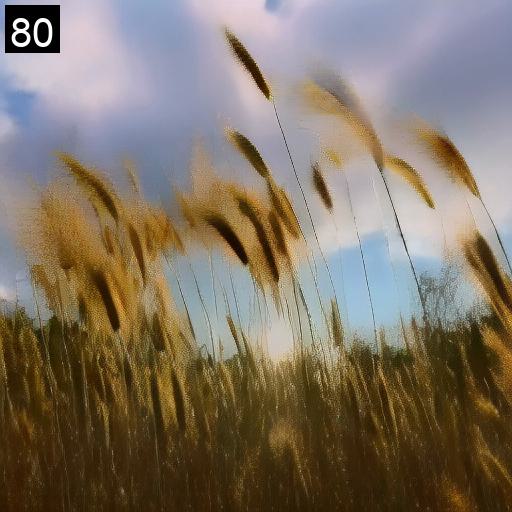} &
        \includegraphics[width=0.2\linewidth]{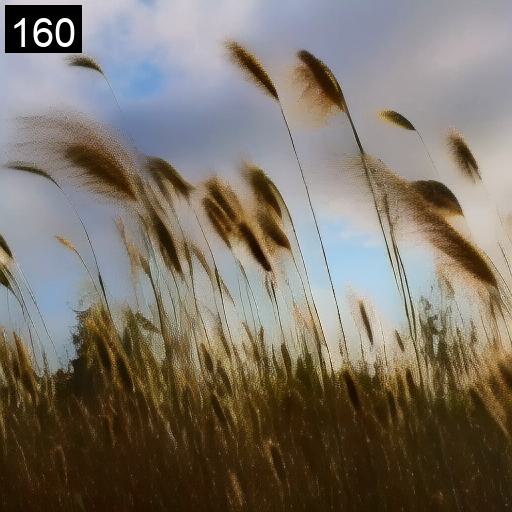} &
        \includegraphics[width=0.2\linewidth]{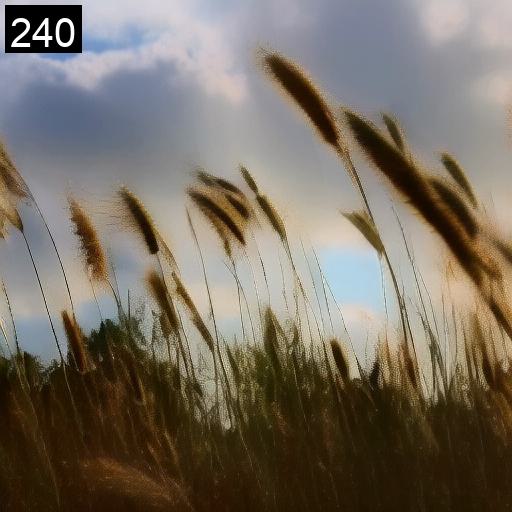} &
        \includegraphics[width=0.2\linewidth]{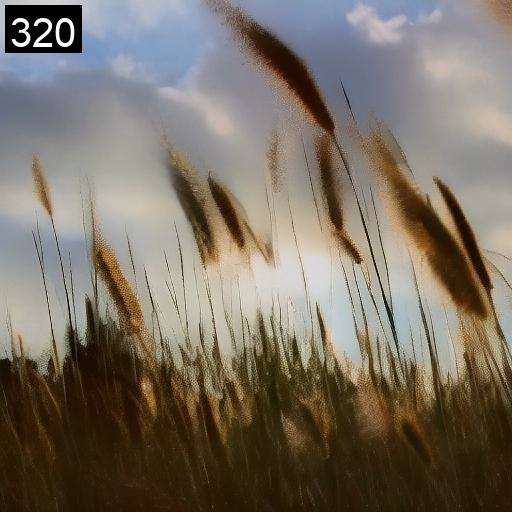} \\
        \multicolumn{5}{c}{\small(e) \textsf{"The dynamic movement of tall, wispy grasses swaying in the wind. The sky above is~$\text{\myldots}$"}} \vspace{2pt}\\
        \includegraphics[width=0.2\linewidth]{fig/opensora_fifo_DDPM_jpgs/6/0.jpg} &
        \includegraphics[width=0.2\linewidth]{fig/opensora_fifo_DDPM_jpgs/6/4.jpg} &
        \includegraphics[width=0.2\linewidth]{fig/opensora_fifo_DDPM_jpgs/6/8.jpg} &
        \includegraphics[width=0.2\linewidth]{fig/opensora_fifo_DDPM_jpgs/6/12.jpg} &
        \includegraphics[width=0.2\linewidth]{fig/opensora_fifo_DDPM_jpgs/6/16.jpg} \\
        \multicolumn{5}{c}{\small(f) \textsf{"a serene winter scene in a forest. The forest is blanketed in a thick layer of snow, which~$\text{\myldots}$"}}
    \end{tabular} 
}
\captionof{figure}{
    Videos generated by FIFO-Diffusion with Open-Sora Plan.
    The number on the top left of each frame indicates the frame index.
    }
\label{fig:qual:opensoraplan_0}

\clearpage
\section{Multi-prompts generation for FIFO-Diffusion}
\label{app:mp}

\subsection{Method}
\label{app:mp_alg}

For multi-prompts generation, we simply change prompts sequentially during the inference.
To be specific, let $\bm{c}_1, \myldots, \bm{c}_k$ be $k$ prompts, and $0=n_0 < n_1<\ldots<n_k$ are increasing sequence of integers.
Then, we use prompt condition $\bm{c}_i$ for $(n_{i-1}+1)^\text{th} \sim n_{i}^\text{th}$ iterations.

\subsection{Qualitative results}
\label{app:mp_qual}

In \cref{fig:mp_1,fig:mp_2}, we provide more qualitative results based on VideoCrafter2.

\scalebox{1}{
    \setlength{\tabcolsep}{1pt}
    \begin{tabular}{ccccc}
        \multicolumn{5}{c}{} \\

        \includegraphics[width=0.2\linewidth]{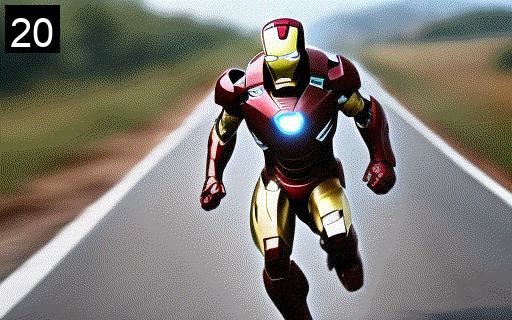} &
        \includegraphics[width=0.2\linewidth]{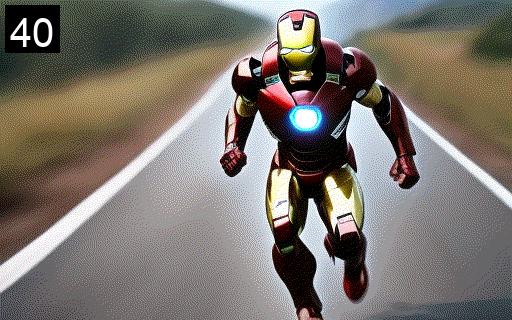} &
        \includegraphics[width=0.2\linewidth]{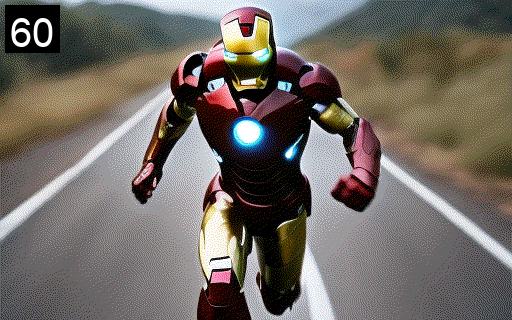} &
        \includegraphics[width=0.2\linewidth]{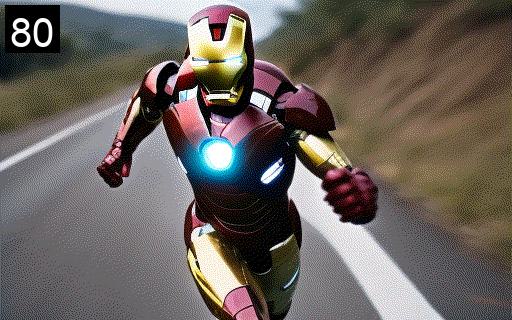} &
        \includegraphics[width=0.2\linewidth]{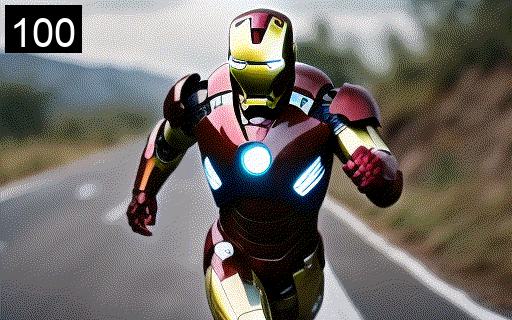} \\
        \includegraphics[width=0.2\linewidth]{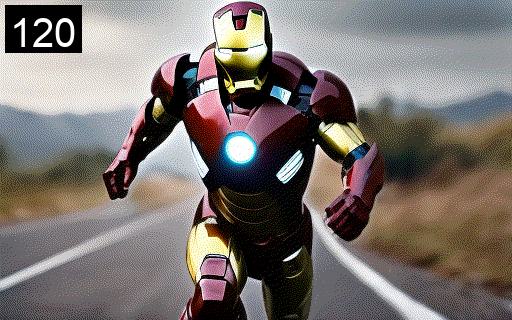} &
        \includegraphics[width=0.2\linewidth]{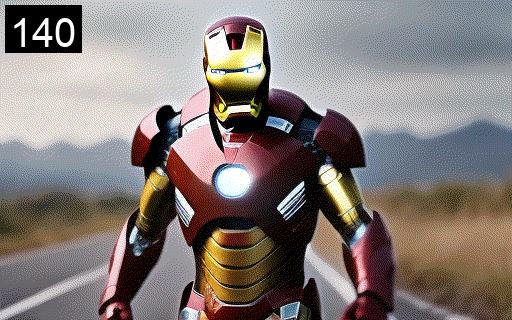} &
        \includegraphics[width=0.2\linewidth]{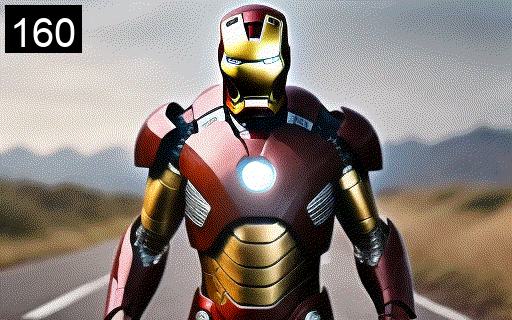} &
        \includegraphics[width=0.2\linewidth]{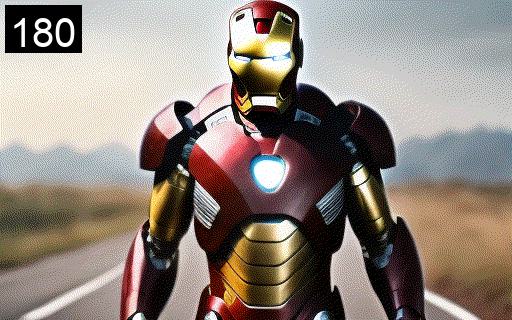} &
         \includegraphics[width=0.2\linewidth]{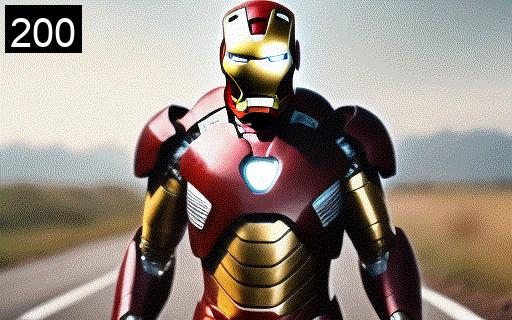} \\
        \includegraphics[width=0.2\linewidth]{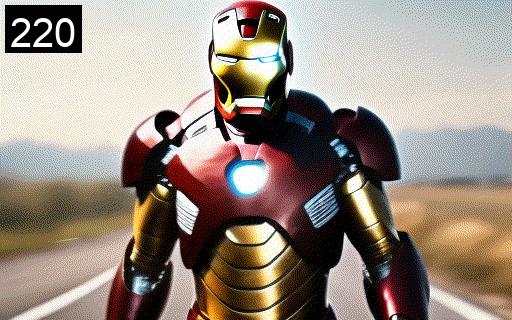} &
        \includegraphics[width=0.2\linewidth]{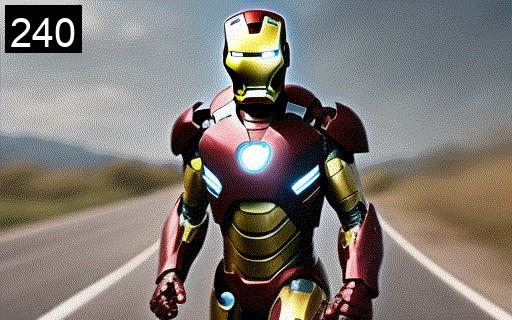} &
        \includegraphics[width=0.2\linewidth]{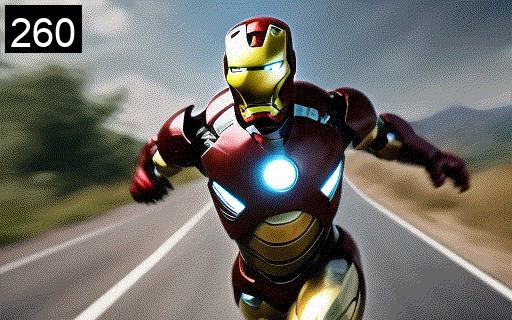} &
        \includegraphics[width=0.2\linewidth]{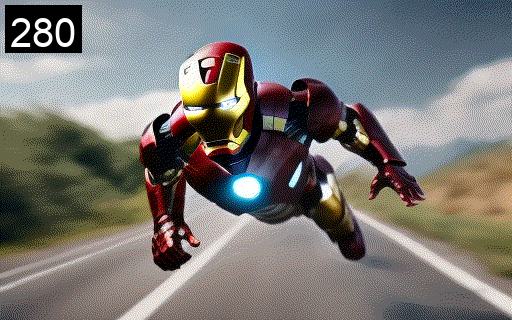} &
        \includegraphics[width=0.2\linewidth]{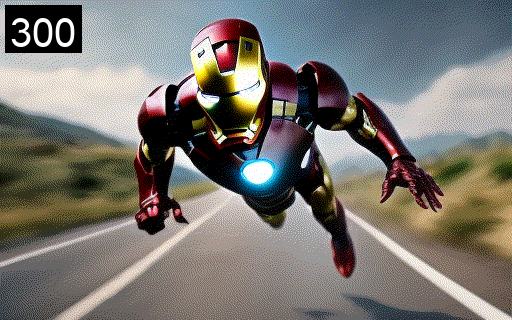} \\
         \multicolumn{5}{c}{(a) \small \textsf{"Ironman \textbf{\textit{running}} $\rightarrow$ \textbf{\textit{standing}} $\rightarrow$ \textbf{\textit{flying}} on the road, 4K, high resolution."}} \vspace{2pt} \\
         
        \includegraphics[width=0.2\linewidth]{fig/mp_jpgs/6/1.jpg} &
        \includegraphics[width=0.2\linewidth]{fig/mp_jpgs/6/2.jpg} &
        \includegraphics[width=0.2\linewidth]{fig/mp_jpgs/6/3.jpg} &
        \includegraphics[width=0.2\linewidth]{fig/mp_jpgs/6/4.jpg} &
        \includegraphics[width=0.2\linewidth]{fig/mp_jpgs/6/5.jpg} \\
        \includegraphics[width=0.2\linewidth]{fig/mp_jpgs/6/6.jpg} &
        \includegraphics[width=0.2\linewidth]{fig/mp_jpgs/6/7.jpg} &
        \includegraphics[width=0.2\linewidth]{fig/mp_jpgs/6/8.jpg} &
        \includegraphics[width=0.2\linewidth]{fig/mp_jpgs/6/9.jpg} &
         \includegraphics[width=0.2\linewidth]{fig/mp_jpgs/6/10.jpg} \\
        \includegraphics[width=0.2\linewidth]{fig/mp_jpgs/6/11.jpg} &
        \includegraphics[width=0.2\linewidth]{fig/mp_jpgs/6/12.jpg} &
        \includegraphics[width=0.2\linewidth]{fig/mp_jpgs/6/13.jpg} &
        \includegraphics[width=0.2\linewidth]{fig/mp_jpgs/6/14.jpg} &
        \includegraphics[width=0.2\linewidth]{fig/mp_jpgs/6/15.jpg} \\
         \multicolumn{5}{c}{(b) \small \textsf{"A tiger  \textbf{\textit{walking}} $\rightarrow$  \textbf{\textit{standing}} $\rightarrow$  \textbf{\textit{resting}} on the grassland, photorealistic, 4k, high definition"}} \vspace{4pt} \\

        \includegraphics[width=0.2\linewidth]{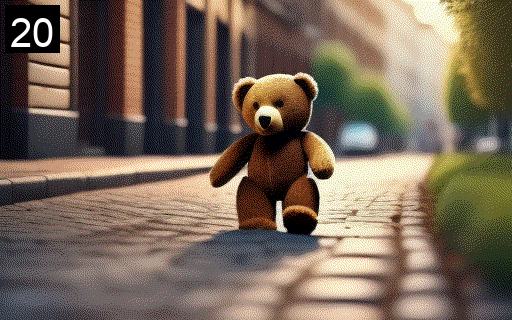} &
        \includegraphics[width=0.2\linewidth]{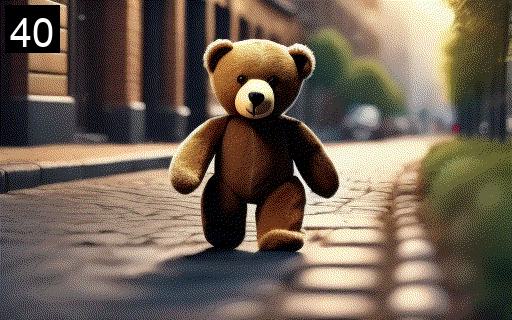} &
        \includegraphics[width=0.2\linewidth]{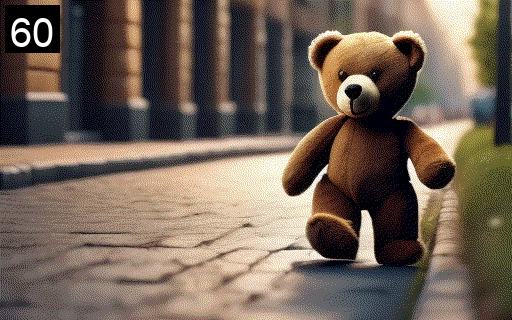} &
        \includegraphics[width=0.2\linewidth]{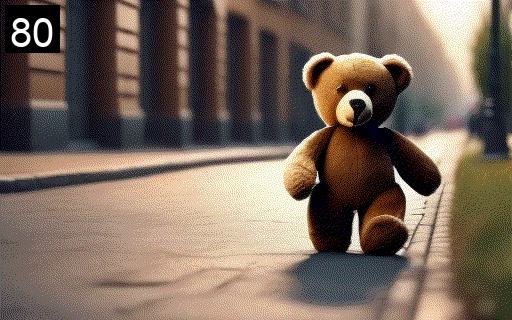} &
        \includegraphics[width=0.2\linewidth]{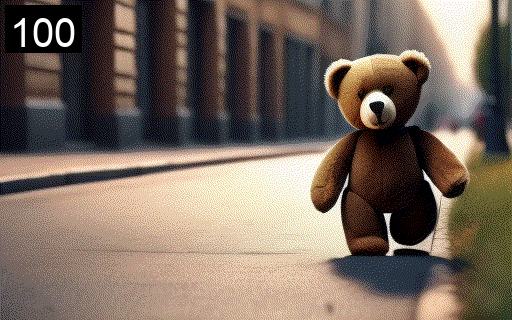} \\
        \includegraphics[width=0.2\linewidth]{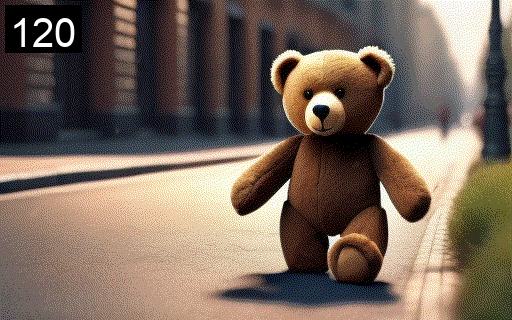} &
        \includegraphics[width=0.2\linewidth]{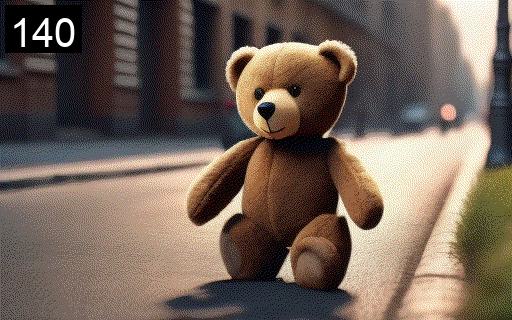} &
        \includegraphics[width=0.2\linewidth]{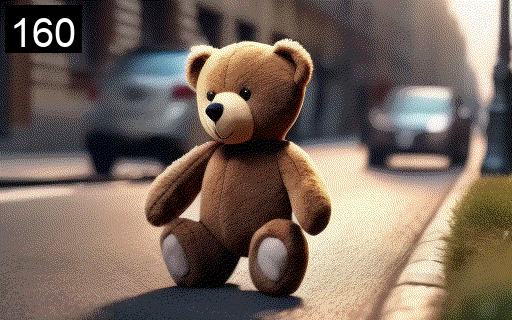} &
        \includegraphics[width=0.2\linewidth]{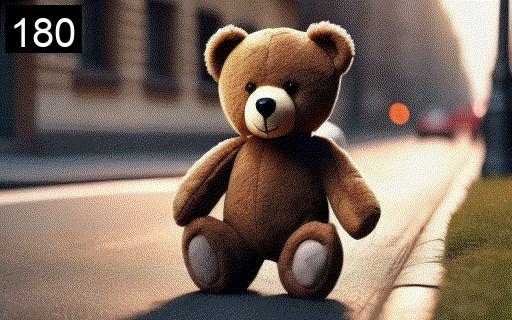} &
         \includegraphics[width=0.2\linewidth]{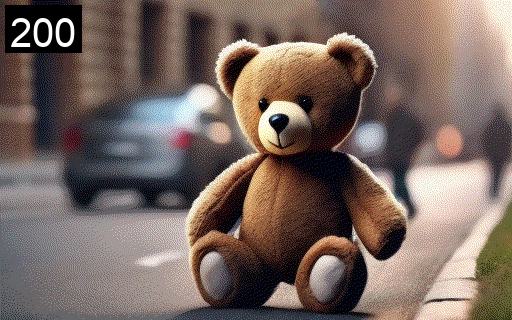} \\
        \includegraphics[width=0.2\linewidth]{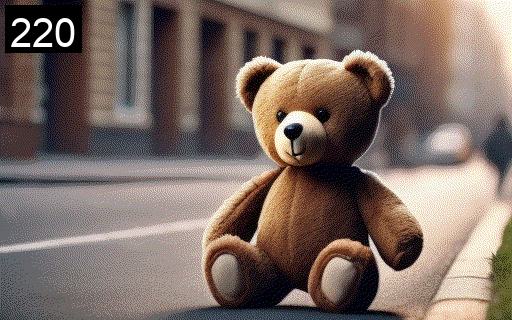} &
        \includegraphics[width=0.2\linewidth]{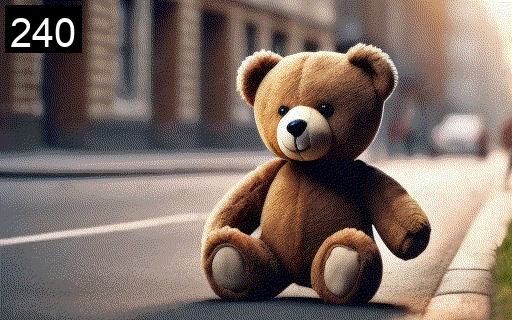} &
        \includegraphics[width=0.2\linewidth]{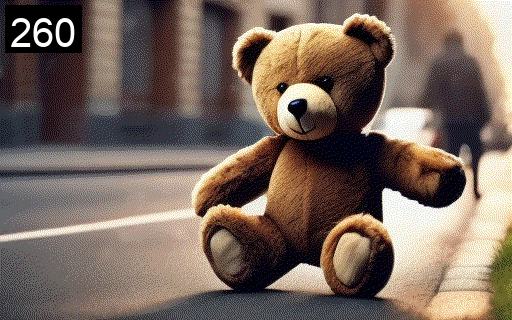} &
        \includegraphics[width=0.2\linewidth]{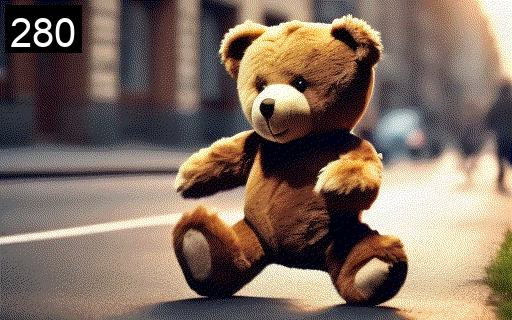} &
         \includegraphics[width=0.2\linewidth]{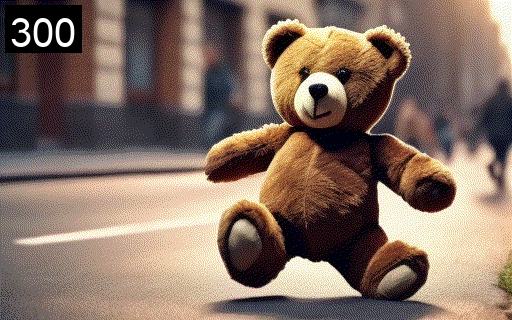} \\
         
         \multicolumn{5}{c}{(c) \small \textsf{"A teddy bear  \textbf{\textit{walking}} $\rightarrow$  \textbf{\textit{standing}} $\rightarrow$  \textbf{\textit{dancing}} on the street, 4K, high resolution."}} \vspace{2pt} \\

    \end{tabular}
}
\captionof{figure}{
    Videos generated by FIFO-Diffusion with three prompts.
    The number on the top left of each frame indicates the frame index.
    }
\label{fig:mp_1}

\clearpage
\scalebox{1}{
    \setlength{\tabcolsep}{1pt}
    \begin{tabular}{ccccc}
        \multicolumn{5}{c}{} \\
        \includegraphics[width=0.2\linewidth]{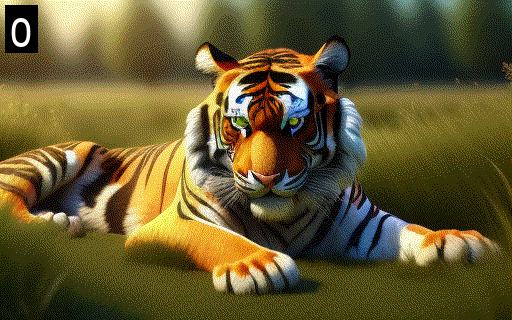} &
        \includegraphics[width=0.2\linewidth]{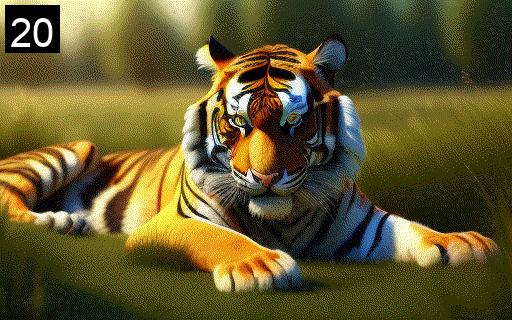} &
        \includegraphics[width=0.2\linewidth]{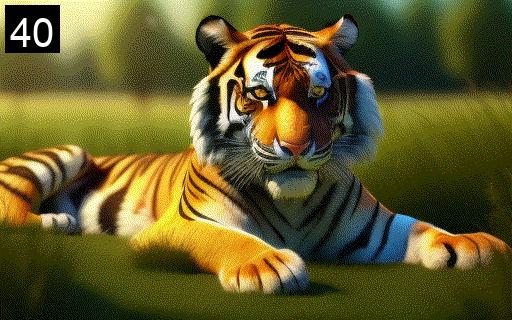} &
        \includegraphics[width=0.2\linewidth]{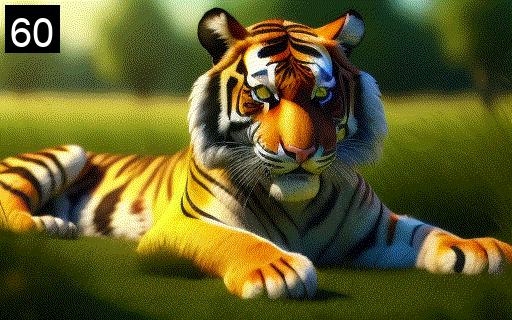} &
        \includegraphics[width=0.2\linewidth]{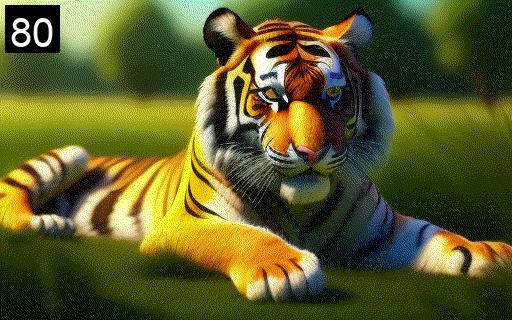} \\
        \includegraphics[width=0.2\linewidth]{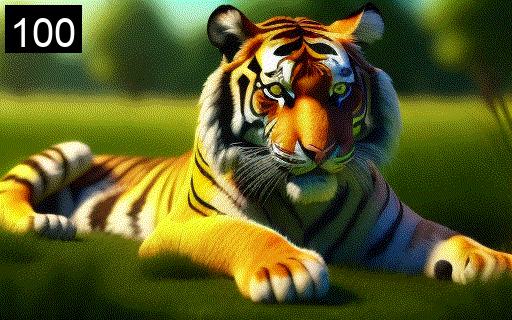} &
        \includegraphics[width=0.2\linewidth]{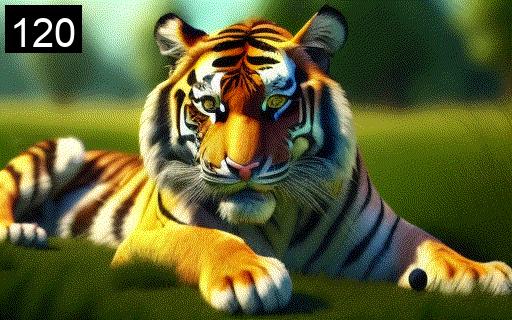} &
        \includegraphics[width=0.2\linewidth]{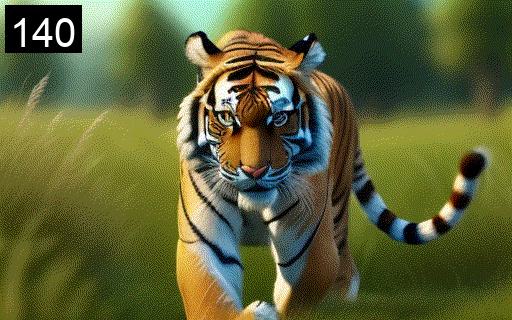} &
        \includegraphics[width=0.2\linewidth]{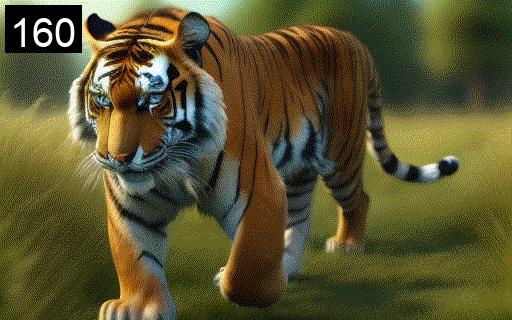} &
        \includegraphics[width=0.2\linewidth]{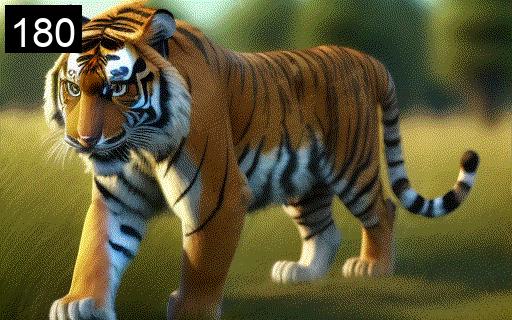} \\
         \multicolumn{5}{c}{\small \textsf{(a) "A tiger  \textbf{\textit{resting}} $\rightarrow$  \textbf{\textit{walking}} on the grassland, photorealistic, 4k, high definition"}} \vspace{4pt} \\
         \includegraphics[width=0.2\linewidth]{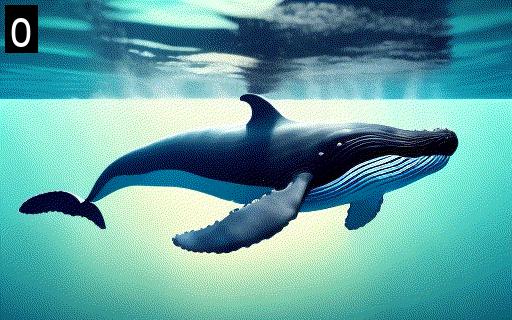} &
        \includegraphics[width=0.2\linewidth]{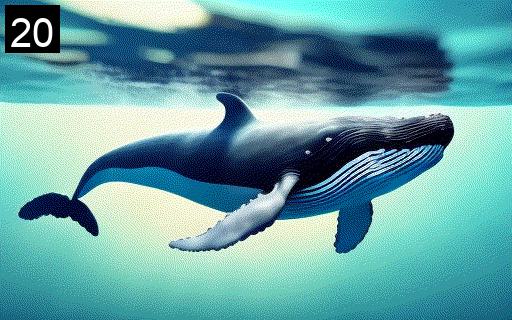} &
        \includegraphics[width=0.2\linewidth]{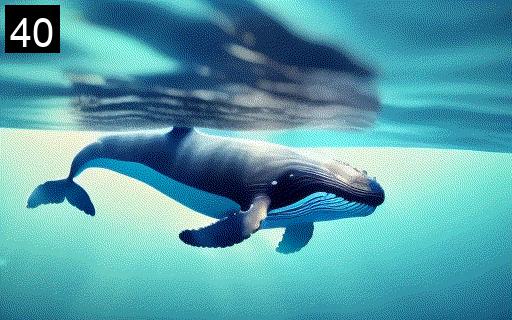} &
        \includegraphics[width=0.2\linewidth]{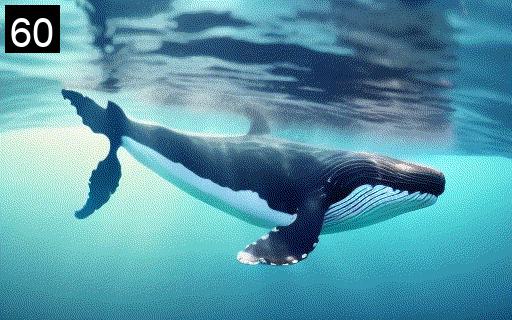} &
        \includegraphics[width=0.2\linewidth]{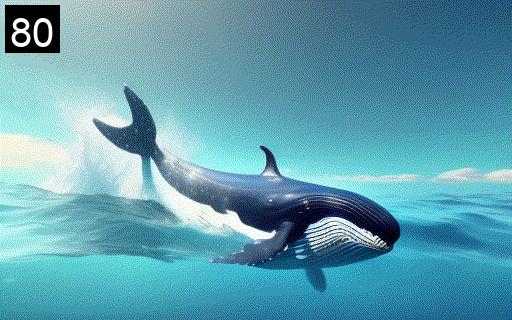} \\
        \includegraphics[width=0.2\linewidth]{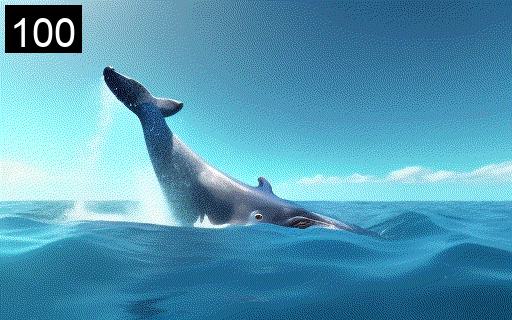} &
        \includegraphics[width=0.2\linewidth]{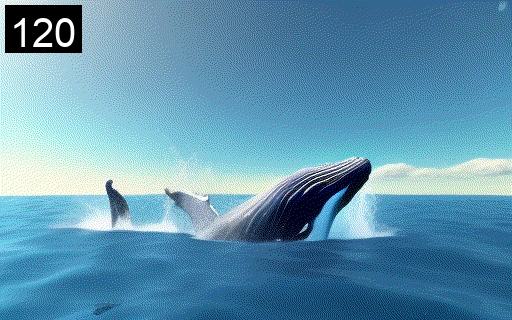} &
        \includegraphics[width=0.2\linewidth]{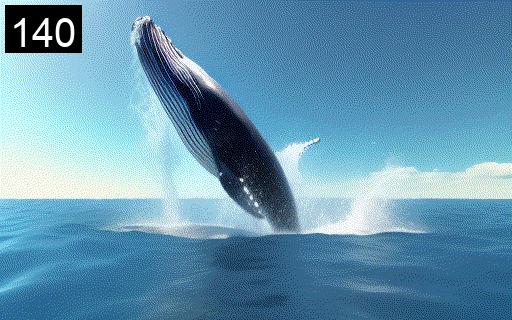} &
        \includegraphics[width=0.2\linewidth]{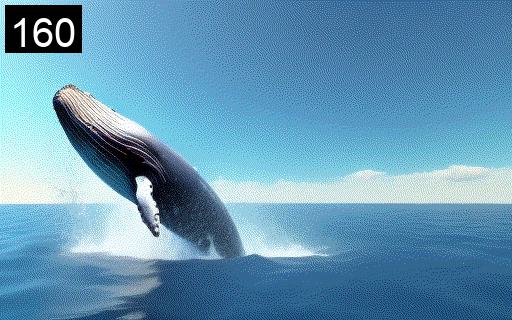} &
        \includegraphics[width=0.2\linewidth]{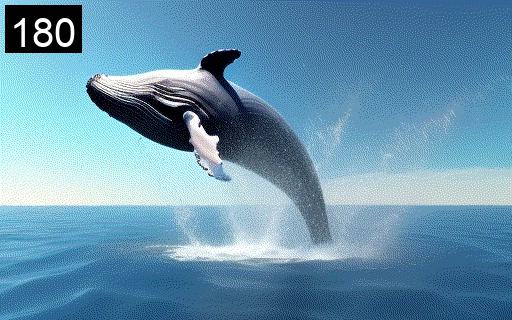} \\
        \multicolumn{5}{c}{\small \textsf{(b) "A whale  \textbf{\textit{swimming on the surface of the ocean}} $\rightarrow$  \textbf{\textit{jumps out of water}}, 4K, high resolution."}}
        \vspace{4pt} \\
         \includegraphics[width=0.2\linewidth]{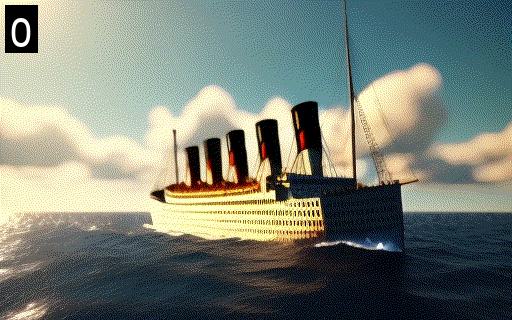} &
        \includegraphics[width=0.2\linewidth]{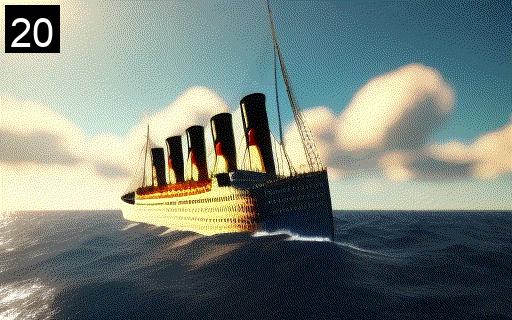} &
        \includegraphics[width=0.2\linewidth]{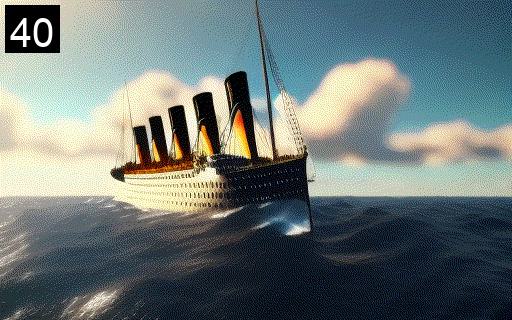} &
        \includegraphics[width=0.2\linewidth]{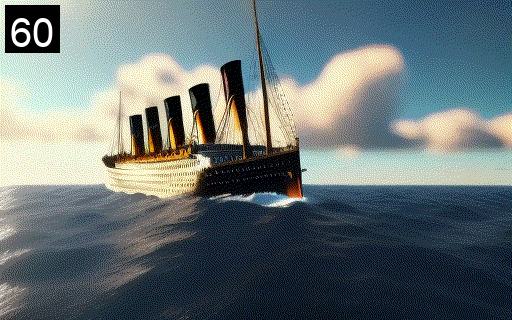} &
        \includegraphics[width=0.2\linewidth]{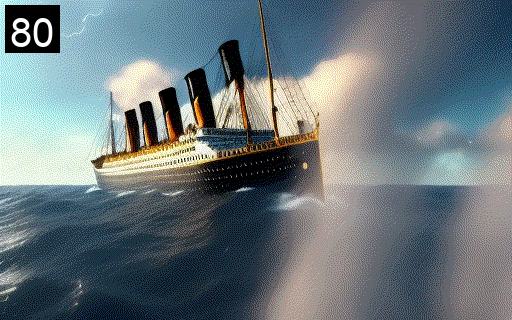} \\
        \includegraphics[width=0.2\linewidth]{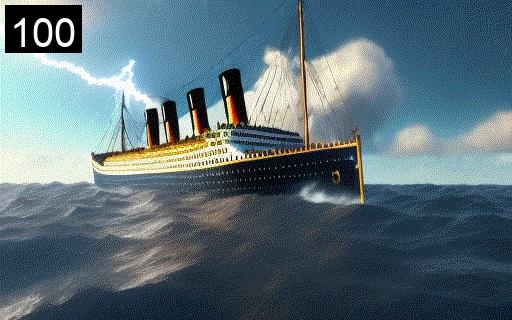} &
        \includegraphics[width=0.2\linewidth]{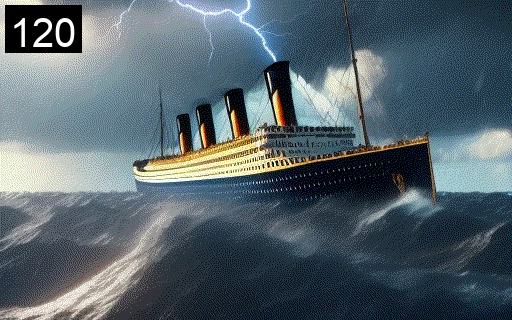} &
        \includegraphics[width=0.2\linewidth]{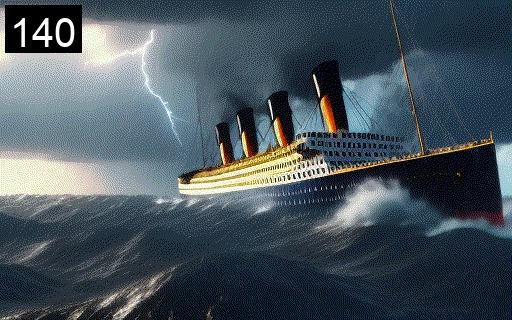} &
        \includegraphics[width=0.2\linewidth]{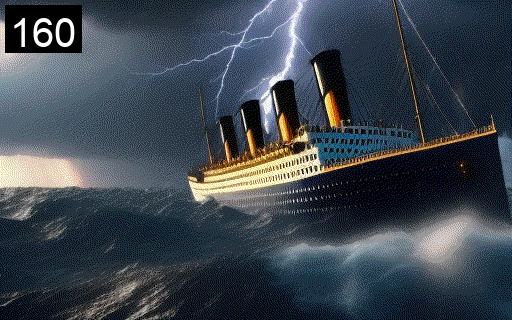} &
        \includegraphics[width=0.2\linewidth]{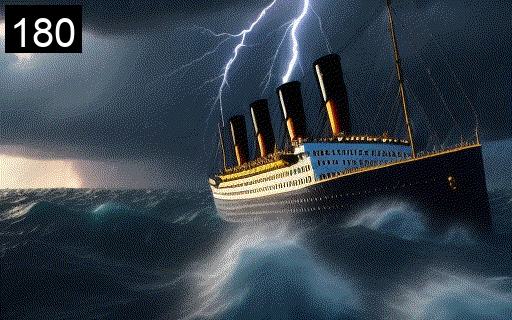} \\
         \multicolumn{5}{c}{(c) \scriptsize \textsf{"Titanic sailing through  \textbf{\textit{the sunny calm ocean}} $\rightarrow$  \textbf{\textit{a stormy ocean with lightning}}, 4K, high resolution."}}\vspace{4pt} \\
         \includegraphics[width=0.2\linewidth]{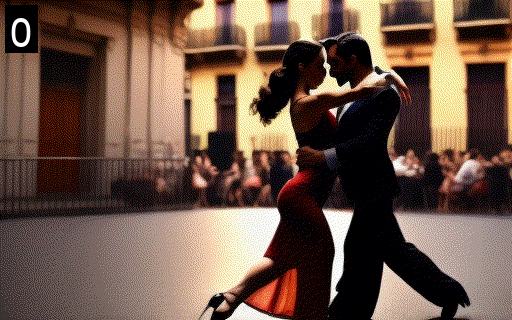} &
        \includegraphics[width=0.2\linewidth]{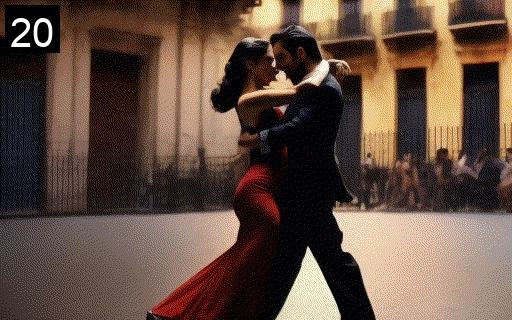} &
        \includegraphics[width=0.2\linewidth]{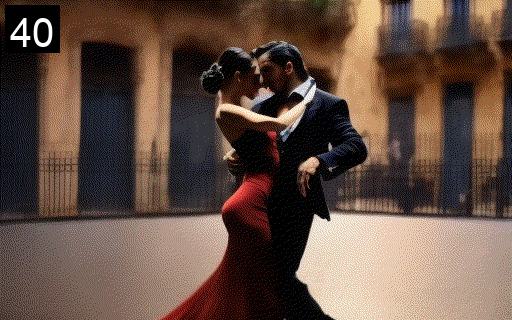} &
        \includegraphics[width=0.2\linewidth]{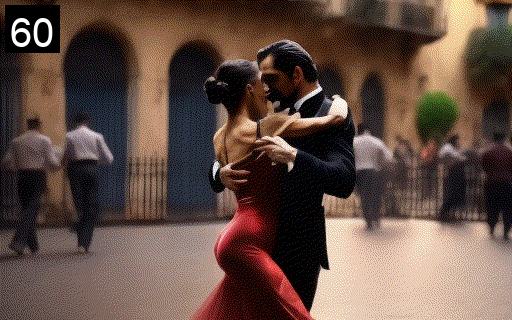} &
        \includegraphics[width=0.2\linewidth]{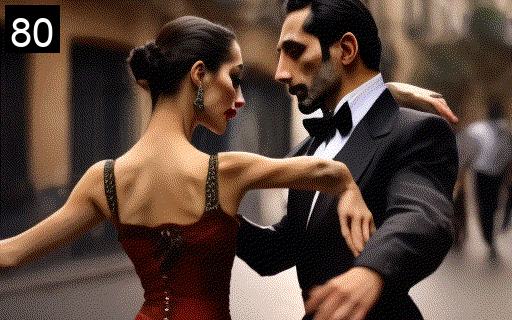} \\
        \includegraphics[width=0.2\linewidth]{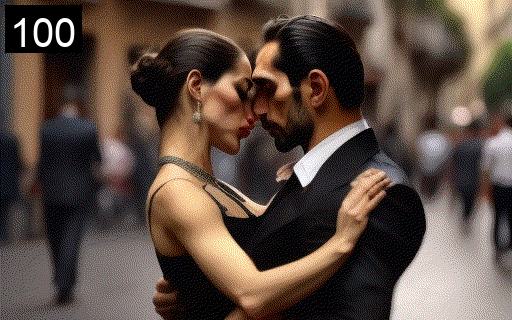} &
        \includegraphics[width=0.2\linewidth]{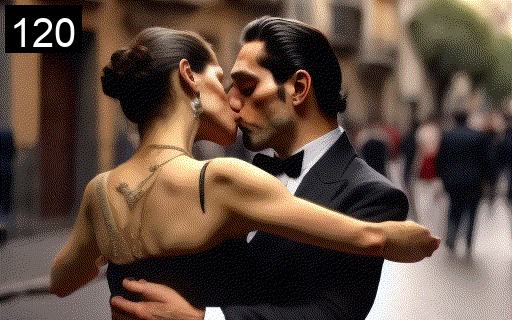} &
        \includegraphics[width=0.2\linewidth]{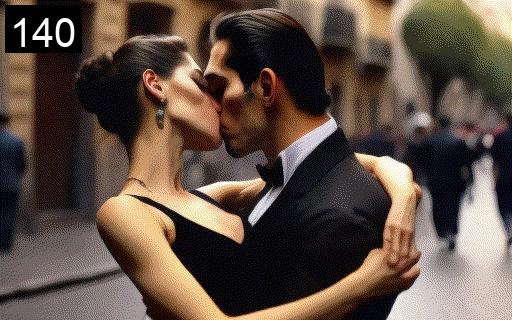} &
        \includegraphics[width=0.2\linewidth]{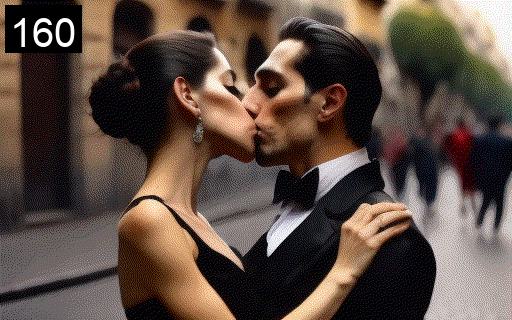} &
        \includegraphics[width=0.2\linewidth]{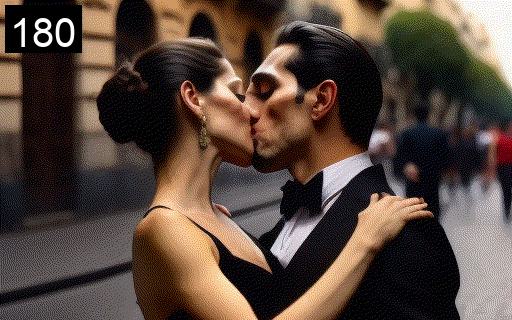} \\
         \multicolumn{5}{c}{(d) \small \textsf{"A pair of tango dancers  \textbf{\textit{performing}} $\rightarrow$ \textbf{ \textit{kissing}} in Buenos Aires, 4K, high resolution."}} 
        
    \end{tabular}
}
\captionof{figure}{
    Videos generated by FIFO-Diffusion with two prompts.
    The number on the top left of each frame indicates the frame index.
    }
\label{fig:mp_2}

%%%%%%%%%%%%%%%%%%%%%%%%%%%%%%%%%%%%%%%%%%%%%%%%%%%%%%%%%%%%%%%%%%%%%%%%%%%%%%%%%%%%%%%%%%%%%%%%%%
\clearpage
\section{Qualitative comparisons with other long video generation methods}
\label{app:qual_comparison}
\begin{figure*}[h]
    \centering
    \renewcommand{\arraystretch}{0.7}
    \scalebox{0.98}{
    \setlength{\tabcolsep}{1pt}
    \hspace{-3mm}
    \begin{tabular}{cccccc}
        \rotatebox[origin=c]{90}{\small Ours\hspace{-14mm}} &
        \includegraphics[width=0.2\linewidth]{fig/videocrafter2_jpgs/17/0.jpg} &
        \includegraphics[width=0.2\linewidth]{fig/videocrafter2_jpgs/17/1.jpg} &
        \includegraphics[width=0.2\linewidth]{fig/videocrafter2_jpgs/17/2.jpg} &
        \includegraphics[width=0.2\linewidth]{fig/videocrafter2_jpgs/17/3.jpg} &
        \includegraphics[width=0.2\linewidth]{fig/videocrafter2_jpgs/17/4.jpg} \vspace{1mm} \\
        % \shortstack[]{FreeNoise\\\vspace{8mm}} &
        \rotatebox[origin=c]{90}{\small FreeNoise\hspace{-14mm}} &
        \includegraphics[width=0.2\linewidth]{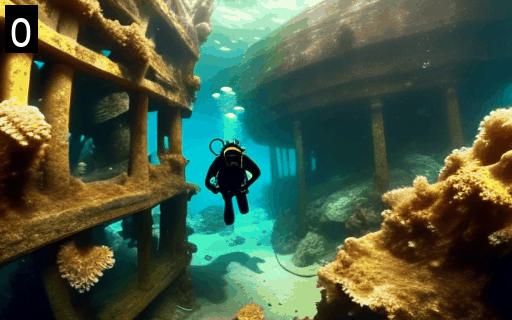} &
        \includegraphics[width=0.2\linewidth]{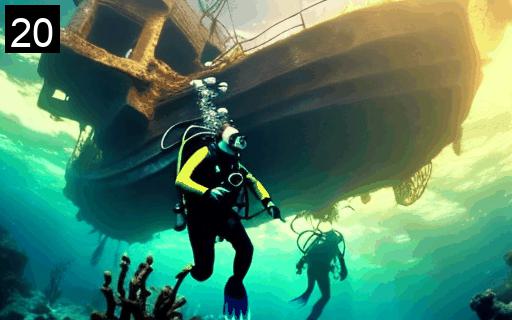} &
        \includegraphics[width=0.2\linewidth]{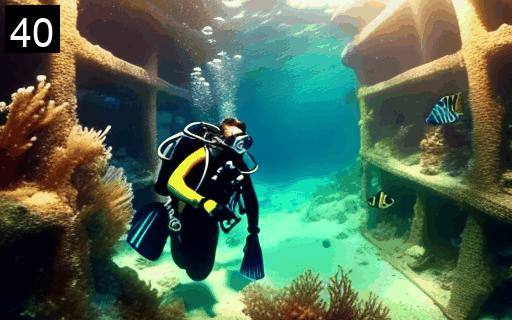} &
        \includegraphics[width=0.2\linewidth]{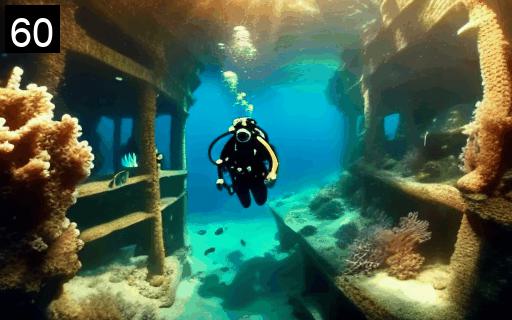} &
        \includegraphics[width=0.2\linewidth]{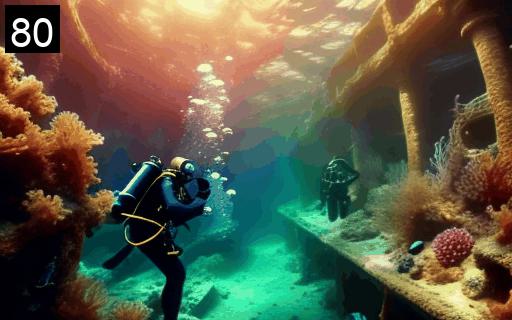} \vspace{1mm}\\
        % \shortstack[]{FIFO-\\Diffusion\vspace{8mm}} &
        \rotatebox[origin=c]{90}{\small Gen-L-Video\hspace{-14mm}} &
        \includegraphics[width=0.2\linewidth]{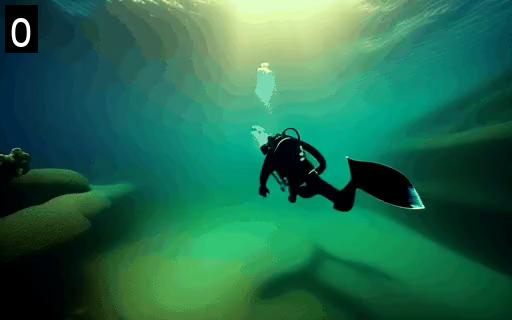} &
        \includegraphics[width=0.2\linewidth]{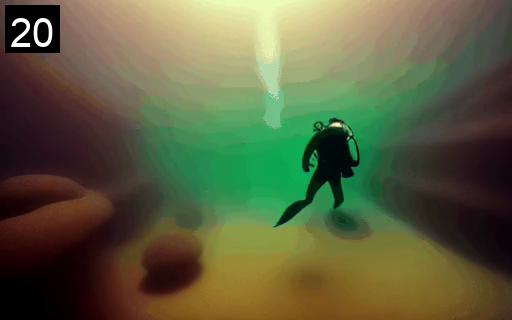} &
        \includegraphics[width=0.2\linewidth]{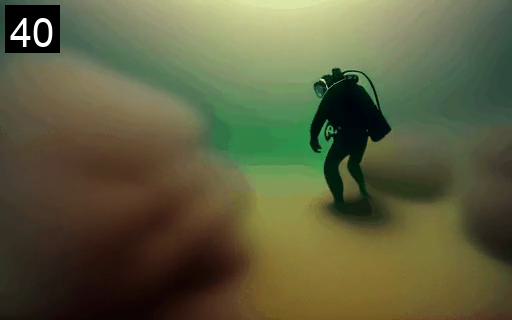} &
        \includegraphics[width=0.2\linewidth]{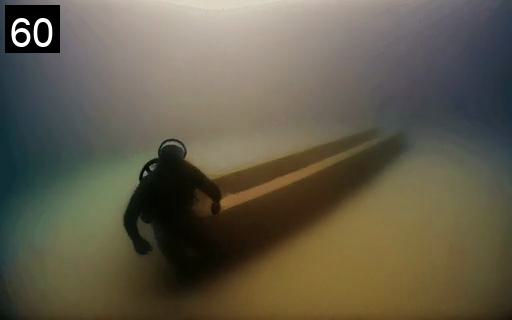} &
        \includegraphics[width=0.2\linewidth]{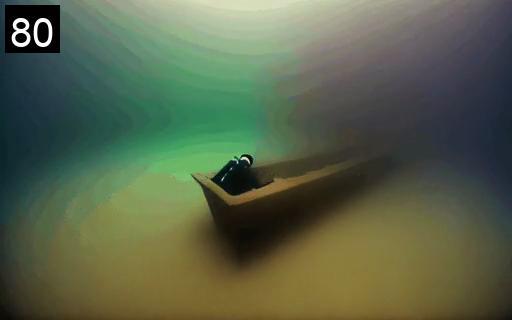} \vspace{1mm} \\
        % \shortstack[]{FIFO-\\Diffusion\vspace{8mm}} &
        \rotatebox[origin=c]{90}{\small LaVie+SEINE\hspace{-14mm}} &
        \includegraphics[width=0.2\linewidth]{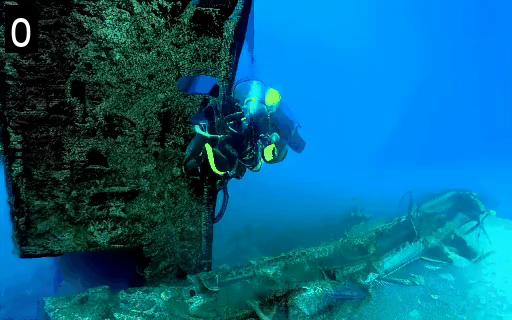} &
        \includegraphics[width=0.2\linewidth]{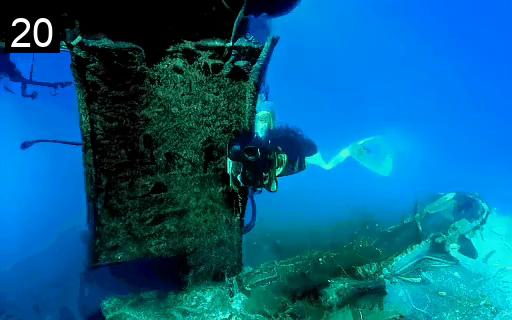} &
        \includegraphics[width=0.2\linewidth]{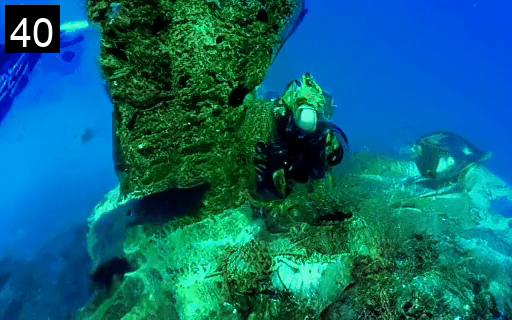} &
        \includegraphics[width=0.2\linewidth]{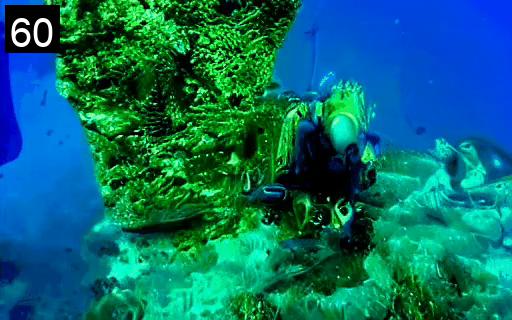} &
        \includegraphics[width=0.2\linewidth]{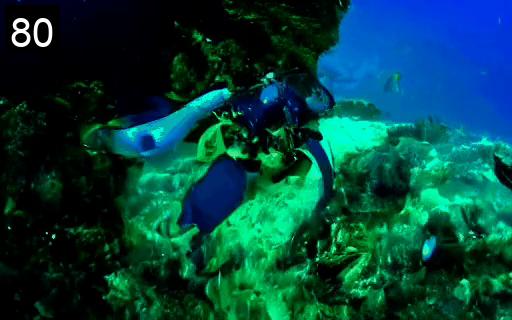} \vspace{1mm} \\
        & \multicolumn{5}{c}{\small (a) \textsf{"A vibrant underwater scene of a scuba diver exploring a shipwreck, 2K, photorealistic."}} \vspace{3pt}\\
        % \shortstack[]{FIFO-\\Diffusion\vspace{8mm}} &
	\rotatebox[origin=c]{90}{\small Ours\hspace{-14mm}} &
        \includegraphics[width=0.2\linewidth]{fig/videocrafter2_jpgs/9/0.jpg} &
        \includegraphics[width=0.2\linewidth]{fig/videocrafter2_jpgs/9/1.jpg} &
        \includegraphics[width=0.2\linewidth]{fig/videocrafter2_jpgs/9/2.jpg} &
        \includegraphics[width=0.2\linewidth]{fig/videocrafter2_jpgs/9/3.jpg} &
        \includegraphics[width=0.2\linewidth]{fig/videocrafter2_jpgs/9/4.jpg} \vspace{1mm} \\
        % \shortstack[]{FreeNoise\\\vspace{8mm}} &
        \rotatebox[origin=c]{90}{\small FreeNoise\hspace{-14mm}} &
        \includegraphics[width=0.2\linewidth]{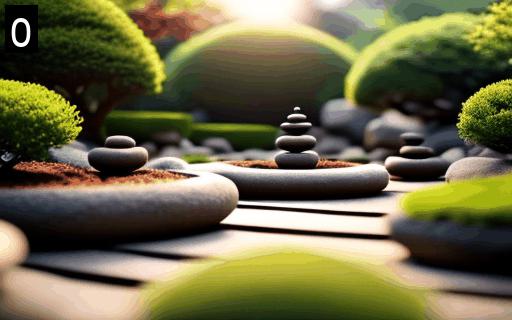} &
        \includegraphics[width=0.2\linewidth]{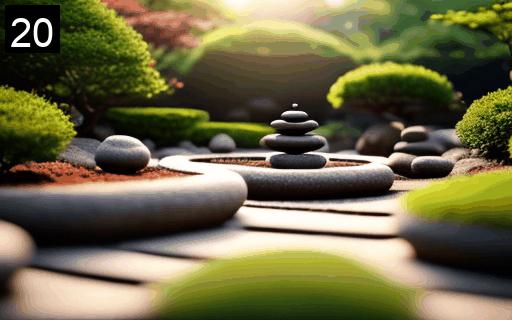} &
        \includegraphics[width=0.2\linewidth]{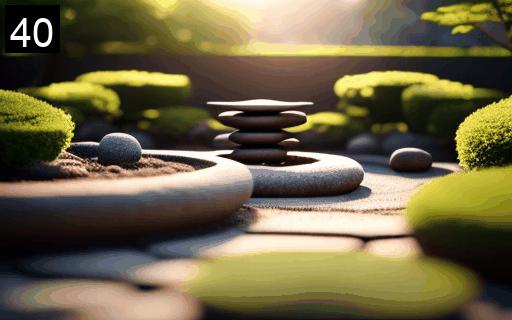} &
        \includegraphics[width=0.2\linewidth]{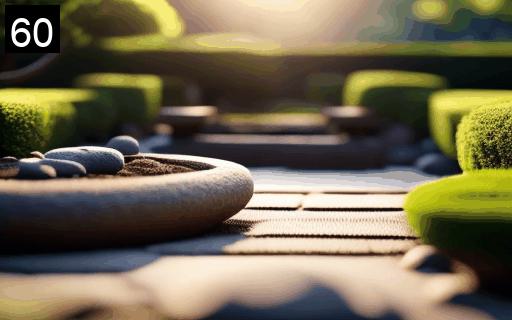} &
        \includegraphics[width=0.2\linewidth]{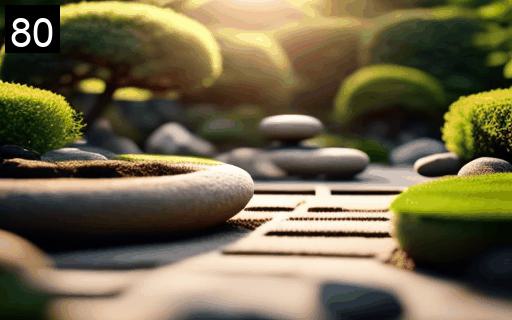} \vspace{1mm}\\
        % \shortstack[]{FIFO-\\Diffusion\vspace{8mm}} &
        \rotatebox[origin=c]{90}{\small Gen-L-Video\hspace{-14mm}} &
        \includegraphics[width=0.2\linewidth]{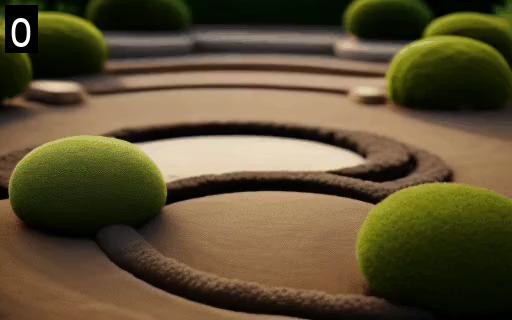} &
        \includegraphics[width=0.2\linewidth]{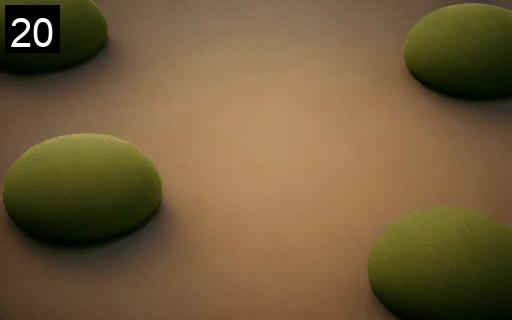} &
        \includegraphics[width=0.2\linewidth]{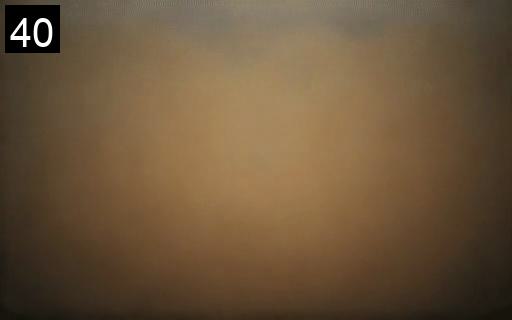} &
        \includegraphics[width=0.2\linewidth]{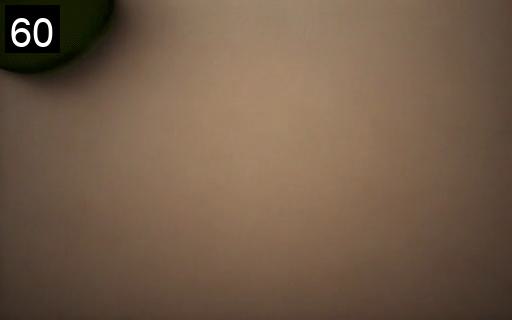} &
        \includegraphics[width=0.2\linewidth]{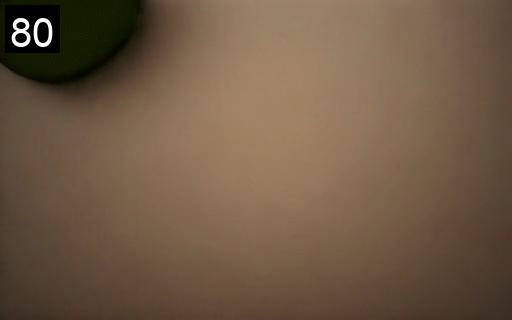} \vspace{1mm} \\
        % \shortstack[]{FIFO-\\Diffusion\vspace{8mm}} &
        \rotatebox[origin=c]{90}{\small LaVie+SEINE\hspace{-14mm}} &
        \includegraphics[width=0.2\linewidth]{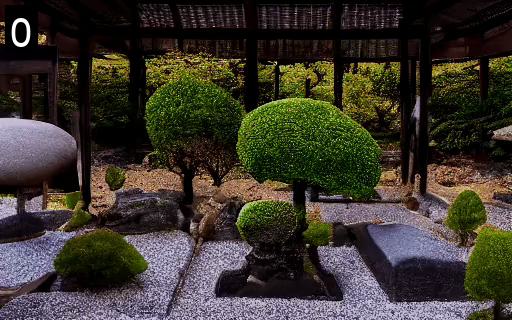} &
        \includegraphics[width=0.2\linewidth]{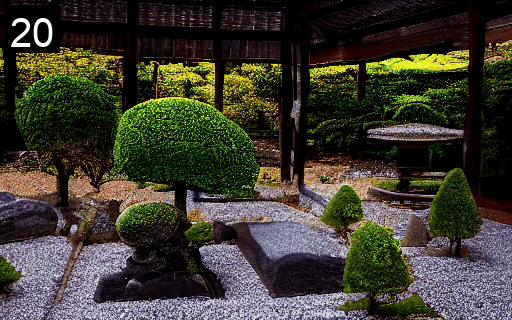} &
        \includegraphics[width=0.2\linewidth]{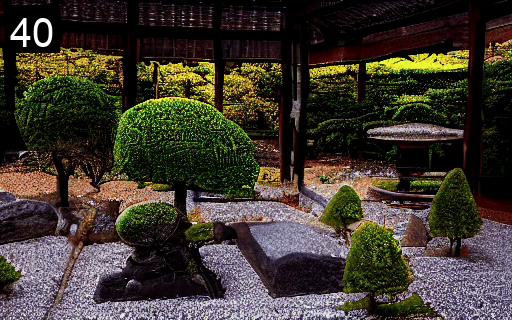} &
        \includegraphics[width=0.2\linewidth]{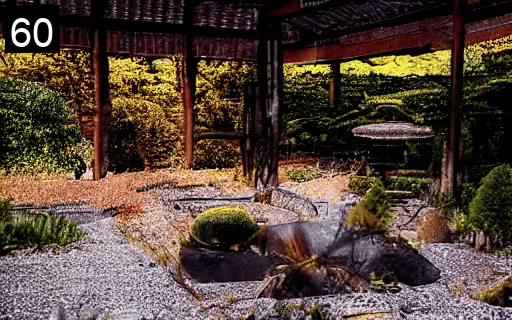} &
        \includegraphics[width=0.2\linewidth]{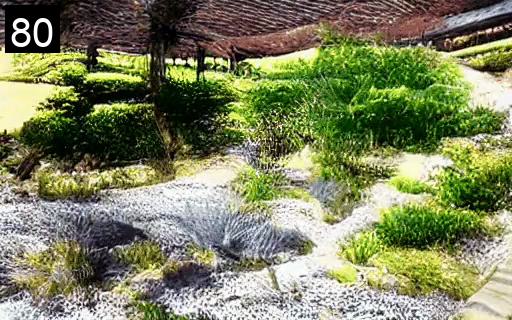} \vspace{1mm} \\
        & \multicolumn{5}{c}{\small (b) \textsf{"A panoramic view of a peaceful Zen garden, high-quality, 4K resolution."}} \vspace{6mm} \\
        
    \end{tabular}
    }
    \caption{
        Qualitative comparisons with other long video generation techniques, Gen-L-Video, FreeNoise, and LaVie + SEINE.
        The number in the top-left corner of each frame indicates the frame index.
    }\label{fig:qual_comparison_app1}
\end{figure*}

\begin{figure*}[h]
    \centering
    \renewcommand{\arraystretch}{0.7}
    \scalebox{0.98}{
    \setlength{\tabcolsep}{1pt}
    \hspace{-3mm}
    \begin{tabular}{cccccc}
    
	\rotatebox[origin=c]{90}{\small Ours\hspace{-14mm}} &
        \includegraphics[width=0.2\linewidth]{fig/videocrafter2_jpgs/8/0.jpg} &
        \includegraphics[width=0.2\linewidth]{fig/videocrafter2_jpgs/8/1.jpg} &
        \includegraphics[width=0.2\linewidth]{fig/videocrafter2_jpgs/8/2.jpg} &
        \includegraphics[width=0.2\linewidth]{fig/videocrafter2_jpgs/8/3.jpg} &
        \includegraphics[width=0.2\linewidth]{fig/videocrafter2_jpgs/8/4.jpg} \vspace{1mm} \\
        % \shortstack[]{FreeNoise\\\vspace{8mm}} &
        \rotatebox[origin=c]{90}{\small FreeNoise\hspace{-14mm}} &
        \includegraphics[width=0.2\linewidth]{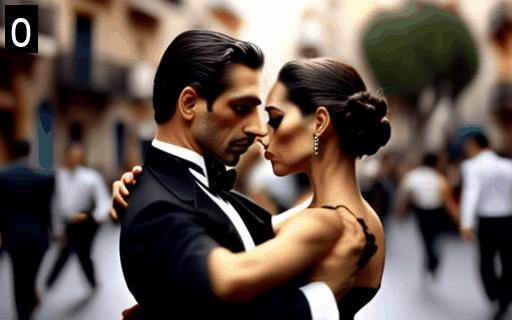} &
        \includegraphics[width=0.2\linewidth]{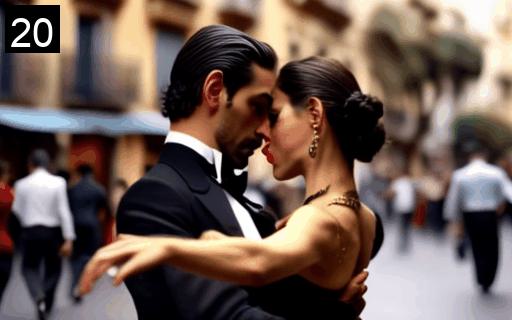} &
        \includegraphics[width=0.2\linewidth]{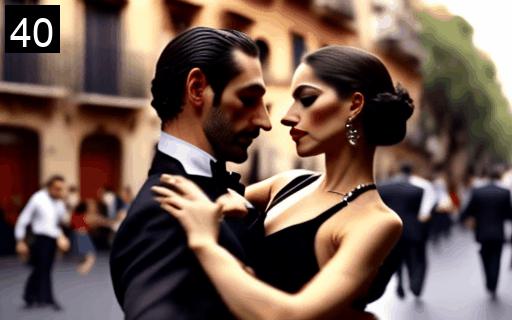} &
        \includegraphics[width=0.2\linewidth]{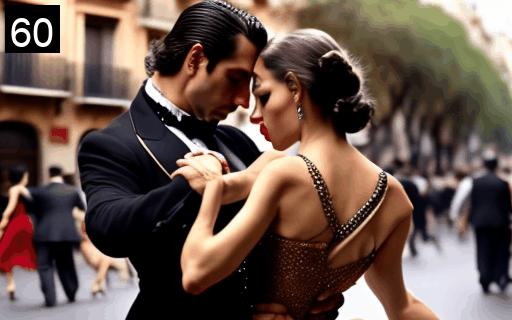} &
        \includegraphics[width=0.2\linewidth]{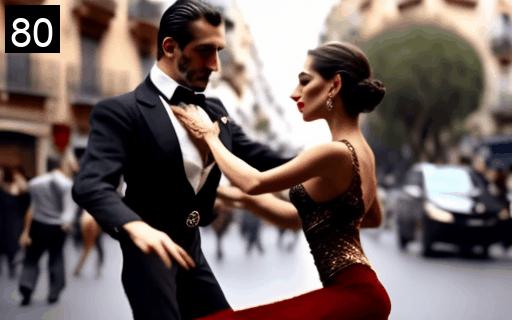} \vspace{1mm}\\
        % \shortstack[]{FIFO-\\Diffusion\vspace{8mm}} &
        \rotatebox[origin=c]{90}{\small Gen-L-Video\hspace{-14mm}} &
        \includegraphics[width=0.2\linewidth]{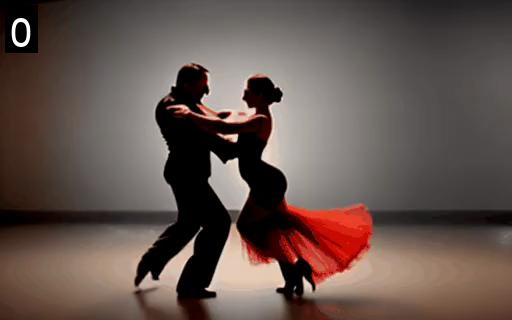} &
        \includegraphics[width=0.2\linewidth]{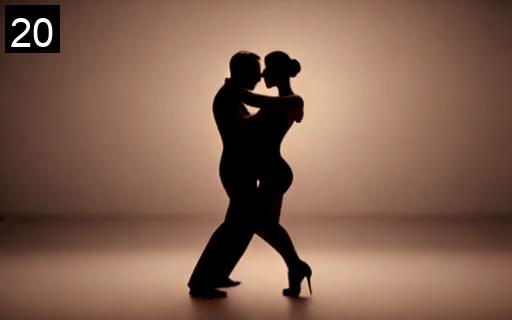} &
        \includegraphics[width=0.2\linewidth]{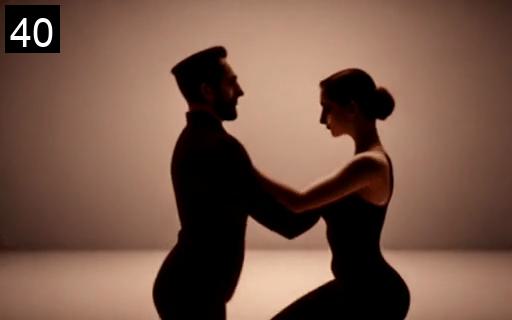} &
        \includegraphics[width=0.2\linewidth]{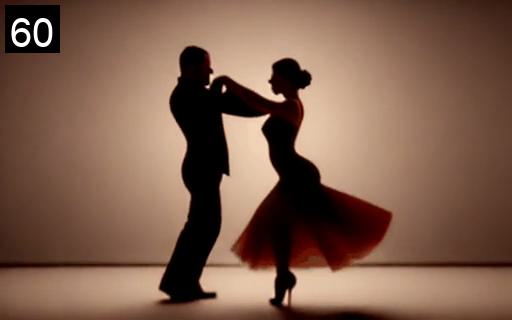} &
        \includegraphics[width=0.2\linewidth]{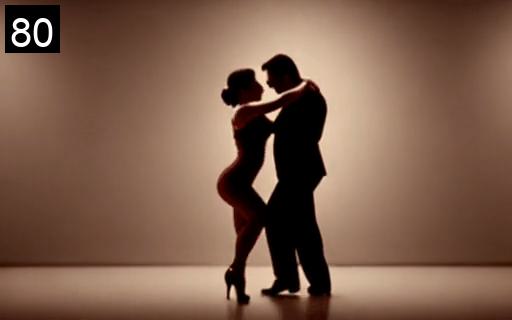} \vspace{1mm} \\
        % \shortstack[]{FIFO-\\Diffusion\vspace{8mm}} &
        \rotatebox[origin=c]{90}{\small LaVie+SEINE\hspace{-14mm}} &
        \includegraphics[width=0.2\linewidth]{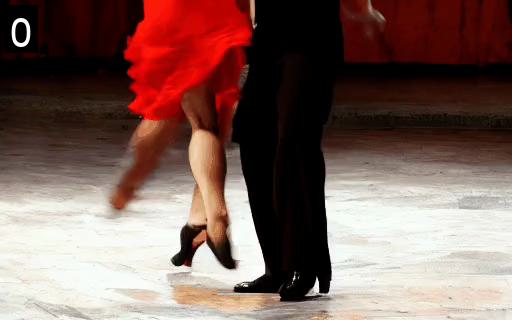} &
        \includegraphics[width=0.2\linewidth]{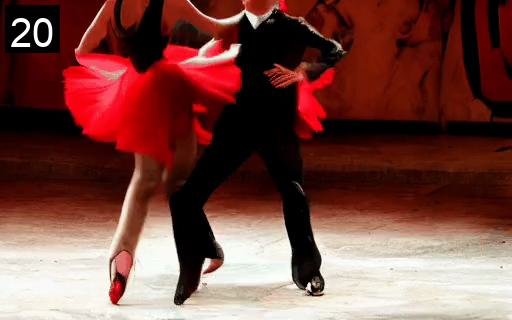} &
        \includegraphics[width=0.2\linewidth]{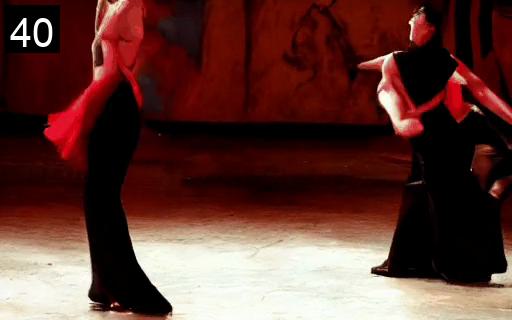} &
        \includegraphics[width=0.2\linewidth]{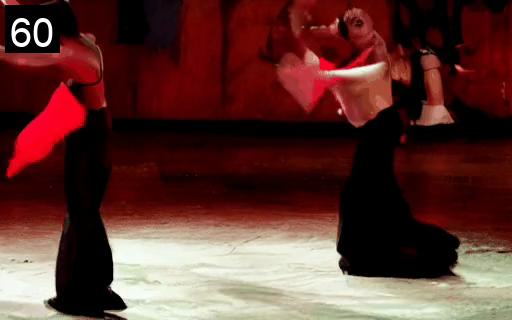} &
        \includegraphics[width=0.2\linewidth]{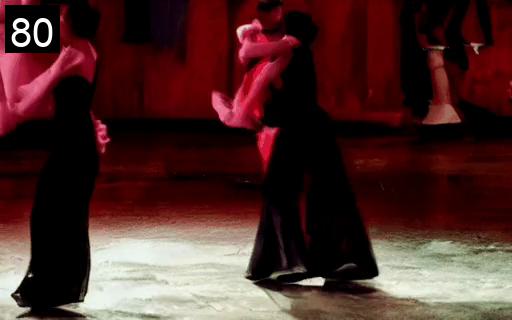} \vspace{1mm} \\
        & \multicolumn{5}{c}{\small (a) \textsf{"A pair of tango dancers performing in Buenos Aires, 4K, high resolution."}} \vspace{6mm}\\
        % \shortstack[]{FIFO-\\Diffusion\vspace{8mm}} &
        
        \rotatebox[origin=c]{90}{\small Ours\hspace{-14mm}} &
        \includegraphics[width=0.2\linewidth]{fig/videocrafter2_jpgs/15/0.jpg} &
        \includegraphics[width=0.2\linewidth]{fig/videocrafter2_jpgs/15/1.jpg} &
        \includegraphics[width=0.2\linewidth]{fig/videocrafter2_jpgs/15/2.jpg} &
        \includegraphics[width=0.2\linewidth]{fig/videocrafter2_jpgs/15/3.jpg} &
        \includegraphics[width=0.2\linewidth]{fig/videocrafter2_jpgs/15/4.jpg} \vspace{1mm} \\
        % \shortstack[]{FreeNoise\\\vspace{8mm}} &
        \rotatebox[origin=c]{90}{\small FreeNoise\hspace{-14mm}} &
        \includegraphics[width=0.2\linewidth]{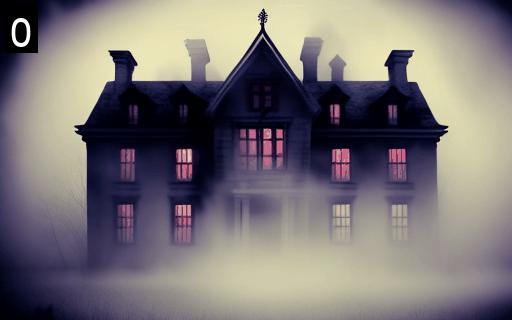} &
        \includegraphics[width=0.2\linewidth]{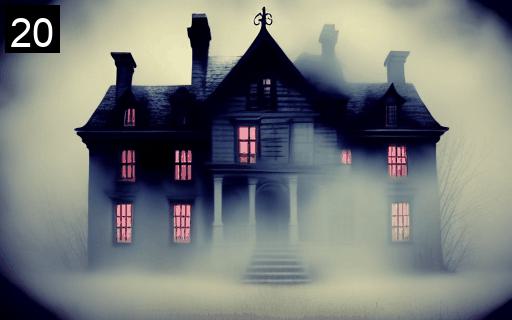} &
        \includegraphics[width=0.2\linewidth]{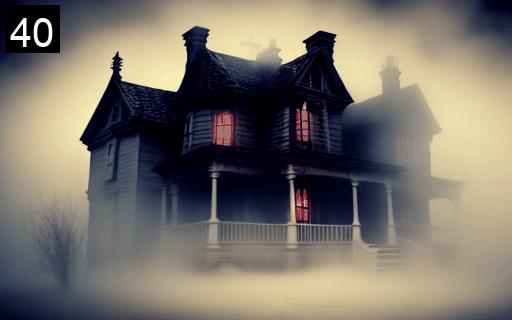} &
        \includegraphics[width=0.2\linewidth]{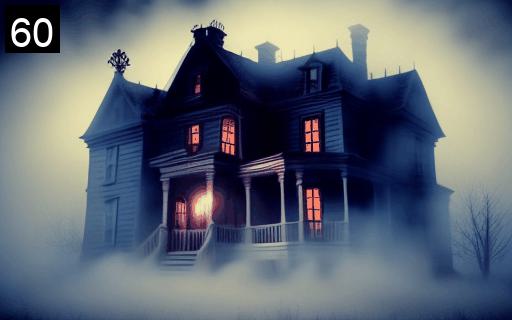} &
        \includegraphics[width=0.2\linewidth]{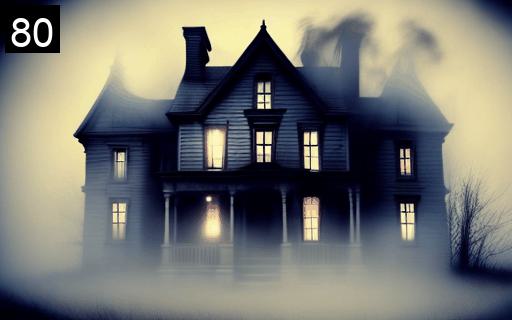} \vspace{1mm}\\
        % \shortstack[]{FIFO-\\Diffusion\vspace{8mm}} &
        \rotatebox[origin=c]{90}{\small Gen-L-Video\hspace{-14mm}} &
        \includegraphics[width=0.2\linewidth]{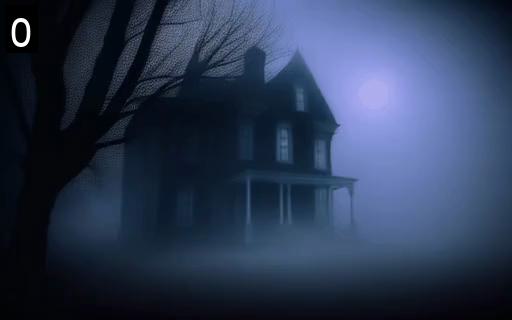} &
        \includegraphics[width=0.2\linewidth]{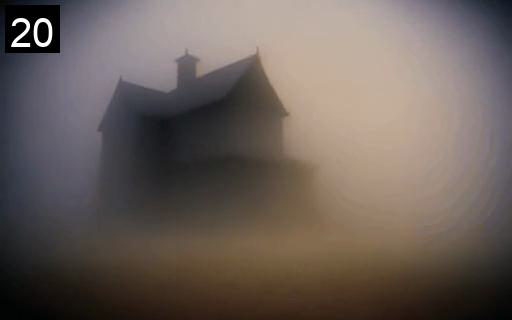} &
        \includegraphics[width=0.2\linewidth]{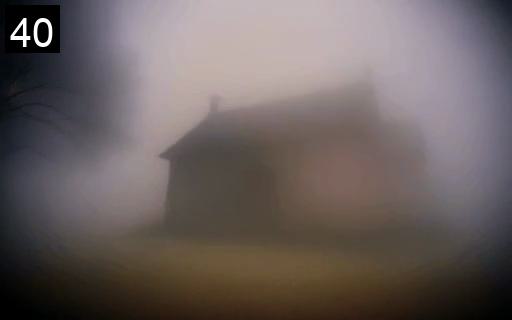} &
        \includegraphics[width=0.2\linewidth]{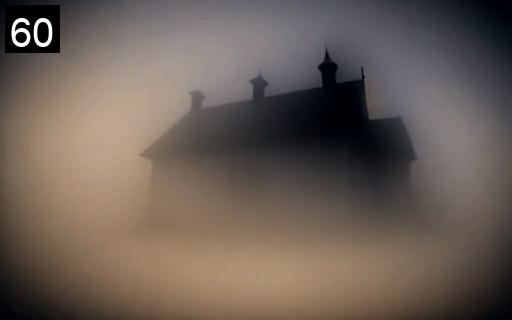} &
        \includegraphics[width=0.2\linewidth]{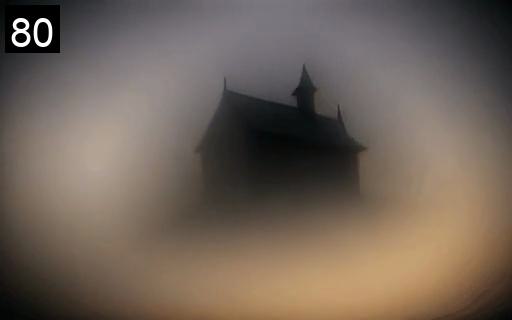} \vspace{1mm} \\
        % \shortstack[]{FIFO-\\Diffusion\vspace{8mm}} &
        \rotatebox[origin=c]{90}{\small LaVie+SEINE\hspace{-14mm}} &
        \includegraphics[width=0.2\linewidth]{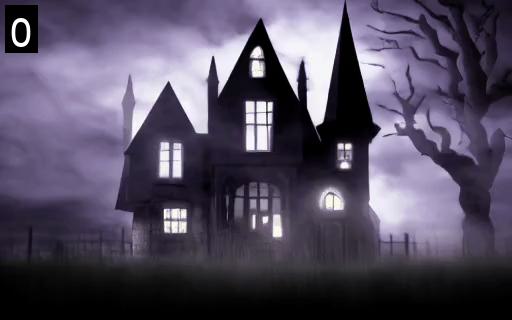} &
        \includegraphics[width=0.2\linewidth]{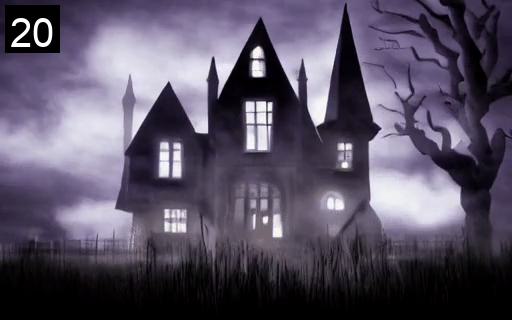} &
        \includegraphics[width=0.2\linewidth]{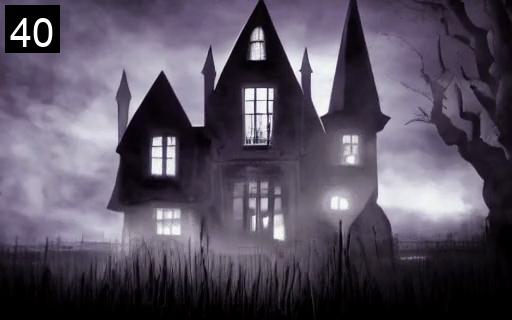} &
        \includegraphics[width=0.2\linewidth]{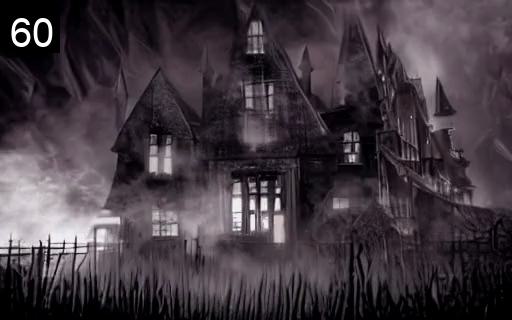} &
        \includegraphics[width=0.2\linewidth]{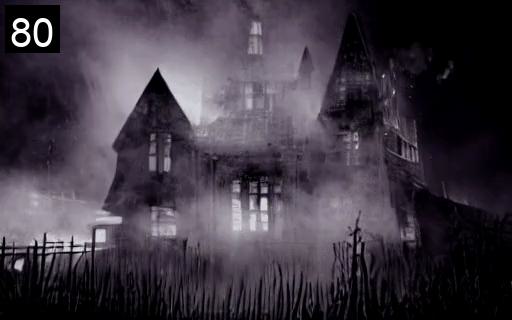} \vspace{1mm} \\
        & \multicolumn{5}{c}{\small (b) \textsf{"A spooky haunted house, foggy night, high definition."}} \vspace{3mm}\\

    \end{tabular}
    }
    \caption{
        Qualitative comparisons with other long video generation techniques, Gen-L-Video, FreeNoise, and LaVie + SEINE.
        The number in the top-left corner of each frame indicates the frame index.
    }\label{fig:qual_comparison_app2}
    \vspace{-3mm}
\end{figure*}

In \cref{fig:qual_comparison_app1,fig:qual_comparison_app2}, we provide more qualitative comparisons with other longer video generation methods, FreeNoise~\citep{qiu2023freenoise}, Gen-L-Video~\citep{wang2023genl}, and LaVie~\citep{wang2023lavie} + SEINE~\citep{chen2023seine}.

%%%%%%%%%%%%%%%%%%%%%%%%%%%%%%%%%%%%%%%%%%%%%%%%%%%%%%%%%%%%%%%%%%%%%%%%%%%%%%%%%%%%%%%%%%%%%%%%%%
\clearpage
\section{Motion evaluation}
\label{app:motion}
\begin{figure}[h]
    \centering
    \includegraphics[width=0.4\linewidth]{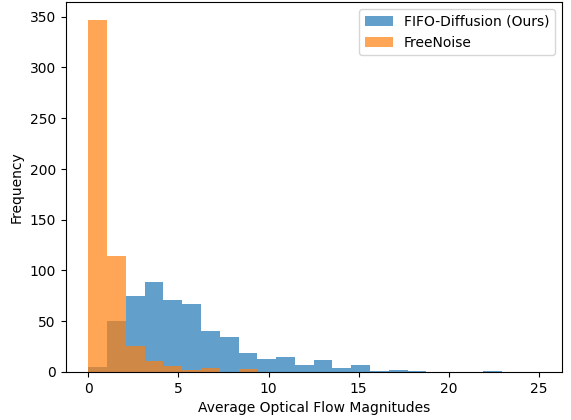} 
    \caption{Comparison of optical flow magnitudes between FIFO-Diffusion and FreeNoise.}\label{fig:optical_flow}
\end{figure}

We measure optical flow magnitudes (i.e. average of optical flow magnitudes) to compare the amount of motion between FIFO-Diffusion and FreeNoise, for the videos generated with randomly sampled prompts from the MSR-VTT~\citep{xu2016msrvtt} test set.
\cref{fig:optical_flow} illustrates that over 65\% of videos generated by FreeNoise are located in the first bin, indicating significantly less motion compared to FIFO-Diffusion. In contrast, our method generates videos with a broader range of motion.

%%%%%%%%%%%%%%%%%%%%%%%%%%%%%%%%%%%%%%%%%%%%%%%%%%%%%%%%%%%%%%%%%%%%%%%%%%%%%%%%%%%%%%%%%%%%%%%%%%
\clearpage
\section{Ablation study}
\label{app:ablation}
\begin{figure*}[h]
    \centering
    \renewcommand{\arraystretch}{0.7}
    \scalebox{0.98}{
    \setlength{\tabcolsep}{1pt}
    \hspace{-3mm}
    \begin{tabular}{cccccc}
        \vspace{3mm}\\
	\rotatebox[origin=c]{90}{\small DD\hspace{-14mm}} &
        \includegraphics[width=0.2\linewidth]{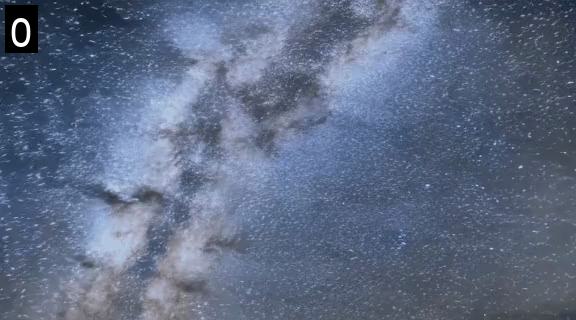} &
        \includegraphics[width=0.2\linewidth]{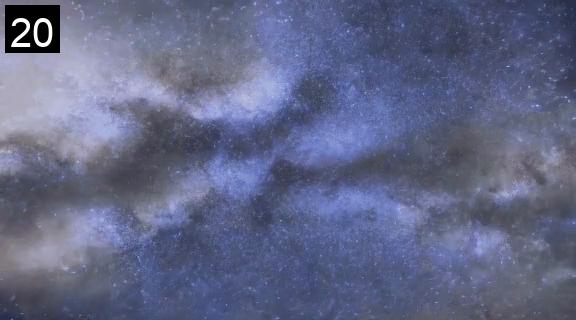} &
        \includegraphics[width=0.2\linewidth]{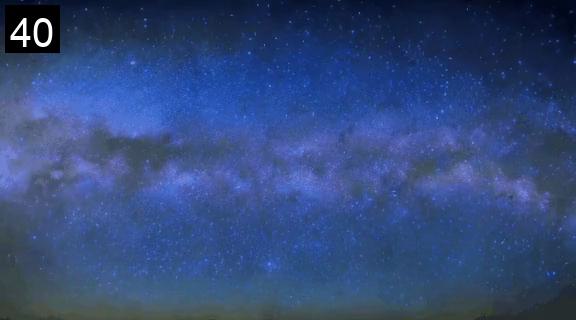} &
        \includegraphics[width=0.2\linewidth]{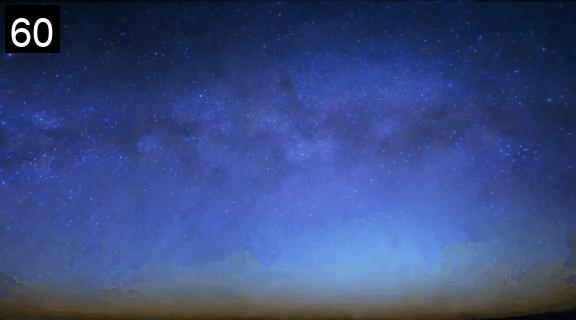} &
        \includegraphics[width=0.2\linewidth]{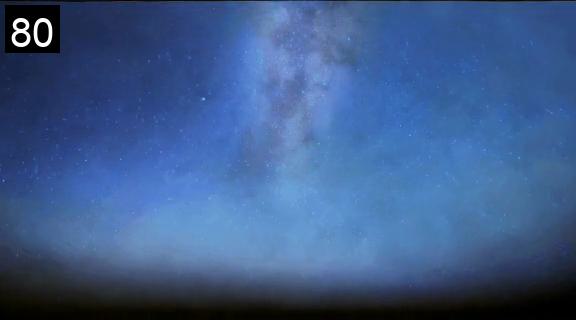} \vspace{1mm} \\
        % \shortstack[]{FreeNoise\\\vspace{8mm}} &
        \rotatebox[origin=c]{90}{\small DD+LP\hspace{-14mm}} &
        \includegraphics[width=0.2\linewidth]{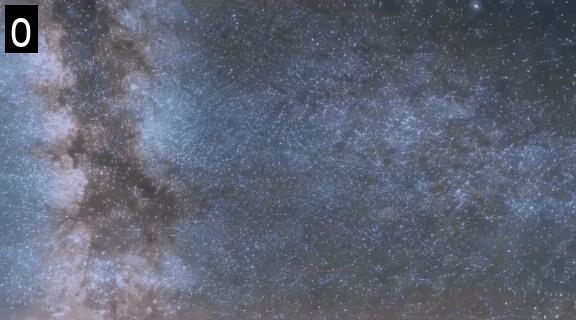} &
        \includegraphics[width=0.2\linewidth]{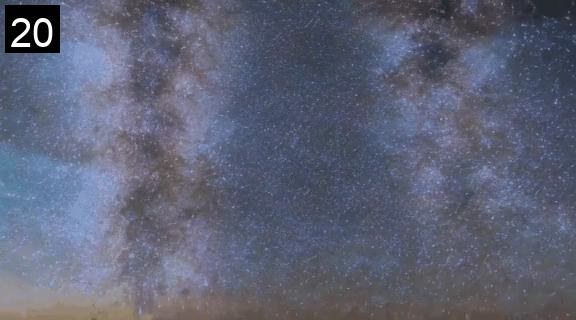} &
        \includegraphics[width=0.2\linewidth]{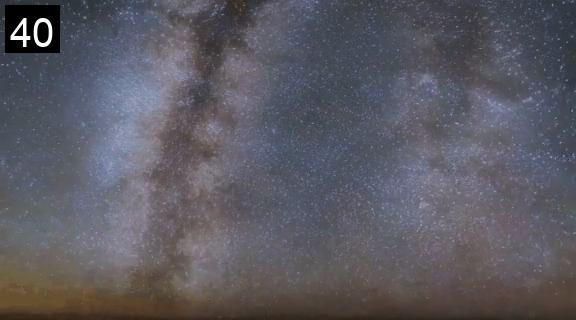} &
        \includegraphics[width=0.2\linewidth]{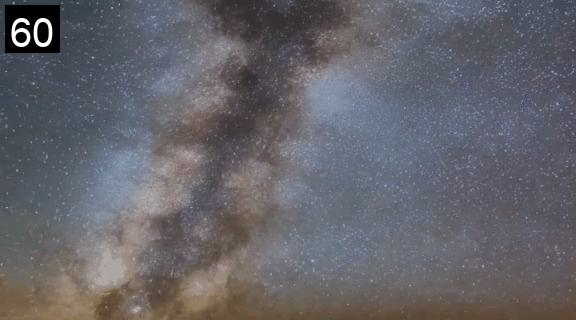} &
        \includegraphics[width=0.2\linewidth]{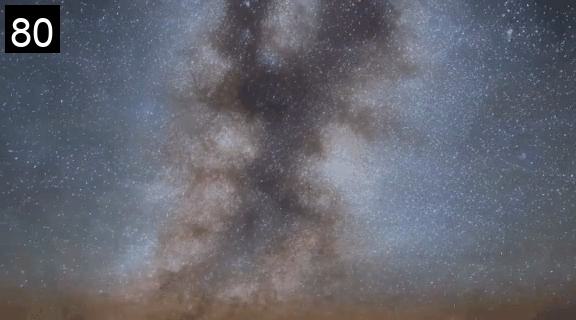} \vspace{1mm}\\
        % \shortstack[]{FIFO-\\Diffusion\vspace{8mm}} &
        \rotatebox[origin=c]{90}{\small DD+LP+LD\hspace{-14mm}} &
        \includegraphics[width=0.2\linewidth]{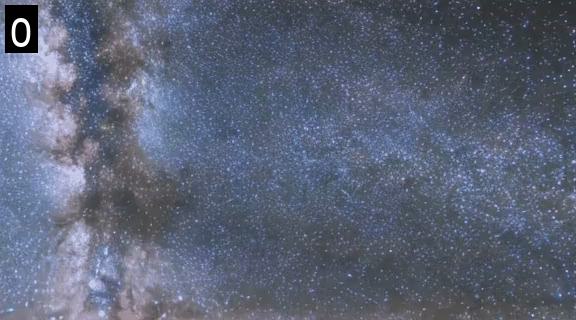} &
        \includegraphics[width=0.2\linewidth]{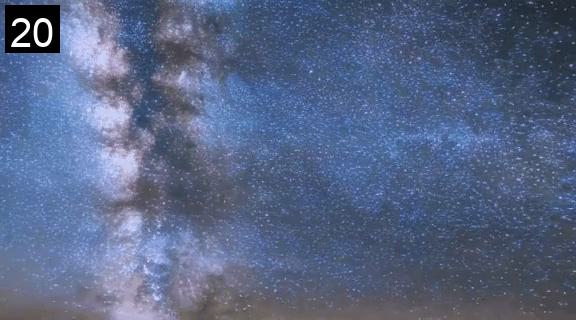} &
        \includegraphics[width=0.2\linewidth]{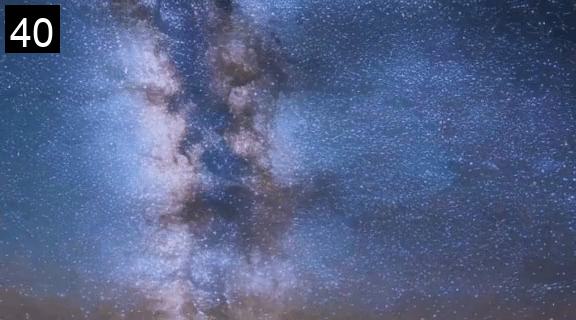} &
        \includegraphics[width=0.2\linewidth]{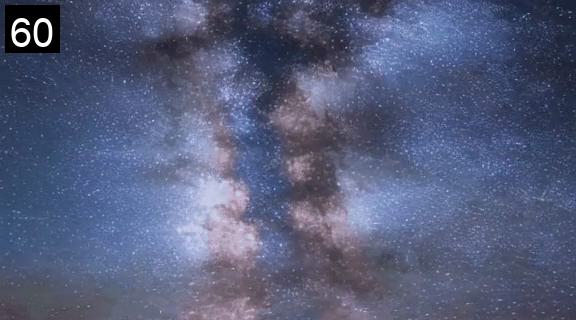} &
        \includegraphics[width=0.2\linewidth]{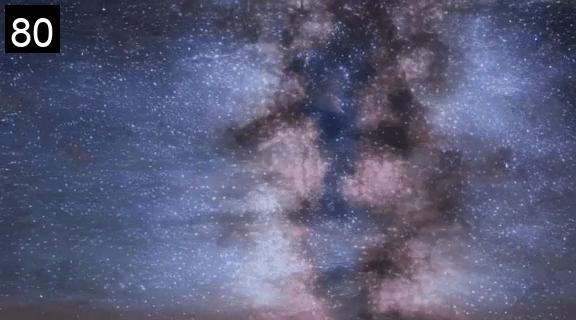} \vspace{1mm} \\
        % \shortstack[]{FIFO-\\Diffusion\vspace{8mm}} &
        & \multicolumn{5}{c}{\small (a) \textsf{"A panoramic view of the Milky Way, ultra HD."}} \vspace{6mm}\\
        \rotatebox[origin=c]{90}{\small DD\hspace{-14mm}} &
        \includegraphics[width=0.2\linewidth]{fig/zeroscope_ablation_jpgs/12/dd/0.jpg} &
        \includegraphics[width=0.2\linewidth]{fig/zeroscope_ablation_jpgs/12/dd/1.jpg} &
        \includegraphics[width=0.2\linewidth]{fig/zeroscope_ablation_jpgs/12/dd/2.jpg} &
        \includegraphics[width=0.2\linewidth]{fig/zeroscope_ablation_jpgs/12/dd/3.jpg} &
        \includegraphics[width=0.2\linewidth]{fig/zeroscope_ablation_jpgs/12/dd/4.jpg} \vspace{1mm} \\
        % \shortstack[]{FreeNoise\\\vspace{8mm}} &
        \rotatebox[origin=c]{90}{\small DD+LP\hspace{-14mm}} &
        \includegraphics[width=0.2\linewidth]{fig/zeroscope_ablation_jpgs/12/dd_lp/0.jpg} &
        \includegraphics[width=0.2\linewidth]{fig/zeroscope_ablation_jpgs/12/dd_lp/1.jpg} &
        \includegraphics[width=0.2\linewidth]{fig/zeroscope_ablation_jpgs/12/dd_lp/2.jpg} &
        \includegraphics[width=0.2\linewidth]{fig/zeroscope_ablation_jpgs/12/dd_lp/3.jpg} &
        \includegraphics[width=0.2\linewidth]{fig/zeroscope_ablation_jpgs/12/dd_lp/4.jpg} \vspace{1mm}\\
        % \shortstack[]{FIFO-\\Diffusion\vspace{8mm}} &
        \rotatebox[origin=c]{90}{\small DD+LP+LD\hspace{-14mm}} &
        \includegraphics[width=0.2\linewidth]{fig/zeroscope_ablation_jpgs/12/dd_lp_ld/0.jpg} &
        \includegraphics[width=0.2\linewidth]{fig/zeroscope_ablation_jpgs/12/dd_lp_ld/1.jpg} &
        \includegraphics[width=0.2\linewidth]{fig/zeroscope_ablation_jpgs/12/dd_lp_ld/2.jpg} &
        \includegraphics[width=0.2\linewidth]{fig/zeroscope_ablation_jpgs/12/dd_lp_ld/3.jpg} &
        \includegraphics[width=0.2\linewidth]{fig/zeroscope_ablation_jpgs/12/dd_lp_ld/4.jpg} \vspace{1mm} \\
        % \shortstack[]{FIFO-\\Diffusion\vspace{8mm}} &
        & \multicolumn{5}{c}{\small (b) \textsf{"A scenic cruise ship journey at sunset, ultra HD."}} \vspace{6mm} \\
        % \shortstack[]{FIFO-\\Diffusion\vspace{8mm}} &
	\rotatebox[origin=c]{90}{\small DD\hspace{-14mm}} &
        \includegraphics[width=0.2\linewidth]{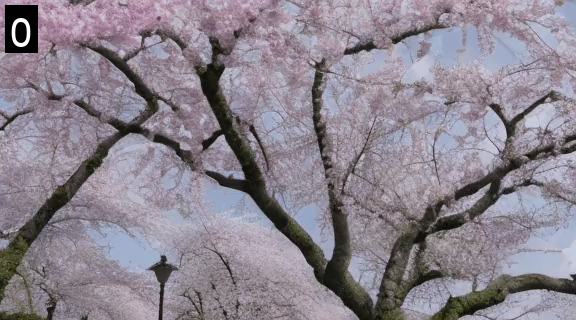} &
        \includegraphics[width=0.2\linewidth]{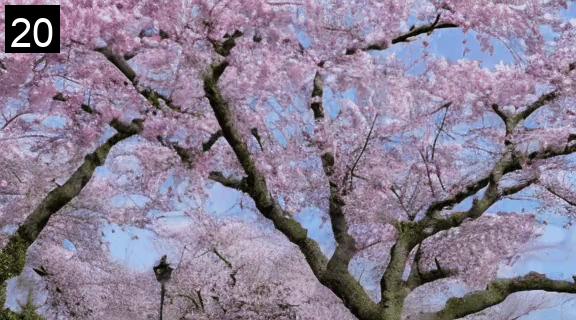} &
        \includegraphics[width=0.2\linewidth]{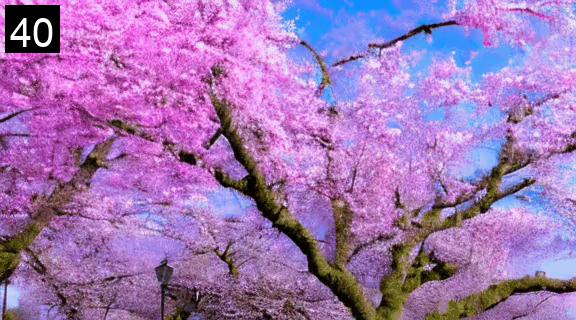} &
        \includegraphics[width=0.2\linewidth]{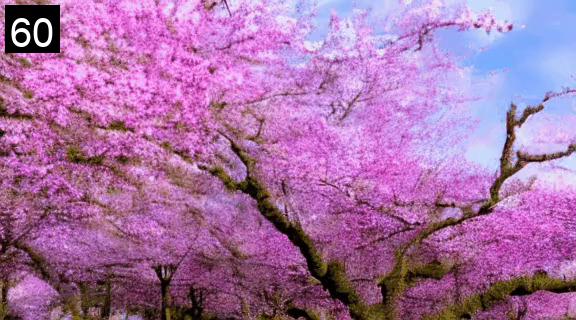} &
        \includegraphics[width=0.2\linewidth]{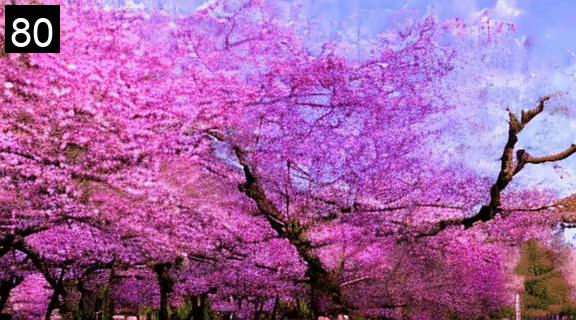} \vspace{1mm} \\
        % \shortstack[]{FreeNoise\\\vspace{8mm}} &
        \rotatebox[origin=c]{90}{\small DD+LP\hspace{-14mm}} &
        \includegraphics[width=0.2\linewidth]{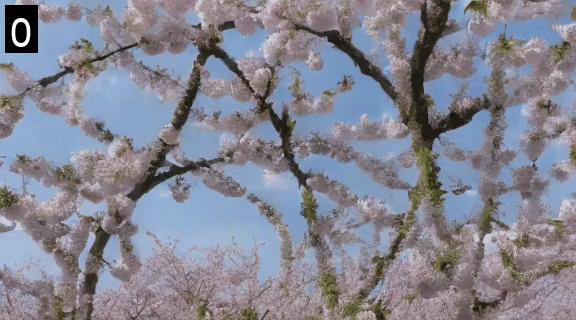} &
        \includegraphics[width=0.2\linewidth]{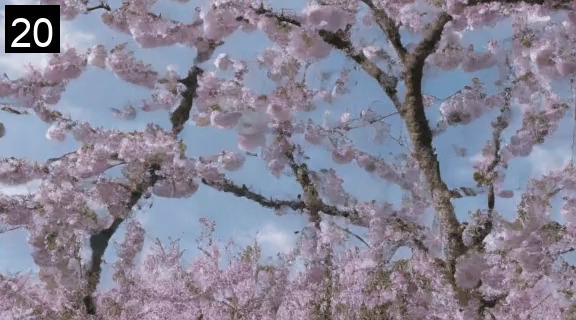} &
        \includegraphics[width=0.2\linewidth]{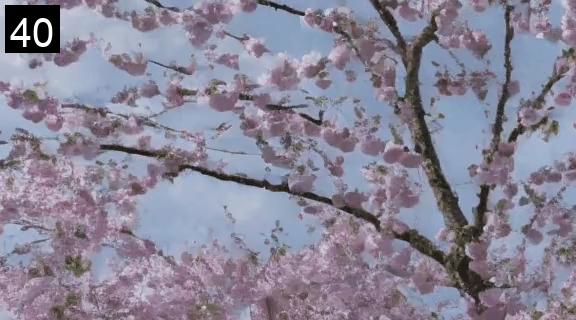} &
        \includegraphics[width=0.2\linewidth]{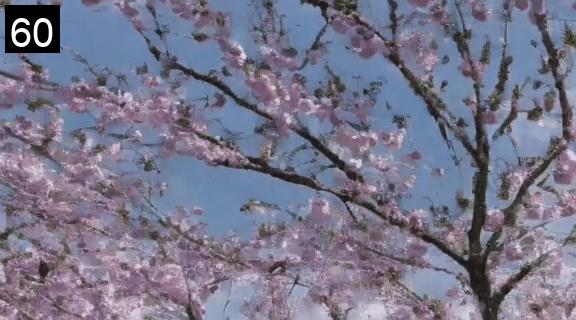} &
        \includegraphics[width=0.2\linewidth]{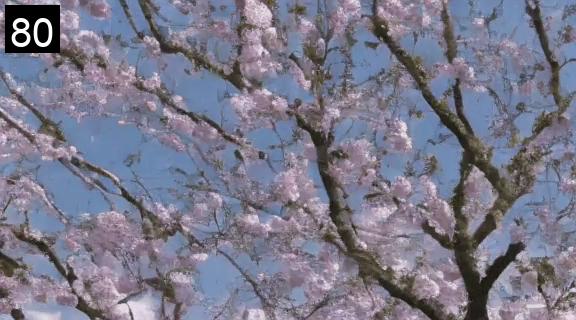} \vspace{1mm}\\
        % \shortstack[]{FIFO-\\Diffusion\vspace{8mm}} &
        \rotatebox[origin=c]{90}{\small DD+LP+LD\hspace{-14mm}} &
        \includegraphics[width=0.2\linewidth]{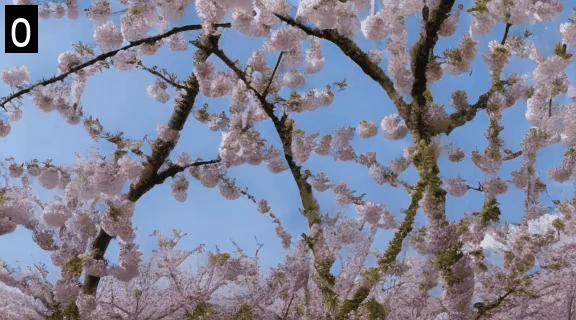} &
        \includegraphics[width=0.2\linewidth]{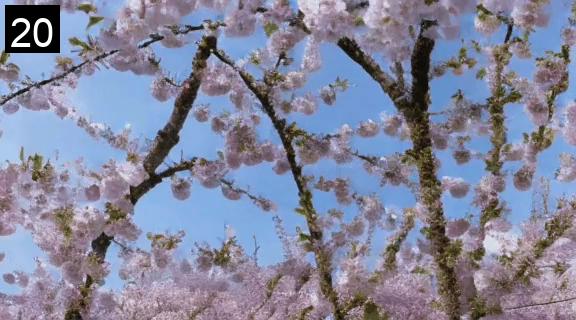} &
        \includegraphics[width=0.2\linewidth]{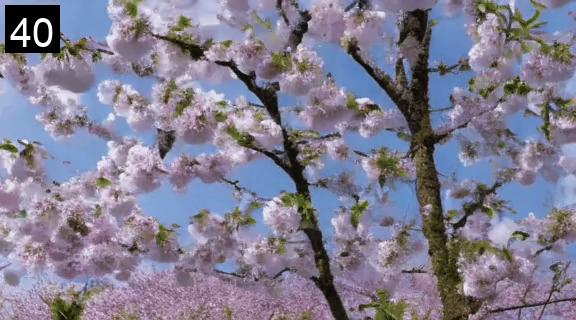} &
        \includegraphics[width=0.2\linewidth]{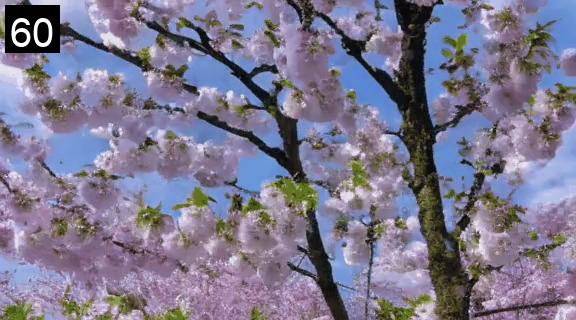} &
        \includegraphics[width=0.2\linewidth]{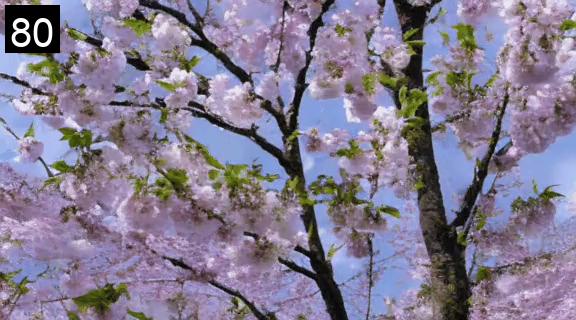} \vspace{1mm} \\
        % \shortstack[]{FIFO-\\Diffusion\vspace{8mm}} &
        & \multicolumn{5}{c}{\small (c) \textsf{"A beautiful cherry blossom festival, time-lapse, high quality."}}\\
        
    \end{tabular}
    }
    \caption{
        Ablation study.
        DD, LP, and LD signifies diagonal denoising, latent partitioning, and lookahead denoising, respectively.
        The number on the top-left corner of each frame indicates the frame index.
    }\label{fig:ablation_1}
    \vspace{-3mm}
\end{figure*}

\begin{figure*}[h]
    \centering
    \renewcommand{\arraystretch}{0.7}
    \scalebox{0.98}{
    \setlength{\tabcolsep}{1pt}
    \hspace{-3mm}
    \begin{tabular}{cccccc}
        \rotatebox[origin=c]{90}{\small DD\hspace{-14mm}} &
        \includegraphics[width=0.2\linewidth]{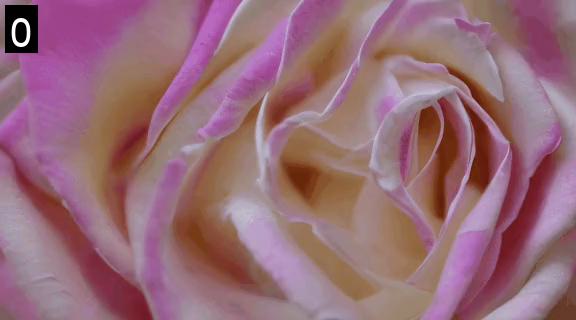} &
        \includegraphics[width=0.2\linewidth]{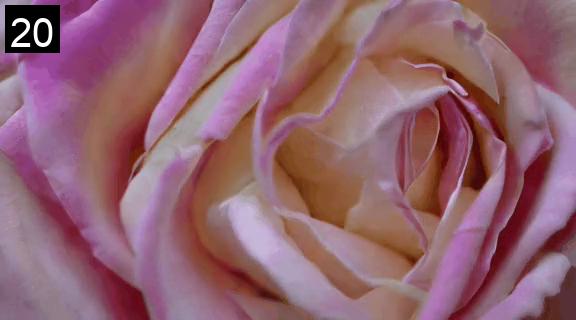} &
        \includegraphics[width=0.2\linewidth]{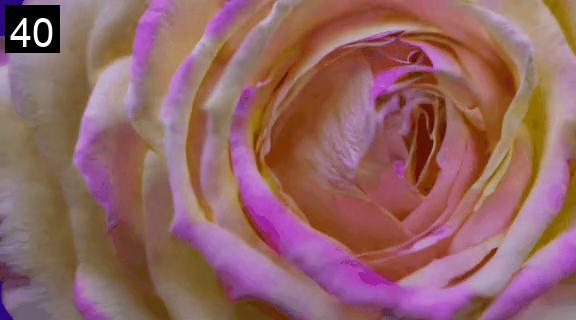} &
        \includegraphics[width=0.2\linewidth]{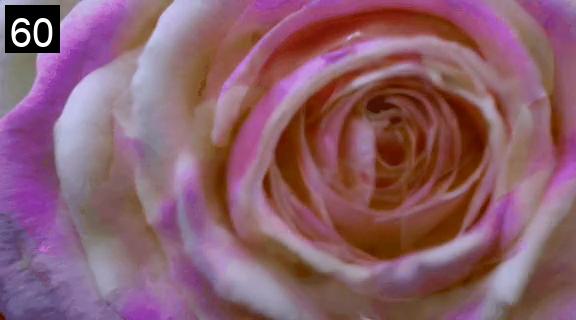} &
        \includegraphics[width=0.2\linewidth]{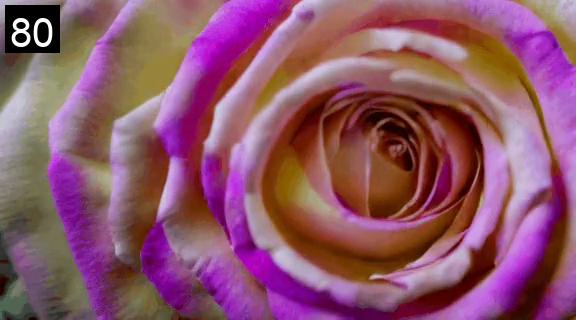} \vspace{1mm} \\
        % \shortstack[]{FreeNoise\\\vspace{8mm}} &
        \rotatebox[origin=c]{90}{\small DD+LP\hspace{-14mm}} &
        \includegraphics[width=0.2\linewidth]{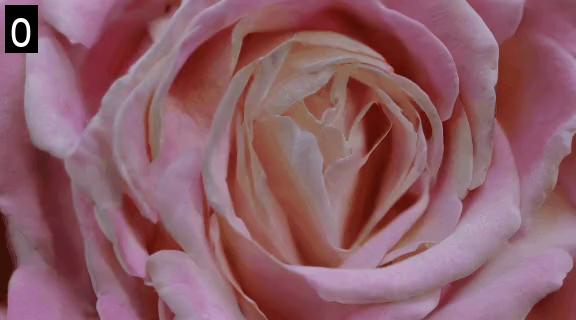} &
        \includegraphics[width=0.2\linewidth]{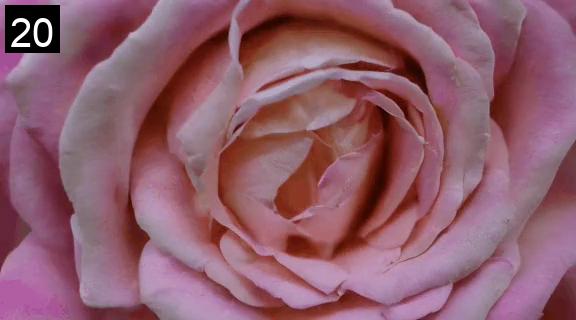} &
        \includegraphics[width=0.2\linewidth]{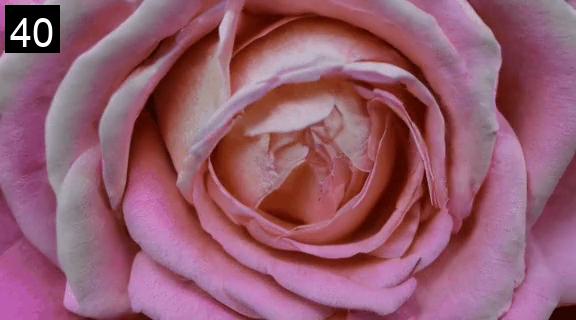} &
        \includegraphics[width=0.2\linewidth]{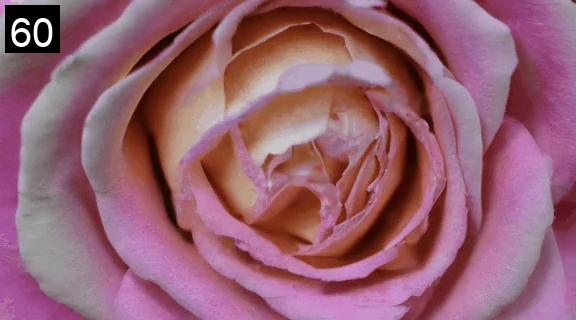} &
        \includegraphics[width=0.2\linewidth]{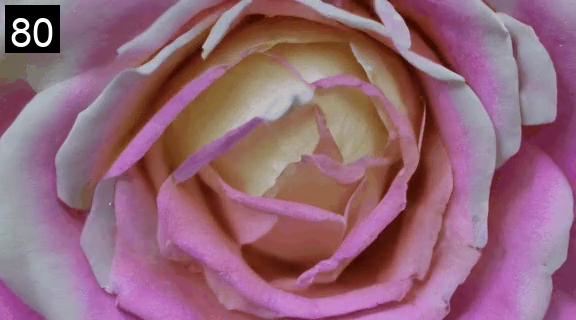} \vspace{1mm}\\
        % \shortstack[]{FIFO-\\Diffusion\vspace{8mm}} &
        \rotatebox[origin=c]{90}{\small DD+LP+LD\hspace{-14mm}} &
        \includegraphics[width=0.2\linewidth]{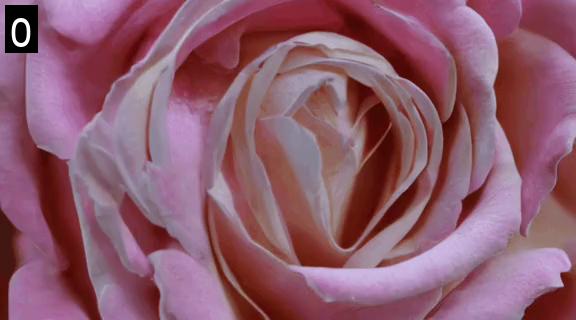} &
        \includegraphics[width=0.2\linewidth]{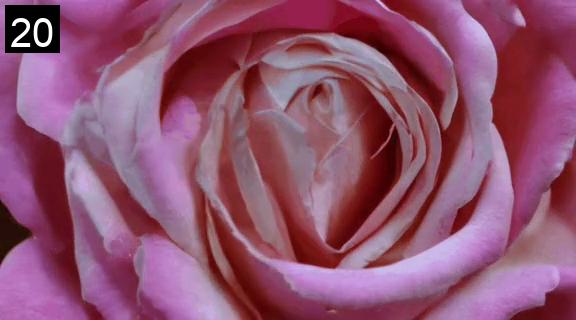} &
        \includegraphics[width=0.2\linewidth]{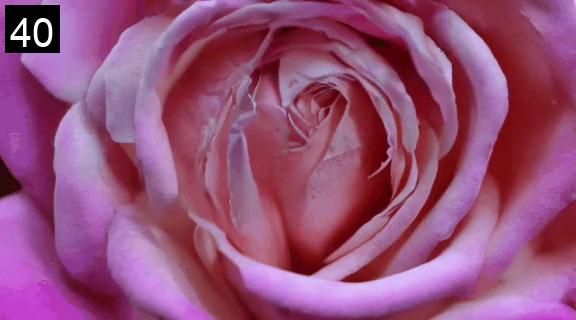} &
        \includegraphics[width=0.2\linewidth]{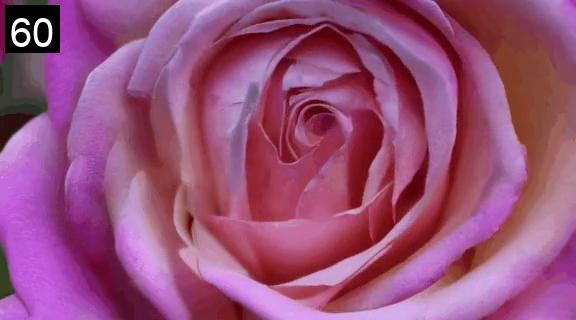} &
        \includegraphics[width=0.2\linewidth]{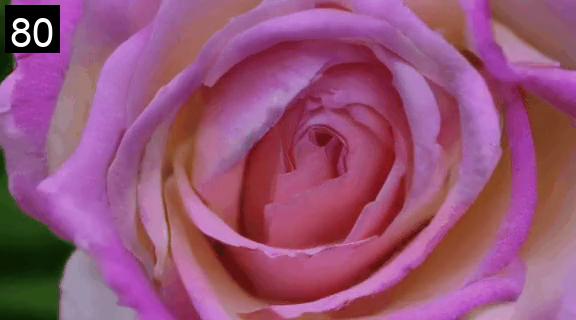} \vspace{1mm} \\
        % \shortstack[]{FIFO-\\Diffusion\vspace{8mm}} &
        & \multicolumn{5}{c}{\small (a) \textsf{"A detailed macro shot of a blooming rose, 4K."}} \vspace{6mm}\\
        % \shortstack[]{FIFO-\\Diffusion\vspace{8mm}} &
	\rotatebox[origin=c]{90}{\small DD\hspace{-14mm}} &
        \includegraphics[width=0.2\linewidth]{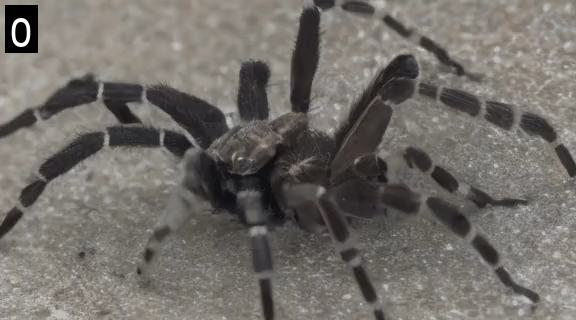} &
        \includegraphics[width=0.2\linewidth]{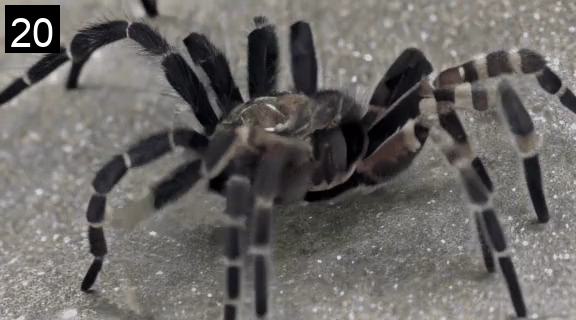} &
        \includegraphics[width=0.2\linewidth]{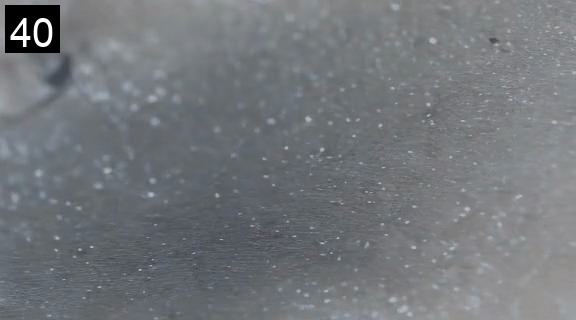} &
        \includegraphics[width=0.2\linewidth]{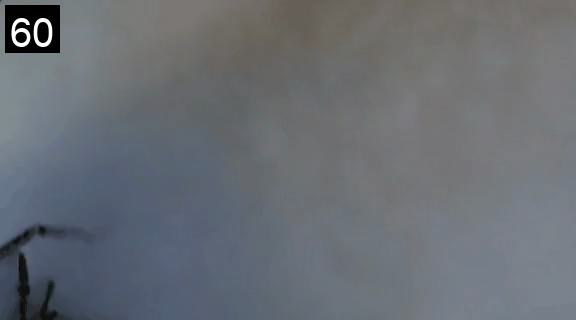} &
        \includegraphics[width=0.2\linewidth]{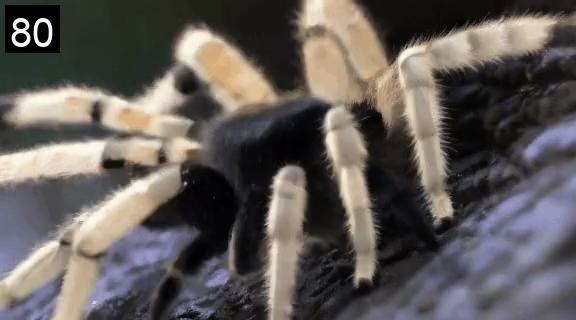} \vspace{1mm} \\
        % \shortstack[]{FreeNoise\\\vspace{8mm}} &
        \rotatebox[origin=c]{90}{\small DD+LP\hspace{-14mm}} &
        \includegraphics[width=0.2\linewidth]{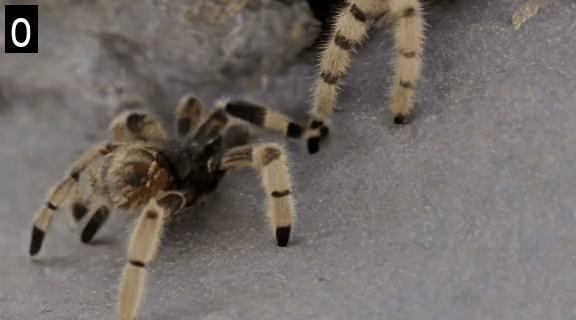} &
        \includegraphics[width=0.2\linewidth]{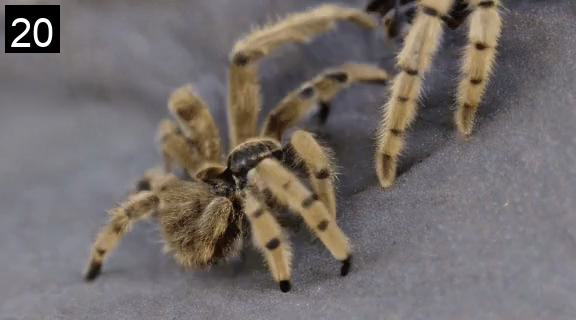} &
        \includegraphics[width=0.2\linewidth]{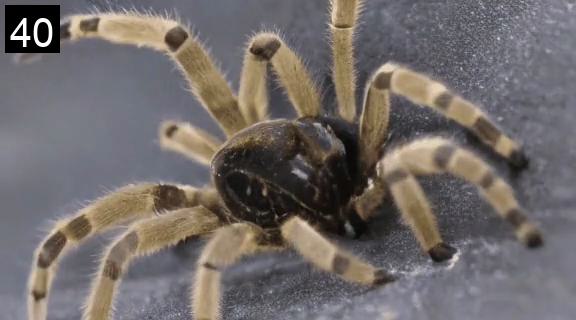} &
        \includegraphics[width=0.2\linewidth]{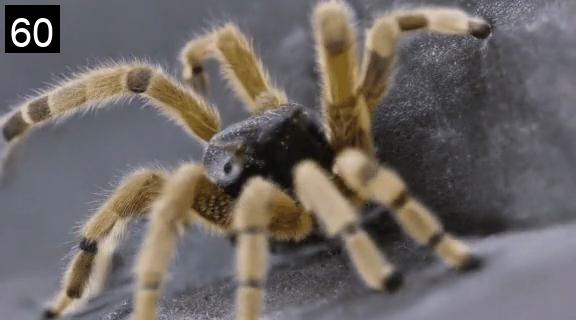} &
        \includegraphics[width=0.2\linewidth]{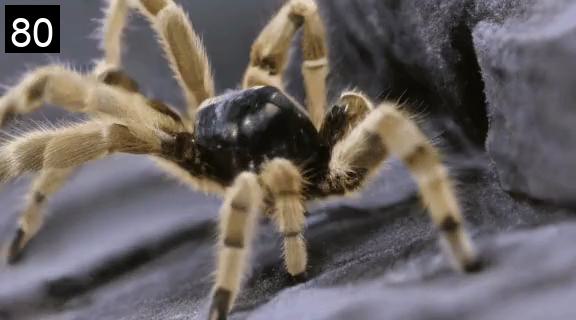} \vspace{1mm}\\
        % \shortstack[]{FIFO-\\Diffusion\vspace{8mm}} &
        \rotatebox[origin=c]{90}{\small DD+LP+LD\hspace{-14mm}} &
        \includegraphics[width=0.2\linewidth]{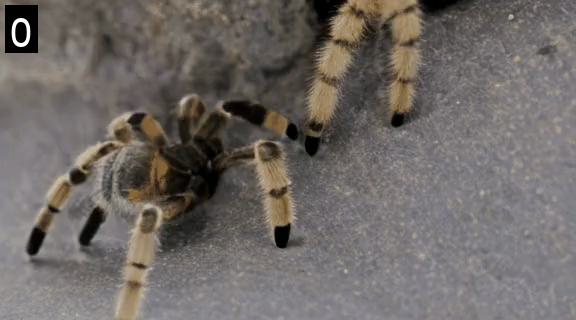} &
        \includegraphics[width=0.2\linewidth]{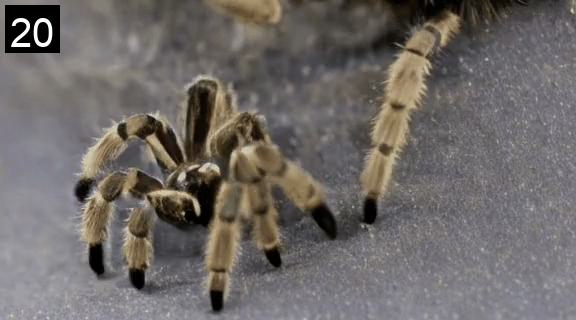} &
        \includegraphics[width=0.2\linewidth]{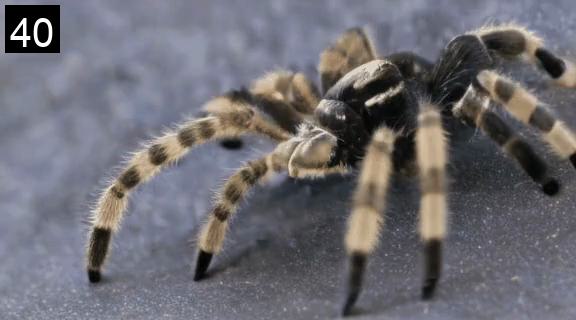} &
        \includegraphics[width=0.2\linewidth]{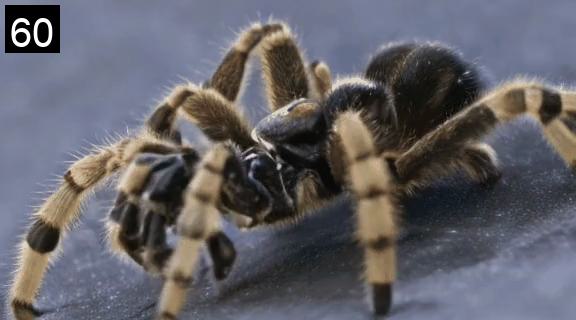} &
        \includegraphics[width=0.2\linewidth]{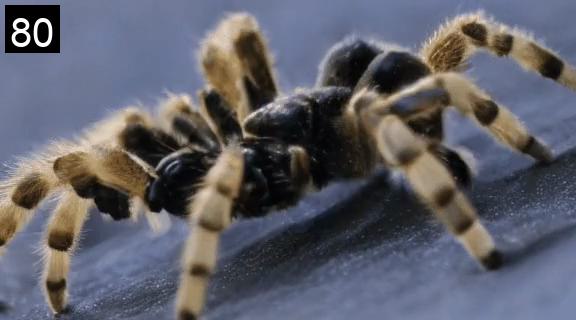} \vspace{1mm} \\
        % \shortstack[]{FIFO-\\Diffusion\vspace{8mm}} &
        & \multicolumn{5}{c}{\small (b) \textsf{"A close-up of a tarantula walking, high definition."}} \\
    \end{tabular}
    }
    \caption{
        Ablation study.
        DD, LP, and LD signifies diagonal denoising, latent partitioning, and lookahead denoising, respectively.
        The number on the top-left corner of each frame indicates the frame index.
    }\label{fig:ablation_2}
    \vspace{-3mm}
\end{figure*}

In \cref{fig:ablation_1,fig:ablation_2}, we conduct an ablation study to investigate the effectiveness of each component in FIFO-Diffusion.
We compare the results of FIFO-Diffusion only with diagonal denoising (DD), with the addition of latent partitioning with n=4 (DD + LP), and lookahead denoising (DD + LP + LD).

\section{Potential Broader Impact}
\label{subsec:impact}
This paper leverages pretrained video diffusion models to generate high quality videos. 
The proposed method can potentially be used to synthesize videos with unexpectedly inappropriate content since it is based on pretrained models and involves no training.
However, we believe that our method could mildly address ethical concerns associated with the training data of generative models.

\end{document}